\documentclass[a4paper,12pt,twoside,openright]{report}

\def\authorname{Sami \ Alabed\xspace}
\def\authorcollege{Girton College\xspace}
\def\authoremail{sa894@cam.ac.uk}
\def\dissertationtitle{RLCache: Automated Cache Management Using Reinforcement Learning.}
\def\wordcount{15,000} 
\usepackage{epsfig,graphicx,parskip,setspace,tabularx,xspace,import} 
\usepackage{subcaption}
\usepackage{booktabs} 
\usepackage{siunitx}  
\usepackage{algorithm,algpseudocode}
\usepackage[hidelinks]{hyperref}

\begin{document}

\pagestyle{empty}
\singlespacing
\begin{titlepage} 

\begin{center}
\noindent
\huge
\dissertationtitle \\
\vspace*{\stretch{1}}
\end{center}

\begin{center}
\noindent
\huge
\authorname \\
\Large
\authorcollege      \\[24pt]
\includegraphics{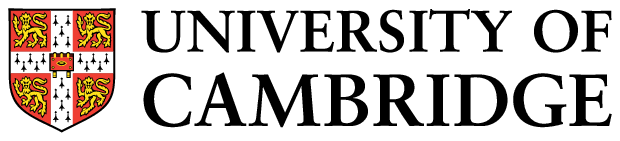}
\end{center}

\vspace{24pt} 

\begin{center}
\noindent
\large
{\it A dissertation submitted to the University of Cambridge \\ 
in partial fulfilment of the requirements for the degree of \\ 
Master of Philosophy in Advanced Computer Science} 
\vspace*{\stretch{1}}
\end{center}

\begin{center}
\noindent
University of Cambridge \\
Computer Laboratory     \\
William Gates Building  \\
15 JJ Thomson Avenue    \\
Cambridge CB3 0FD       \\
{\sc United Kingdom}    \\
\end{center}

\begin{center}
\noindent
Email: \authoremail \\
\end{center}

\begin{center}
\noindent
\today
\end{center}

\end{titlepage} 

\newpage
\vspace*{\fill}

\onehalfspacing
\newpage
{\Huge \bf Declaration}

\vspace{24pt} 

I \authorname of \authorcollege, being a candidate for the M.Phil in
Advanced Computer Science, hereby declare that this report and the
work described in it are my own work, unaided except as may be
specified below, and that the report does not contain material that
has already been used to any substantial extent for a comparable
purpose.

\vspace{24pt}
Total word count: \wordcount

\vspace{60pt}
\textbf{Signed}: 

\vspace{12pt}
\textbf{Date}:

\vfill

This dissertation is copyright \copyright 2019 \authorname. 
\\
All trademarks used in this dissertation are hereby acknowledged.

\newpage
\vspace*{\fill}

\singlespacing
\newpage
{\Huge \bf Acknowledgements}
\vspace{24pt} 

I would like to express gratitude to my supervisor Dr. Eiko Yoneki for valuable and constructive suggestions during the planning and development of this research work. 
I would also like to especially thank Michael Schaarschmidt for the constant mentoring, intellectual discussions, and encouragements throughout the various ups and downs of this project.
Additionally, I would like to thank Raad Aldakhil and Mihai Bujanca for their proof-reading of this report and comments.

\newpage
\vspace*{\fill}

\singlespacing
\newpage
{\Huge \bf Abstract}
\vspace{24pt} 

This study investigates the use of reinforcement learning to guide a general purpose cache manager decisions. 
Cache managers directly impact the overall performance of computer systems. 
They govern decisions about which objects should be cached, the duration they should be cached for, and decides on which objects to evict from the cache if it is full.
These three decisions impact both the cache hit rate and size of the storage that is needed to achieve that cache hit rate. 
An optimal cache manager will avoid unnecessary operations, maximise the cache hit rate which results in fewer round trips to a slower backend storage system, and minimise the size of storage needed to achieve a high hit-rate.

Current approaches assume characteristics of underlying data and use rule-based mechanisms that are tailored for the general case.
Caches are exposed to a changing workload regularly, especially the caches at a lower level of the stack.
For example, a cache in front of a general purpose database that is used by several different services observes a varying level of traffic patterns.
Rule-based methods are static and do not adapt to changes in workload, resulting in the cache operating in a sub-optimal state.
Using reinforcement learning, the system learns ideal caching policies tailored to individual systems and the traffic pattern it observes without any prior assumption about the data. 

This project investigates using reinforcement learning in cache management by designing three separate agents for each of the cache manager tasks.
Furthermore, the project investigates two advanced reinforcement learning architectures for multi-decision problems: a single multi-task agent and a multi-agent.
We also introduce a framework to simplify the modelling of computer systems problems as a reinforcement learning task.
The framework abstracts delayed experiences observations and reward assignment in computer systems while providing a flexible way to scale to multiple agents.

Simulation results based on an established database benchmark system show that reinforcement learning agents can achieve a higher cache hit rate over heuristic driven algorithms while minimising the needed space.
They are also able to adapt to a changing workload and dynamically adjust their caching strategy accordingly.
The proposed cache manager model is generic and applicable to other types of caches, such as file system caches. 
This project is the first, to our knowledge, to model cache manager decisions as a multi-task control problem.

\newpage
\vspace*{\fill}

\pagenumbering{roman}
\setcounter{page}{0}
\pagestyle{plain}
\tableofcontents
\listoffigures

\onehalfspacing

\chapter{Introduction}
\graphicspath{{01-introduction/figs/}}

\pagenumbering{arabic} 
\setcounter{page}{1} 

Computer systems need to store their data in a reliable backend that ensure their durability and availability. 
These systems often include a cache for temporarily storing information closer to where it is needed to reduce the congestion on the permanent backends.
Caches are designed to provide fast read and write access to the data locally.

Caches have long been used to boost the performance of computer systems, such as in operating systems, file systems, and web servers.
This work focuses on web caches, but many ideas and design decisions are general and transferable to other cache systems. 
Web caches are critical to the performance of many systems and directly linked to user satisfaction \cite{arapakis2014impact, amazonMoney}. 
An experiment conducted by Google in $2009$ demonstrated that increasing web search latency $100$ to $400$ milliseconds reduces the daily
number of searches per user by $0.2\%$ to $0.6\%$ \cite{brutlag2009speed}.
Hence, optimising the performance of caches is important and the focus of our work.

However, the effectiveness of the cache is limited by how many times a client requests an object that is stored in the cache.
Furthermore, the cache cannot grow in size indefinitely, it needs to be managed and cleared of unwanted objects.
Hence the need for a Cache Manager (CM).
CMs use a set of methods and strategies to decide on: what objects to cache, how long they should be cached for, and what to remove when the cache is full. 
An optimal CM focuses on storing hot keys for long duration without occupying more space than necessary.
Occupied space is an important factor for caches, especially when the cache is on constrained devices such as smartphones.

Current approaches utilise heuristics to optimise performance for the general case.
While heuristics are designed to work for the general case, they do not push the system to its optimal performance and to further optimise the system developers write a set of hard-coded rules.
Additionally, even after tuning the system once, changes in the traffic patterns render the system in a sub-optimal configuration. 
Traffic patterns commonly fluctuate in social networks and news sites, where a sudden news outbreak causes a significant shift in the traffic pattern.

This study investigates using Reinforcement Learning (RL) to automate the optimisation of the CM decisions. 
RL has the ability to learn optimal heuristics tailored to each unique pattern and dynamically adjust its decision-making policies by monitoring the system metrics. 
However, the delay in computer system poses a real challenge to RL algorithms that model their interaction as a sequential decision problem.
When decisions are only observed in the future, there needs to be a mechanism to correlate RL agent's action to the update and impact to the computer system Service-Level-Objective (SLO).
Additionally, a typical computer system has several SLOs that may require separate RL agents to learn different policies.
Therefore, there is a need for an abstraction that simplifies the delayed experience and allow as many agents to observe the impact of their actions simultaneously.

This work successfully implemented and evaluated several different RL agents for each one of the CM tasks.
Furthermore,  we investigated a multi-task agent that combines the three RL agents into a single agent that learns all the three key actions a CM makes.
We addressed the issues of applying RL algorithms to computer systems by designing a scalable framework that monitors system metrics, relates past actions to current observations, and issues them to relevant RL agents.
In our simulation running using YCSB benchmark\cite{cooper2010benchmarking}, an established benchmark in the database community, the RL agents were able to learn patterns in the traffic and outperform traditional heuristic while requiring less storage and the ability to handle changes in the traffic pattern.

This dissertation's key contributions are:
\begin{itemize}
	 \item Applied advanced RL approaches to solve a long-researched problem in computer systems while tackling the issues that arise due to the gap between applied computer system research and RL typical controlled environment. Providing a detailed discussion and implemented solution on issues such as delayed experience, modelling problem, and using SLOs to craft reward signals for RL agents.
    \item Implemented an online CM system that dynamically adapts to changing workload and learn optimal strategies for making caching decision (\autoref{section:rlcache_strategy}), eviction decisions (\autoref{section:rl_eviction}), and TTL estimation (\autoref{sec:ttl_agent}).
    \item This work, as far as we are aware, is the first to successfully introduce the use of a multi-task RL approach in a computer system (\autoref{sec:multi_task}).
    \item Designed a framework for system problems when actions have a delayed impact and demonstrated it by building three separate agents each learning an independent delayed task (\autoref{sec:observer}). 
\end{itemize}

This project does not address the latency concern overhead of using RL neither considers the distributed case of a CM, we discuss both cases in the future work \autoref{section:distributed_cache_manager} section.
We argue that our work stands by itself regardless, as the CM SLOs are the hit-rate, capacity, and utilisation, so optimising these provides a significant improvement to the status quo.

The dissertation structure is as follows:
\autoref{chapter:background} provides a brief background to caching, CMs, reinforcement learning, and using reinforcement learning in a computer system context. Additionally, it surveys current literature.
After that, \autoref{chapter:cache_manager}  discusses the design behind the observer framework architecture, a framework that simplifies scaling and using RL for computer systems. 
Furthermore, the section covers the algorithms, models, and implementation details of the RL agents this work uses.
Next, \autoref{chapter:evaluation} covers the evaluation setup and evaluates the several agents' configurations and compare them to traditional algorithms.
Finally, \autoref{chapter:summary} summarises the findings and discuss possible future directions. 

\chapter{Background and Related Work}\label{chapter:background}
\graphicspath{{02-background/figs/}}

\section{Background}
This chapter sets the context for this work and summarises the background.
The first section introduces CMs and their common algorithms. 
Next, it presents reinforcement learning (RL) and two popular RL algorithms this work uses: Deep Q-network and Soft Actor-Critic.
Finally,  the last section surveys related work in applying optimisation techniques to CMs and compares this project to the literature.
\section{The Caching Problem}\label{sec:caching_background}
\subsection{Introduction to caches}
Latency was found to be linked directly to users satisfaction and revenue generation in multiple studies \cite{arapakis2014impact, amazonMoney, brutlag2009speed}. 
Many applications use databases hosted in data centres to store persistent information.
These databases write to single or distributed disks to ensure the persistence and durability of the information.
However, the combined cost of network communication and the long processing time to these databases pose a bottleneck to the system, making it a challenge to scale the database to handle a large number of client requests.
The desire to improve the performance of the system motivated the system community to look into a fast storage system that has less durability and consistency guarantees but scales easily. 
These systems are called caches. 
The cache stores a subset of the data and act as a layer that provides fast data access to the client while reducing the traffic stress on the origin server \cite{webCachingPerformance,rabinovich2002web}.
The cache can be located locally on a device (e.g. a smartphone memory), on the edge of a network (e.g. a content delivery network CDN \cite{choi2011survey}), hosted near the database server (e.g. Redis \cite{carlson2013redis}), or a mix of them all.  

\begin{figure}[!htb]
	\centering
	\includegraphics[width=1.0\linewidth]{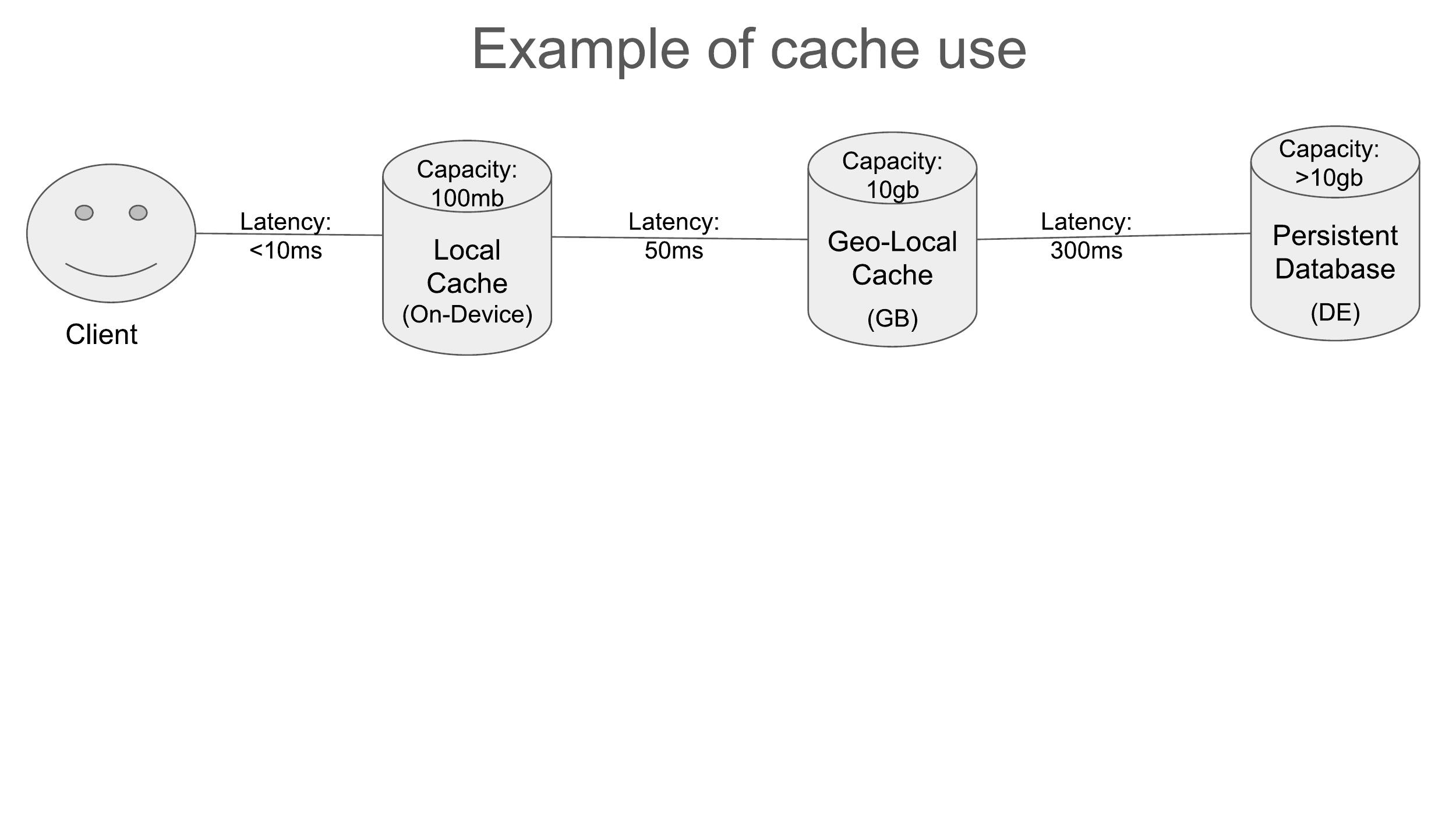}
	\caption[Example of a simple system using caches.]{A simple system that use geo-local cache, local cache, and a persistent database. With an example of possible sizes of those caches.}
	\label{fig:cachehellofig}
\end{figure}
To demonstrate the flow of the data between the end-user, cache, and a database, imagine a use-case of a cache in a news site where a video of the latest news will be played by all new visitors of the site.
The video is pushed to a geocache close to the users' location to improve their perceived latency and avoid relying a single backend system. 
When the video is requested, it is played from the cache providing a better experience to the end-users.
These users will also cache the content on their device for a much faster replay experience, illustrating the ability of the layering stacks as an onion, as shown in figure \ref{fig:cachehellofig}.

Caches are useful if the stored contents are requested, to ensure that the relevant objects are cached and the cache size does not grow indefinitely, cache managers (CM) are used.
The next subsections discuss the main mechanisms that CMs govern. 
\subsection{Caching Strategy}\label{sec:caching_decision}
Caching strategy decides whether an object should be cached or not.
When a request asks for data that exists in the cache, that is called a \textit{cache hit}, and if it is not in the cache, it is a \textit{cache miss}. 
On a cache miss, the data is fetched from the slower backend and served to the client. 
Application developers have to decide in advance on a caching strategy that chooses which objects to cache. 
These decisions are often based on the type of data and access pattern.
Here we consider two of the common caching strategies with their advantages and disadvantages.

\subsubsection{Write-through} 
In write-through policy, the application writes new data updates to both the cache and the backend at the same time and only reports success when the write succeeds in both places at the same time.
This approach is useful for applications that have read-after-write access pattern.
For example, a social media post by a celebrity is very likely to be requested shortly after a write, thus proactively caching their media post is the optimal strategy and improves the performance for early requesters.
The disadvantage of this approach is the slower write speed as the write has to happen in two places, in addition to the potential of being an unnecessary operation. 
Unnecessary write operations pose an overhead when the job is a batch write job where the inserted items will not necessarily be queried after, or when the request itself does not generate additional read, for example after a log entry.

\subsubsection{Write-on-read} 
In write-on-read loads data into cache only on reads.
Upon a cache miss, write-on-read strategy retrieves the data from the slower backend and remembers the results for next time the item is requested. 
The data remains in cache until invalidated or exceed its expiry time.
The disadvantage of this approach is that in applications where the client is guaranteed to issue a read after a write, there is always going to be at least a single cache miss causing an additional network trip to the slower backend.

\subsection{Data Staleness}\label{background:ttl}
Data staleness occurs when an item is updated in the backend but not in the cache. 
This can happen when a cache write fails but an update to the backend succeeds, or if a database update bypassed the cache.
To ensure data freshness and to not clutter the cache with long-lasting objects, a CM sets a Time to live (TTL) on an object.
TTL is an integer that represents the seconds an object is stored in a cache until it is evicted (deleted). 
This mechanism guarantees a time-bounded weak consistency model \cite{terry2013replicated} (TTL bounded), web-caches depend on this model to reduce manual invalidation requests.
An optimal TTL is large enough to maximise the hit-rate of an object and expires before the cache observes an update or invalidation to this object.

A common mechanism for configuring a TTL is to configure a fixed (absolute) value at development time that is applicable to all objects.
The value usually is a large number that the application SLO tolerates but not necessarily the optimal.
Overestimating the TTL populates the cache needlessly with unwanted objects leading to the cache filling faster while underestimating the TTL leads to more cache misses.
%
%
\subsection{Cache Capacity}\label{background:eviction}
Caches are limited in size, especially the ones deployed on a client device.
When a cache exceeds its maximum capacity, a \textit{cache eviction} algorithm decides on which elements to evict.
There are several other cache eviction policies surveyed in \cite{podlipnig2003survey}. We chose the three most widely used algorithms as our baselines.

\subsubsection{Least recently used (LRU)}
LRU evicts the least recently used item based on the order of queries on the cache. 
LRU is the most widely used algorithm due to its simplicity and applicability to real-world patterns. 
The rationale behind the algorithm is that if a client requests an item every K time, the item will remain in the cache of size K indefinitely.
Thus, popular items will be cached as long as the caching capacity allows it.
LRU only uses the access time when making a decision, and this can be a disadvantage when the data is sharded (partitioned) across multiple databases that have varying access times to their objects some are slower than others.
For example, assume that retrieving object A takes twice as long as another object B. If the access pattern favours B, LRU will evict the one that was not accessed recently (A) and not the one that will improve the overall system performance (B).

\subsubsection{First in first out (FIFO)}
FIFO is a simple algorithm that evicts the item that was first inserted in the cache regardless of the access pattern.
FIFO is an useful algorithm for linear events. e.g. in web browsing, a client going back to the last visited page is a common browsing pattern, but returning to pages that they visited much earlier is uncommon.

\subsubsection{Least frequently used (LFU)}
LFU incorporates information about the access pattern by tracking objects cache hits and evicting the object with the least number of hits.
It is most useful for applications where a popular object is accessed throughout the application life-cycle, like caching the icon of a website.
However, LFU fails to capture real-world traffic patterns.
For example, in a social media a popular post might receive a large number of hits at some time frame, then lose interest with time, a pure LFU cache will keep that post in cache even when users are not interested in it anymore. 

\subsection{Remarks}
This section provided a quick overview of the caching problem, and highlighted that there are always at least two ways to make decisions depending on the use case.
Providing a solid argument for using dynamically adjustable policies that learn from the data and its access pattern.
The next section covers the topic of reinforcement learning and the difficulty of using it directly with a system problem.
\section{Reinforcement learning}
\subsection{Reinforcement Learning Essentials}
\begin{figure}[!htb]
	\centering
	\includegraphics[width=1.0\linewidth]{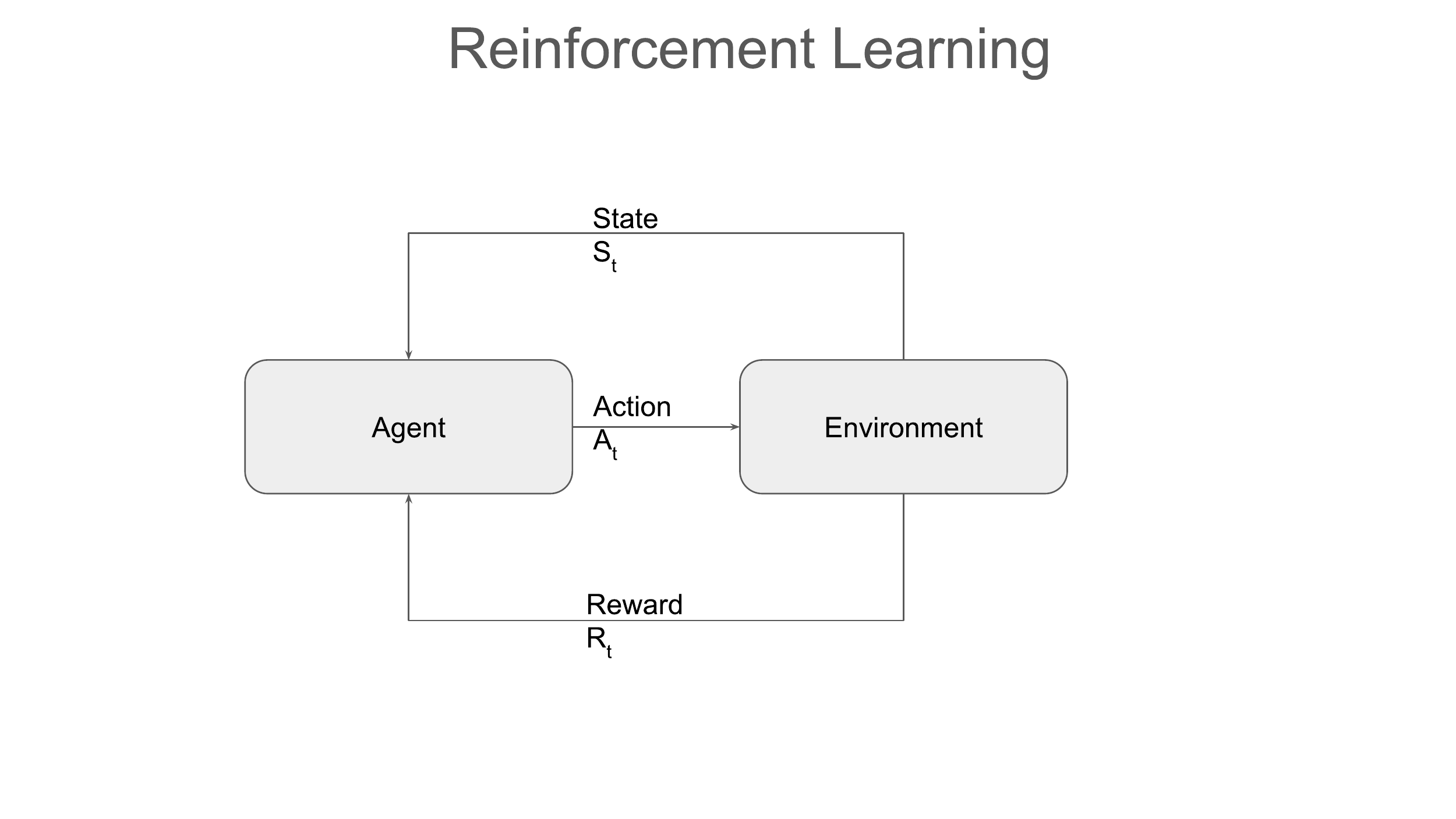}
	\caption[Simplified reinforcement learning environment architecture.]{The interaction between an agent and an environment in a reinforcement learning setting.}
	\label{fig:rlfig}
\end{figure}

Reinforcement learning (RL) is a sub-type of machine learning that learns how to operate in any setting dynamically. 
It has been successfully applied in various problems such as playing Atari games \cite{mnih2013playing}, or control complex robots \cite{kober2013reinforcement}.
Formally, RL is a sequential decision process where an agent learns optimal policy through interaction with the environment  \cite{Sutton1998ReinforcementIntroduction}, illustrated in figure \ref{fig:rlfig}.
After each action, the agent observes the environment and learns a correlation between an action and its impact on the environment. 
The agent receives a reward from the environment that indicates how good the action was, and uses it to learn optimal actions. 

The goal of an agent is to maximise the expected sum of future rewards, that is done by training over many steps (state-action transitions), and choosing an action that returns the maximum discounted expected reward function.
Sutton et al. \cite{Sutton1998ReinforcementIntroduction} proved that given enough training samples, maximising over expected reward function yields an optimal policy.
A powerful property of RL algorithms is their ability to explore the environment.
At each step, the agent decides whether to exploit (reuse) the action that it knows has the highest expected reward, or to explore other possible actions.
The exploration-exploitation trade-off is what makes RL flexible and able to handle new data that it has not seen during the training phase.
 
RL algorithms fall into two categories, model-based and model-free. 
In a model-free, the agent has no previous knowledge of the system and performs actions in the real environment in order to learn their impact on the environment.
Thus model-free requires no input from the developer, is very flexible, and easy to set-up.
Conversely, model-based algorithms learn a model of the environment, then the agent learns from the interaction with the resulting model.
The model provides many more simulation steps for the agent, allowing it to converge towards the optimal policy faster. 
However, a model-based approach is very challenging to design as the model has to reflect the real environment accurately.
In a computer system where the environment is stochastic and subject to abnormalities all the time, a model-based approach remains an active research field. 
This project uses a model-free approach to demonstrate its feasibility in a real computer system problem.
There are many model-free algorithms available, in the next subsections we discuss two we have used in this work that is also widely used for many applications.

\subsubsection{Deep Q-Network (DQN)}\label{agent:dqn}
Q-learning \cite{watkins1992q} is a reinforcement learning algorithm that learns optimal discrete actions for an environment.
It works by learning an action-value function $Q(s, a)$ that takes in an action $a$ and a state $s$ and outputs a value of how good the action in that state is based on the reward from environment.
The agent collects the Q-value of all possible state and action pairs in a dynamic programming approach, and then it navigates the environment by choosing the action that maximises the Q-value in a given state. 
This approach worked early-on for small games without many state and action pairs.
However, for complicated tasks such as large scale computer systems storing all possible states is not feasible due to the high memory consumption of the dynamic programming approach.

The recent success in combining RL with deep learning \cite{Arulkumaran2017DeepSurvey} showed that it is possible to replace the state-action map with a neural network approximator.
As such, RL had numerous successes in Atari games \cite{mnih2013playing, Silver} that were previously computationally prohibitive. 
The deep learning model approximates the state-action results without having to store every pair in memory. 
The model learns feature weights from agent action-reward observation and updates the neural network weights through back-propagation.

Combining Q-learning with deep learning motivated DQN \cite{van2016deep}. 
DQN is a widely used general-purpose an off-policy algorithm that can learn from data without interacting with the real environment.
An off-policy algorithm is attractive in computer system problems as it enables easier scaling and decoupling of processes.
Follow-up work \cite{mnih2015human} introduced additional enhancements to the standard DQN, which includes a dual-network and using a replay buffer, these helped to make the algorithm robust by delaying the effect of an update on the network causing to be less receptive to divergences.
Furthermore, it allows the agent to warm up by first learning from logs and demonstration such as in DQfD \cite{hester2018deep}.
For these reasons, we utilise DQN in RLCache.

\subsubsection{Soft Actor Critic (SAC)}\label{agent:sac}
\begin{figure}[!htb]
	\centering
	\includegraphics[width=1.0\linewidth]{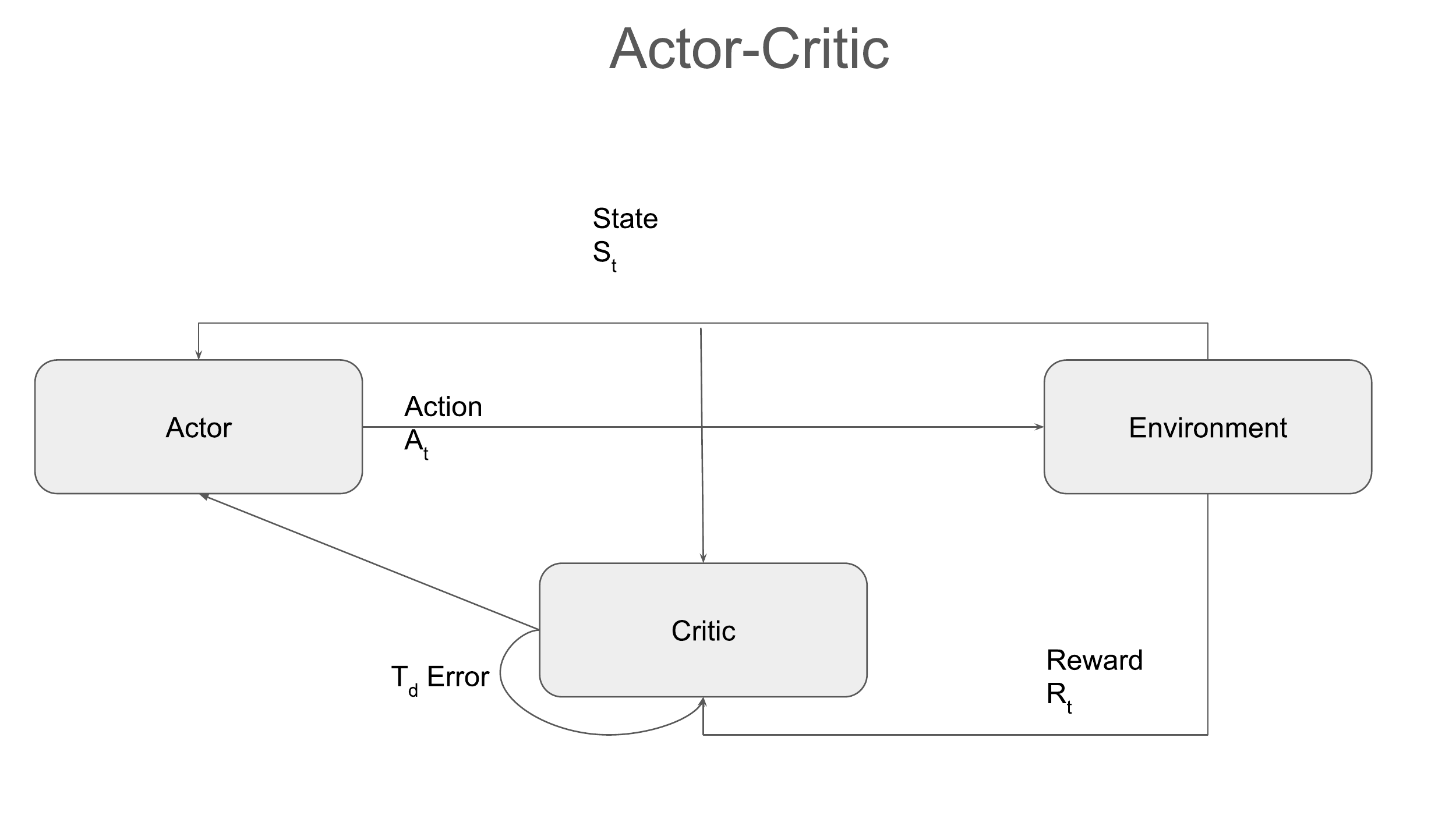}
	\caption[Simplified actor-critic architecture.]{The interaction between an actor and critic in an environment.}
	\label{fig:actorfig}
\end{figure}

Another recently emerged algorithm is SAC \cite{haarnoja2018soft, haarnoja2018softApplications}.
The actor-critic algorithms learn both a policy (actor) and a value (critic) function.
The policy function approximates the best action in a current state, similar to DQN.
The value function represents how good is a given state, by estimating the total reward that is possible for the agent to get starting from this state.
\autoref{fig:actorfig} shows the actor learning the policy function based on the feedback from the critic, while the critic is learning the value directly from the environment reward function\cite{Sutton1998ReinforcementIntroduction}. 

SAC is a robust off-policy algorithm that learns continuous actions in an environment.
It tackles the exploration problem differently than most RL algorithms. 
Typical RL algorithms aim to maximise the reward, while SAC maximises both the entropy of policy and the reward.
The entropy refers to the noise a random variable introduces, in this case, the action value, because SAC works for continuous actions the action value variable takes a real-number value.
Low entropy results in SAC assigning the same value to the variable, while a high entropy encourages exploring other values.
This way SAC is always exploring, and it is never stuck into local maxima.
We omit the maths involved as it is not relevant and recommend the reader to consult the original paper \cite{haarnoja2018soft} for more details.
We used an open source implementation of SAC and DQN provided through the RLGraph framework \cite{Schaarschmidt2018rlgraph}.

\subsection{Reinforcement Learning in Computer Systems}
Imagine a large scale system capable of self-optimised its resources in a way that best fit the SLOs.
For example, this system can be a given CPU utilisation targets and set of possible actions, and it is tasked to maximise the CPU utilisation by assigning tasks to the system. 
Such a system requires less maintenance, and allow non-system experts to achieve optimal performance without extensive knowledge of underlying architecture.
That motivates the reinforcement learning in computer system research field. 

RL was notoriously tricky to apply to computer systems problems due to the large configurations and states a system observes rendering RL algorithms inefficient and incapable of learning in those settings.
The introduction of deep learning into RL renewed the interest of the systems research in employing RL in key research areas of computer systems. 
For example, Mao et al. \cite{mao2016resource} used RL to learn optimal scheduling strategy of tasks in a large scale cluster, other work in the system includes optimising indexing in a database  \cite{Sharma2018}, and optimising distributed stream processor \cite{Li2018}.   
While RL has an enormous potential in computer systems, there are obstacles that make it challenging to integrate RL into systems.

\subsubsection{Delayed State Transition}
Typical RL algorithms perform an action and observe a state transition to the next state immediately. 
For example, in an Atari game, an agent choosing a jump action, the system can immediately observe the jump and its effect on the environment.
Contrarily, computer systems observe the changes several minutes after performing an action. 
We tackle this issue in this work by introducing a framework to abstract this (\autoref{sec:observer}).

\subsubsection{Sample Inefficiency}
RL algorithms are notoriously samples inefficient, i.e. they require many training samples and observations before they learn optimal strategies for the system.
Large quantities of training data are difficult to obtain in non-production system. 
The current approach is to use a simulator to generate near realistic workload and observe it in a pre-production environment until the model learns how to handle that type of data.
The simulator generates workload from a real production system trace or a collection of open-source benchmarks. 
We describe in detail our approach to building a simulator in \autoref{sec:simulator}.

\section{Related work}

\textbf{Eviction strategies.} Survey by Podlipnig et al. \cite{podlipnig2003survey} introduces LRU, LFU, and FIFO argues that LRU is the most optimal strategy that works for the majority of caches.
We evaluated the RL approach against the traditional strategies in \autoref{sec:eval_eviction}. 
Results show that RLCache is capable of learning caching strategy in a changing in the workload better than the traditional approaches.

\textbf{Caching strategies.} 
Alireza et al. \cite{adaptiveCaching} used RL to create an adaptive caching policy using DQN for network objects caching.
The policy learnt the distribution of popular contents and achieved near optimal results.
Their work is similar to RLCache caching strategy described in \autoref{section:rlcache_strategy}.
However, the strategy we describe generalises to other problem outside the network caching domain and can learn result-sets, object keys, and other information.
Authors in \cite{ali2012intelligent} explored a machine learning approach to predict the demand of an object before caching it or replacing it.
Their results outperformed LRU in replacing objects and improved the hit-rate of the system on a various capacity scale. 
Unfortunately, we were not able to obtain an artefact of their code to reproduce their results and compare it to ours for the workload described in \autoref{sec:simulator}.
A working supervised learning approach provides a reasonable motivation for a RL algorithm, as the RL agent will learn the same policy the supervised learning did.
While the supervised learning approach will fail if exposed to different traffic pattern and data than what it observed during training phase. 
The RL approach generalises to new data that it did not observe during the training phase.

\textbf{TTL estimation}. 
We drew inspiration from Schaarschmidt et al. \cite{schaarschmidt2016learning} work in estimating TTL.
They used a variant of DQN for continuous action space, namely normalised advantage function \cite{gu2016continuous} to estimate CDN objects' TTL, they optimised for the result set of a cached key.
Their algorithm's reward function maximised the number of cached objects if the cache is empty.
In our work, we used a soft-actor critic algorithm (SAC), SAC offers a better sampling efficiency, robustness, and is better suited for continuous problems \cite{haarnoja2018softApplications}, details of the algorithm is described previously in \autoref{agent:sac}.
Furthermore, their work optimised the cache query results, while our work focuses on the cache keys its unique patterns while minimising the capacity needed.

\textbf{Delayed experience}. 
The work in \cite{schaarschmidt2016learning} proposed a delayed experience injection algorithm to calculate rewards for a CDN objects TTL estimation. 
The algorithm stored an observed state for exactly estimated-TTL in a queue.
At each time-step, a thread consumes an entry from the queue and observe the difference between the stored state and the current environment.
We extend that concept further and incorporate it into a framework that generalises to other problems.
The framework incorporates computer system termination signals and SLO metrics.
Additionally, it scales to a large number of agents learning concurrently.
We discuss the design in-details in \autoref{sec:observer_timing}.

\textbf{RL in computer systems}. 
We draw inspiration from \cite{mao2016resource} where they used RL for resource management, to improve the sampling efficiency of their algorithm they pause the simulation and output multiple actions per-step, we utilise the same technique in the RLCache eviction strategy described in \autoref{section:rl_eviction} and output multiple keys to evict per eviction step.
Other related work in computer systems \cite{DutreilhUsingWorkflow, tesauro2006hybrid, tesauro2005utility} experimented with replacing the handcrafted threshold policies that maintain SLOs with machine learning based one.
Under the same principle, we replace the cache hit rate and utilisation SLOs with RL enhanced one.
While \cite{Sharma2018, lift2018} looked into optimising key-value stores and databases index selection using RL, and caches are a high-performance type of key-value stores, they do not use indexes thus our work is in a separate domain.

\textbf{Remarks.}
This work distinguishes itself from the aforementioned respected works by developing a framework that incorporates various key contributions and generalises them, and fully automate the CM by designing agents capable of performing these tasks.

\chapter{RLCache: Intelligent Caching}\label{chapter:cache_manager}
\graphicspath{{03-cache-manager/figs/}}
\section{Overview}
\begin{figure}[!htb]
	\centering
	\includegraphics[width=1\linewidth]{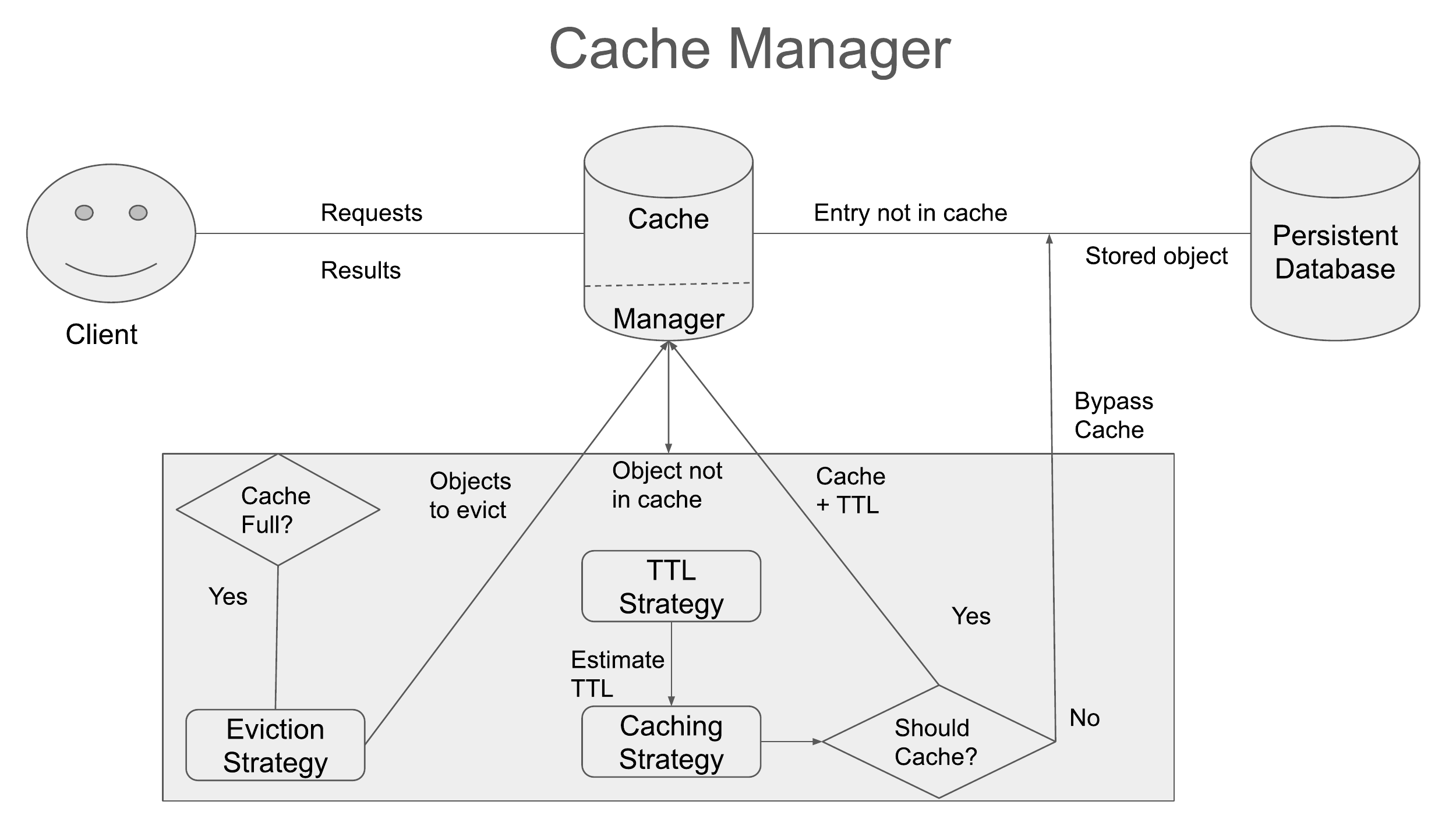}
	\caption[Cache manager workflow.]{The cache manager workflow and various components in the system.}
	\label{fig:cachemanagerfig}
\end{figure}

Caches are commonly used in computer systems to improve the performance of the overall system, they are added at various computation stacks and as such are a primary target for optimisation.
A common use-case for a cache is to improve the performance of a large-scale database.
Large-scale databases support a diverse range of workloads. 
Thus, caches need a cache manager (CM) to ensure the caching decisions provides the best performances trade-offs.
A CM decides whether an object should be cached, the duration it should be cached, and if the cache is full what objects to remove, illustrated in figure \ref{fig:cachemanagerfig}.

For example, BigTable \cite{Chang2008} supports the combination of both batch jobs and adhoc queries. 
Applying a caching layer on it helps with improving scalability and performance.
However, because of the diverse nature of workloads, the cache needs to be finely tuned to reach optimal performance.
That provides an opportunity to apply RL to the caching layer and dynamically learn ideal actions based on the traffic patterns.
A smart CM can improve the cache hit-rate by not caching batch-job tasks which minimises the memory size required, thus results in reduced financial costs and increases in performance.
Additionally, it learns keys that have the least impact on the system if they were evicted, resulting in reduced cache misses stemming from a poor eviction decision.

The rest of this chapter discusses applying RL to CM's components starting with introducing the observer framework that is responsible for orchestrating the communication between the CM and the RL algorithms.
\section{RLObserver: Framework for delayed experience observations and modelling}\label{sec:observer}
\subsection{RLObserver Architecture}
Computer systems, as can be seen in the cache management problem, made of multiple components that each make a decision that contributes to the overall stability of the system. 
These decisions impact is only observed much later in the process.
For example in the caching problem, the TTL estimation and caching strategy are two independent processes that contribute to the overall hit-rate of the cache, the CM can only observe the impact of their decision after the TTL is up or if the entry receives an invalidation update.
A core contribution of this work is introducing a framework for delayed experience observations that enables multiple components to independently make intelligent decisions. 

\begin{figure}[!htb]
	\centering
	\includegraphics[width=1.0\linewidth]{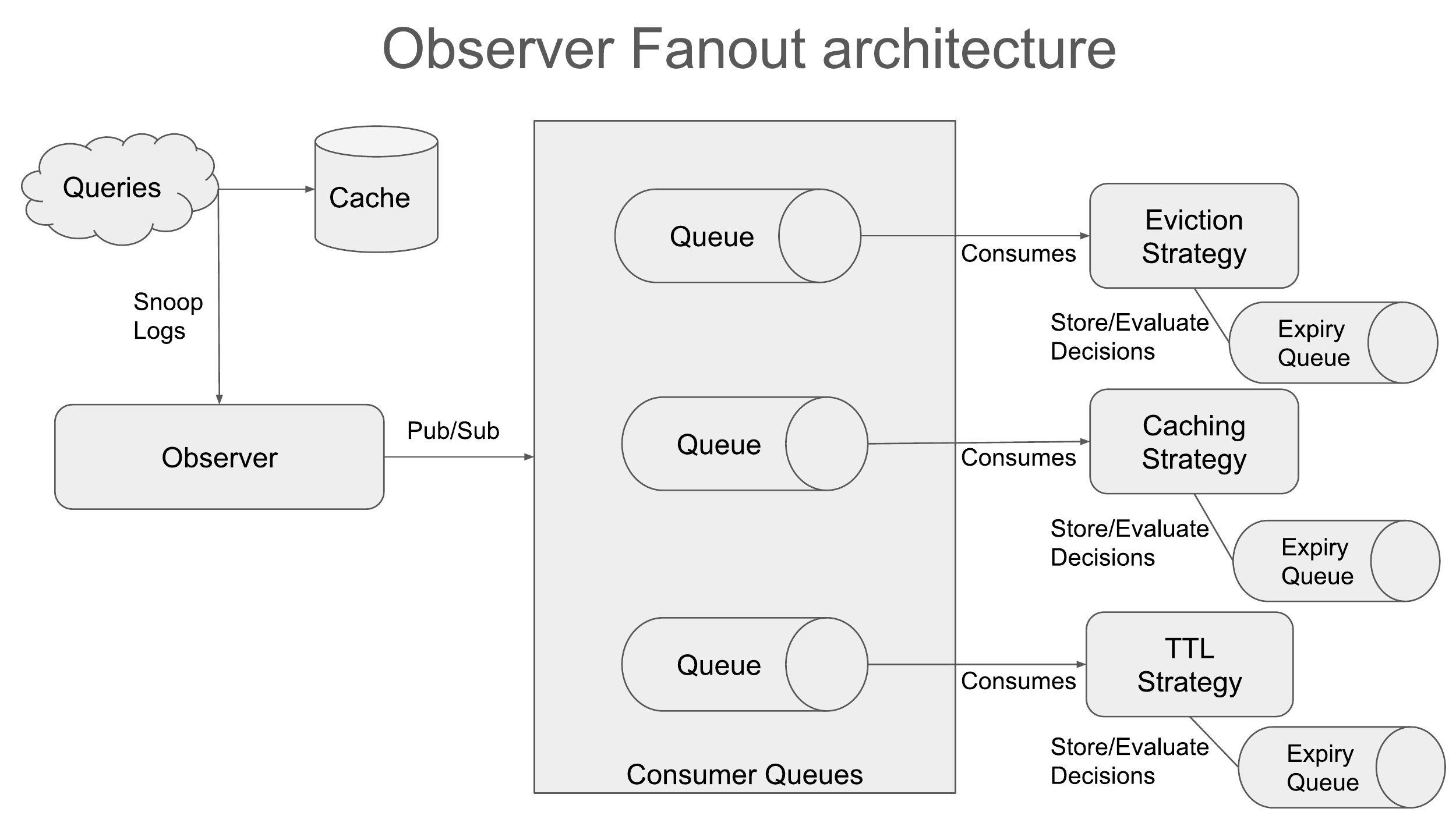}
	\caption[Observer architecture.]{The workflow of the observer architecture and its fan-out design.}
	\label{fig:observerfig}
\end{figure}

Figure \ref{fig:observerfig} shows the observer architecture, the observer distributes the system metrics observations to interested components in a fan-out fashion.
Fan-out architecture \cite{chen2011scaling} allows the system to distribute the observation to multiple components without blocking the processor for responses, enabling it to scale for a large number of agents.
By decoupling the observer logic from strategies and providing a flexible mechanism to register interest in various system metrics, it allows the system to scale as components can be added and removed without affecting each other.
When a CM receives a read request, if the item in the cache it would signal a cache hit observation to the observer, if not in the cache it would send a cache miss. 
On an invalidate request (change or addition of objects) the CM sends an invalidation signal to let the components know the value they hold is no longer considered a fresh copy.
Additionally, the framework allows the components to define a callback function for when an entry expires after TTL seconds.
This procedure allows the component to learn from decisions that had no positive or negative impact on the system. 

\subsection{Observer Delayed experience}\label{sec:observer_timing}
When a system makes a decision, the environment reports the impact of the decision much later.
Therefore, the system needs to store incomplete transactions into an expiring queue.
When a caching strategy makes a decision it enqueues the decision and a snapshot of the state into the queue.
If the CM receives non-terminal updates (i.e. cache hits), it retrieves a mutable state of the stored experience and updates the hit count.
Otherwise, if the update invalidates what is in the cache, the stored experience is then considered completed and updated to reflect the reason for termination, and then the components invoke the termination procedure.
If the CM does not observe any more interaction with the object for the duration of the maximum configurable TTL seconds, the queue will invoke an eviction procedure.
The eviction procedure is a customs procedure defined by the subscriber components, and it handles the workflow of when there is no more observation related to the (action, state) pair in the given observable time.

The expiry queue is a dictionary from key to stored object to provide a fast $O(1)$ access to the object, and a min heap to keep track of objects that are next going to expire. 
On a separate thread, the expiry heap is monitored every second for keys that have expired, when a key is expired, an observation is enqueued in the observer with expiration as a reason and key, value, and the cache metadata so the CM strategies can handle this later.
\section{RLCache: Caching Strategy}\label{section:rlcache_strategy}
Caching strategy decides if an object after a read or write should be cached or not.
An optimal caching strategy aims to maximise the cache hit rate while minimising the number of items occupying the cache by predicting if an object is going to be requested again in a short time-frame.

Since real-life workloads vary, a binary caching strategy such as the one described in \autoref{sec:caching_decision} will not be able to capture unique object access patterns in different workloads. 
Take for example Twitter, where some users have a large number of followers who reads the user posts as soon as post them, and some users who produce a large number of tweets but have a fewer number of followers who do not read their posts.
An intelligent cache should learn the patterns of those users and base the caching decision accordingly.

Needlessly caching objects increase the network latency as these objects are written to both the cache and persistent backend, in a write-heavy workload this is not desirable.
Additionally, a binary decision to caching all the content is challenging due to capacity constraints, especially on constrained devices like smartphones, and even dedicated hosted caches have a limited capacity compared to the persistent backend.
Therefore, a RL based caching strategy is desirable, and the rest of this section discusses the RL model, algorithm, and reward.
\subsubsection{Agent Design}
\begin{figure}[!htb]
	\centering
	\includegraphics[width=1\linewidth]{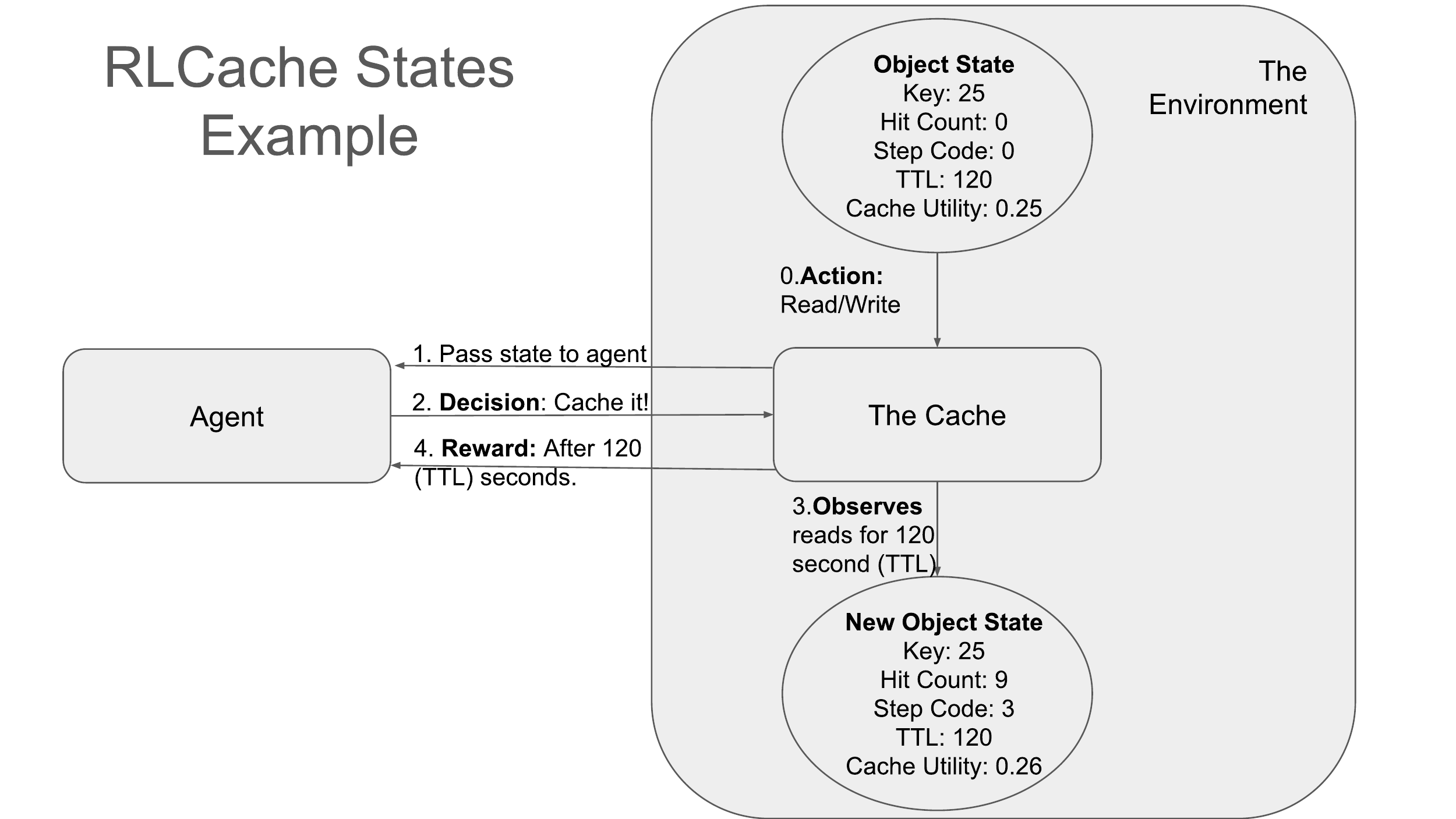}
	\caption[Illustration of RLCache environment's cached state.]{The state representation of an object as observed by the agent.}
	\label{fig:rlcachestatefig}
\end{figure}

The CM has no explicit end-state as the system operates indefinitely or until it crashes.
Therefore, RLCache agents designs observe a reward after every step or update without any terminal signal.
Each cached object is an independent state in the Markov decision process (MDP) \cite{howard1960dynamic}.
The state transition represents the transition of the cached object from the initial state without any cache hits, to the final state that includes termination reason as illustrated by \autoref{fig:rlcachestatefig}.
The final state for each object is when the object is invalidated, evicted, or expired from the cache until then the cache state circulate under the hit state.

We justify the continuous system design by experimenting with an episodic design, where we introduced an arbitrary terminal state (after every $10,000$ operations) for the system to observe an update and reset its internal states. 
However, the results were underwhelming in comparison to the continuous system as the agent was not able to adapt quickly when the workload changes, this is visible when the workload switches to write-heavy and as such results in a large number of invalidation causing the termination signal to be reached much quicker.
Additionally, real-world CM has no explicit termination signal and it operates until failure. 

\subsubsection{State}\label{section:rlcache_strategy_state}
Learning the importance of an object and its behaviour requires knowledge of its unique identifier (the key), the access pattern, and the result set.
Thus, the agent state consist of a continuous space vector resulted by concatenating: \textit{object key}, \textit{operation type}, \textit{results set}, \textit{object metadata}, encoded \textit{termination reason}, and the \textit{hit count}.
\begin{itemize}
	\item \textit{Object key}: For the agent to learn important (hot) keys, the state needs to encode the key's unique representation.
	To transform the object key from the String representation to a number before feeding it to the tensor, RLCache maintains a dictionary of observed keys to their encoding as is usually done in natural language processing tasks.
	\item \textit{Operation type}: The operation type encodes the invocation reason of the caching strategy.
	The operation type distinguishes between an object queried from a Get request or from a Set to help the agent learn when the workload is a write-heavy or read-heavy workload.
	\item \textit{Results set}: Results set are the values stored in the persistent database associated with the object, passed into an embedding layer as commonly used in natural language translation tasks, to produce a vector of continuous space. 
	The results set being part of the state helps the agent learn properties about the requested object.
	For example, the cache learns an association between '.jpg' in the results set and being a hot item.
	\item \textit{Object metadata}: The metadata contains a vector of result set size, TTL, and query retrieval times, guide the agent in learning to cache results that tend to be slower to retrieve and maximise the utilisation of available cache space. 
	\item \textit{Termination reason}: The termination reason helps the agent distinguish between an object that reached the end of its life cycle because of cache expiry, eviction, invalidation, or because of a miss. 
 	Thus the agent can learn the patterns of objects that are susceptible to many invalidation.
	\item \textit{Hit count}: Contains the number of hits the object observed until it reached its termination state. 
	Therefore, the agent able to learn hot objects that received more hits are favourable than other objects.
\end{itemize}

This state representation provides the necessary context to learn individual object behaviour.
Keys that have high hit rate tend to be heavy-read keys, keys that are invalidated frequently are heavy-write, adding result set metadata allows the agent to maximise the utilisation of the cache capacity. 
\subsubsection{Action}
Before a CM caches an object, it queries the caching strategy agent for a decision, whether to cache it or not. 
The caching decision is a discrete binary problem, should cache ($1$) or should not cache ($0$).
\subsubsection{Reward}
An optimal caching strategy aims to maximise the cache hit rate while minimising needless caching operations. 
The agent capture that objective using its reward function.
The reward function described in \ref{algo:caching_strategy_reward} aims to reward the agent based on its caching decision.
\begin{itemize}
	\item The reward scales with the number of read requests to the object (hits), so the agent has the incentive to cache hot objects. 
    If the agent mispredicts that a cache read will follow, the agent is punished, thus discouraging the agent from caching entries that tend to be invalidated or not read until they are expired. 
	\item The reward rewards the cases where the agent does not cache an object and the object is not requested during its TTL lifespan to motivate the agent to avoid caching heavy-write and unpopular objects.
	The reward encodes a punishment for when the agent does mispredict.
\end{itemize}
Additional optional scaling functions can be used on top of the reward function to reflect the nature of the backend.
Scaling the reward function by the time it takes to retrieve the item from the backend allows the agent to favour caching items that tend to be slower to retrieve.
Scaling the reward function by the ratio of cache hits to the size of the result set discourages caching items that tend to be large in size but are not requested often, minimising the storage capacity needed to achieve the same hit-rate.
\begin{algorithm}[!htb]
	\caption{RLCache: Caching Strategy Reward Function}\label{algo:caching_strategy_reward}
	\begin{algorithmic}
		\Procedure{reward($action, observation$)}{}
		\If{$action == should\_cache$}
		\State $reward = hit\_count$
		\Else \Comment {Did not cache entry}
		\If{$observation == cache\_miss$}
		\State $reward = -1$
		\Else
		\State $reward = 1$
		\EndIf
		\EndIf
		\EndProcedure
	\end{algorithmic}
\end{algorithm}

\subsubsection{Algorithm}
Algorithm \ref{algo:caching_strategy} shows the caching strategy. 
The agent uses an off-policy algorithm to learn from the interactions.
The advantage of using an off-policy algorithm is the ability to learn without interacting with the real environment.
The property to learn from data directly simplifies the process of scaling the caching strategy, as discussed in \autoref{section:distributed_cache_manager}.

The caching strategy operates using two procedures.
The first is the decision making one, \textit{Should Cache}.
This procedure responsible to decide whether to cache an object or not, the output is a boolean to indicate if it should be cached.
The procedure utilises the RL agent to make its decision, before feeding the input to the agent it needs to be tensorised.
The key and value (result set) are converted to an integer representation using the embedding layer. 
After converting the object into its state representation, the agent consumes the state vector representation and pass it to its network.
The network outputs a Q-value that predicts how good an action is given current state.
To encourage exploration, the agent chooses the action that maximises the Q-value with probability $1 - \epsilon$, otherwise, it uniformly samples a random action. 
Formally, the epsilon is used to indicate the probability of choosing a random action uniformly instead. 
The epsilon starts at $1$, exploring random actions aggressively at the beginning, but linearly decays until it reaches a $0.1$.
This configuration allows the agent to choose the value the network is most confident while leaving room for exploration that avoids the agent being stuck in a local maximum when the traffic pattern shifts. 
When it chooses an action, the observer framework creates an incomplete experience and stores it in an expiring queue, as discussed in \autoref{sec:observer_timing} as the environment cannot provide the reward directly after but only in the future.
The incomplete experience contains an immutable original state and the agent's action. 

The second procedure is a non-blocking \textit{observe} whose purpose is to update the agent's network weights by rewarding (or punishing) the agent.
It is invoked by the observer (see \autoref{sec:observer}) to update the state associated with the object key.
When the CM observes an action to an object that it made a decision on previously, it retrieves the stored state, creates a new transition to the new state while keeping the original state intact.
If the new state is a cache hit, the cache hit count associated with that key increases. 
Otherwise, the environment calculates the reward using the reward function described above and passes it to the agent.
The agent updates the network weight through back-propagation based on how good the action was.
When an entry hits a final state, it completes its MDP, and the observer removes it from the observed queue.

\begin{algorithm}[!htb]
	\caption{RLCache: Caching Strategy.}\label{algo:caching_strategy}
	\begin{algorithmic}
		\State \textbf{Given:} \begin{itemize}
			\item an off-policy RL algorithm $A$.	\Comment{e.g. \textbf{DQN}}
			\item exploration decaying steps $Es$ 
		\end{itemize}
		\State Initialise an empty replay memory $R \gets \phi$ 
		\State Initialise an empty expiry observer queue $Q \gets \phi$
		\State Initialise $A$ with random weights
		\State Initialise $\epsilon \gets 1.0$ an exploration probability that linearly decays until $0.05$ for $Es$ steps.

		\While{\textit{CacheManager} is alive} \Comment{Continuous System}
			\Procedure{Should Cache}{$key, values, TTL, operation\_type$}
				\State $s \gets  convertToState(key, value, TTL, operation\_type)$
				\State Select action $\alpha = A(s)$ with exploration probability $\epsilon$
				\State Create incomplete experience $i_e \gets (s, \alpha, time, operation\_type)$
				\State Insert incomplete experience into $Q[key] = i_e$ for $TTL$ seconds.
				\State return the action $\alpha$
			\EndProcedure

			\Procedure{\textbf{async }Observe}{$key,observation\_type$}
				\If{$observation\_type == cache\_hit$}
					\State $stored\_state = Q[key]$
					\State increment $stored\_state$ hit rate
				\Else \Comment{Cache miss, invalidate, or $Q$ entry expiry}
					\State $stored\_experience \gets Q[key]$
					\State Compute reward $r = reward(stored\_experience)$
					\State insert complete experience into $R$
					\State delete entry associated with the key in $Q$
					\State Calculate the loss and update $A$ network's weight
				\EndIf
			\EndProcedure
		\EndWhile 
	\end{algorithmic}
\end{algorithm}

\section{RLCache: Eviction Strategy}\label{section:rl_eviction}
Eviction strategy governs the removal of the objects from the cache process once the cache is full.
Optimal eviction strategy removes objects that are no longer used or has the least harmful impact on the SLO.
The traditional eviction policy will not be able to handle a case of mixed access pattern without relying on handwritten rules, here comes this work.
RLCache learns information about the key, result set, and other system information and derives a bespoke policy for each object.

To motivate using RL for eviction strategy, imagine a partitioned database where some keys are stored on faster (or under-loaded) servers, while others are on slower ones.
When the cache exceeded its maximum capacity, and an eviction must happen before caching any new objects.
Traditional algorithms mentioned in \autoref{background:eviction} will evict objects without taking into account the overall impact on the system as they only have simple information about the cache, a problem RLCache does not suffer from. 

\subsubsection{Agent Design}
\begin{algorithm}[!htb]
	\caption{RLCache: Eviction Strategy}\label{algo:eviction_strategy}
	\begin{algorithmic}
		\State \textbf{Given:} \begin{itemize}
			\item an off-policy RL algorithm $A$.	\Comment{e.g. \textbf{DQN}}
			\item exploration decaying steps $Es$ 
		\end{itemize}
		\State Initialise an empty replay memory $R \gets \phi$ 
		\State Initialise an empty expiry action queue $Q \gets \phi$
		\State Initialise an empty dictionary to hold observed stats $D \gets \phi$
		\State Initialise $A$ with random weights
		\State Initialise $\epsilon \gets 1.0$ an exploration probability that linearly decays until $0.2$ for $Es$ steps.

		\While{\textit{CacheManager} is alive} \Comment{Continuous System}
			\Procedure{Trim Cache}{$cache$}
				\State Initialise empty list of keys to evict $L \gets \phi$
				\For{$key$ in $cache$} 
					\State Retrieve from $Q$ the state $s$ associated with the key.
					\State Select action $\alpha = A(s)$ with exploration probability $\epsilon$
					\If{$\alpha == 1$} \Comment{evict}
						\State Append the $key$ to $L$ 
					\EndIf 
					\State Create incomplete experience $i_e \gets (s, \alpha, time, operation Type)$ 
					\State Insert incomplete experience into $Q[key] = i_e$ for $TTL$ seconds.
				\EndFor
				\State \Return return the list of keys $L$
			\EndProcedure

			\Procedure{\textbf{async }Observe}{$key, observation\_type$}
				\If{$observation\_type == write $} \Comment{New write to cache}
					\State $s \gets convertToState(key,value,TTL,operation\_type)$
					\State Insert incomplete experience into $Q$ for $TTL$ seconds.
				\ElsIf{observation Type $=$ cache Hit}
					\State $stored\_state = Q[key]$ 
					\State increment $stored\_state$ cache hit rate
				\Else \Comment{Cache miss, invalidate, or $Q$ entry expiry}
					\State $stored\_experience \gets Q[key]$ 
					\State Compute reward $r = reward(stored\_experience)$
					\State insert complete experience into $R$
					\State delete entry associated with the key in $Q$
					\State Calculate the loss and update $A$ network's weight
				\EndIf
			\EndProcedure
		\EndWhile 
	\end{algorithmic}
\end{algorithm}

Much like the caching strategy agent described in \autoref{section:rlcache_strategy}, the state design mirrors the one used in the caching strategy.
However, the algorithm, action and reward differ.
The details of the RLCache eviction's algorithm are illustrated in listing \ref{algo:eviction_strategy}, while we discuss the rest of the model below.

\subsubsection{State}
We have experimented with two different state design.
\begin{enumerate}
	\item \textbf{Single-state cache:} Models the whole cache as a single entity to allow the agent to learn relative importance between keys.
	For example, the agent learns to compare keys based on their size and hit-rate.
	This approach does not scale when the cache size increases, as each state will have a copy of the whole cache, and each transition as well.
	The state grows exponentially, making it very difficult for the agent network to learn any distinctive features.
	\item \textbf{Independent objects state:} Treats each cached object as a separate state, and feed a single object at a time into the agent, in a similar manner to the caching strategy (\autoref{section:rlcache_strategy}).
	This approach yielded the best result, as the agent objective is to minimise cache misses and evict keys that are unlikely to be read again, regardless of how they compare to each other. 
\end{enumerate}
The rest of this work considers only the second approach.

\subsubsection{Action}
When the cache is full, RLCache tasks the eviction strategy to nominate object(s) to evict from the cache and make space for new entries.
The eviction decision is a discrete binary problem, should evict the object ($1$) or should not ($0$).
Initially, we would sample uniformly from the cache until we find an object to evict, thus mimicking the traditional eviction algorithm of one eviction per call. 
However, this approach yields less training samples for the agent to learn from and overall slowed the system significantly. 
Instead, taking the idea proposed in \cite{mao2016resource}, RLCache pauses the execution of the cache, scans every cached item and outputs an eviction decision on it, at the end of the procedure RLCache evicts all entries that the strategy chose.
Consequently, the eviction strategy agent receives many updates to learn from.

\subsubsection{Reward}
\begin{algorithm}[!htb]
	\caption{RLCache: Eviction Strategy Reward Function}\label{algo:eviction_strategy_reward}
	\begin{algorithmic}
		\Procedure{reward($stored\_state, action, observation$)}{}
		\If{$observation == expiry$} \Comment{expiry of TTL after the agent decision}
			\If{$action == 1$} \Comment{If should evict} 
				\State $reward = 1$ \Comment{object was evicted and not read after.}
			\Else \Comment{Object was not evicted}
				\State $reward = hit\_count_{observation\_time} -  hit\_count_{decision\_time}$
			\EndIf
		\EndIf
		\If{$observation = invalidate$}
			\If{$action == 1$}  \Comment{If should evict} 
				\State $reward = 1$ \Comment{eviction followed with an invalidation.}
			 \Else
			 	\State $reward = -1$ \Comment{no eviction but invalidated after.}
			\EndIf
		\EndIf
		\If{$observation == cache\_miss$}
			\State $reward = -1$			
		\EndIf
		\State \Return $reward$
		\EndProcedure
	\end{algorithmic}
\end{algorithm}

Optimal eviction strategy minimises the cache misses caused by premature evictions.
The reward function described in listing \ref{algo:eviction_strategy_reward} punishes the strategy for evicting an object that caused a cache miss.
Eviction strategy, mirrors the caching strategy in both the objective function and the action it takes, there are two actions an eviction strategy can do:
\begin{itemize}
	\item Evicts an object: the environment rewards the agent if the evicted object is later invalidated, and will punish it if it caused a cache miss.
	\item The environment rewards the agent for not evicting an object that the client reads later. 
	Thus, motivating the agent to hold on useful objects. 
	However, if the object is later invalidated the environment issues a punishment to the agent.
\end{itemize}

\section{RLCache: TTL Estimation Strategy}\label{sec:ttl_agent}
Adding a cache expiration time to objects avoids cluttering the cache and ensures data is always fresh.
The TTL strategy needs to estimate a TTL that expires as soon as the object is no longer expected to be read again or before it is invalidated.
Doing so maximises the cache hit rate and minimises over-populating the cache with objects that are no longer useful.
In a specialised type of caches such as CDNs, TTL estimation is very crucial for the overall performance of the system. 
High TTL overpopulate the cache by keeping objects longer in it, while low TTL results in fewer cache-hits.
Furthermore, there is a significant performance impact for sending manual invalidation of an object if still exists in the cache as it requires to send invalidation to all subscribed clients, causing a network latency overhead. 
The traditional approach of using fixed TTL values cannot optimise for changing content and changing traffic patterns.
While RLCache learns optimal TTL parameter from observing the access patterns and the content of cached objects and estimates the optimal TTL that reduces the cache invalidation and utilises the cache better. 
Dynamically changing TTL is useful even when the traffic pattern is static.
For example, the TTL for an object that contains a website logo should be much higher than the TTL for an object that retrieves posts from a social media platform. 
In the former, the website logo is unlikely to change, hence there is no need to invalidate as often as a social media object which may change on a much faster rate.

\subsubsection{Agent Design}
Like the previous two strategies, TTL estimation treats each object as an independent state rather than the whole cache as one large state.
However, TTL estimation is a continuous problem with a continuous action space, unlike the discrete actions that both eviction and caching strategies use.
As such the agent method differs from the previous two strategies, the method has to support continuous actions and needs to be an off-policy algorithm to facilitate scaling-out, SAC described earlier in \autoref{agent:sac} fulfils both requirements.
\subsubsection{State}
To estimate the TTL of an object, the agent needs to know information about both the object and overall cache.
\begin{itemize}
	\item \textbf{Object information:} Object unique key, the value set, the value set size, hit counts, and termination code, similar to the two other strategies discussed above.
	\item \textbf{Cache information:} Information about the ratio of cached entities to the capacity of the cache (cache utility).
	Cache utility described in \cite{schaarschmidt2016learning} allows the agent to reason with the trade-off between estimating higher TTL when the cache is emptier and when to be conservative with its estimation when it is close to its capacity.
\end{itemize}
\subsubsection{Action}
Estimating the number of seconds the entry should remain in the cache for is a continuous problem.
The agent outputs a number between $[0, \infty)$ to indicate the object's TTL in seconds.
Theoretically, the time is unbounded and the agent should be able to predict any number up to infinity and given infinite number of observations the agent will be able to learn the optimal range for the TTL value.
In practice, an upper bound limit is set that allows the agent to converge faster as it has fewer values to try.
\subsubsection{Reward}
\begin{algorithm}[!htb]
	\caption{RLCache: TTL Estimation Reward Function}\label{algo:ttl_strategy_reward}
	\begin{algorithmic}
		\Procedure{reward}{$t_{inv}, t_{exp}, utility, hits$}
		
		\If{$observation\_type == hit$}
			\State $r = 1$
		\ElsIf{$observation\_type == eviction$} 
			\State $r = 0$
		\Else \Comment{Miss, expire, or invalidate}
			\State $t_{diff} = absolute((t_{exp} + 1) / t_{inv}) * -1$
		\EndIf
		\State \Return $r * utility$
		\EndProcedure
	\end{algorithmic}
\end{algorithm}

Using the difference between invalidation time and estimated time is a straightforward way to calculate how good an estimate is. 
While it works for write-heavy workloads, invalidation are rare in read-heavy workloads like web-based workloads \cite{breslau1999web}.
For a general purpose CM, this means those workloads do not observe reward majority of the time. 
Therefore, the environment rewards the agent for every hit an object receives to encourage it to assign higher TTL for objects that have a higher read ratio.
Objects that do not observe any read will have an overall less reward than objects that do, causing the agent to favour the heavy-read objects.
Adding cache utility to the reward function as did \cite{schaarschmidt2016learning} encourages higher TTL when the cache is empty and a conservative TTL when the cache is close to its capacity, hence the second term of the reward function that scales the reward by the cache utility. 
Combining both reward functions provides a reward for all cached object.
Objects that do not receive invalidation will receive a reward based on their cache hit performance and the cache utility.
As the environment punishes the agent if an object is invalided, the agent is motivated to minimise the TTL value it outputs.
This complex trade-off is captured in the reward function in algorithm \ref{algo:ttl_strategy_reward}.
By punishing entries that receive an invalidation for any reason, the agent learns how the other strategies (such as the eviction one) operates as well and provide support for them through an accurate TTL.
We maintain a decision monitoring queue, to store transitions that were caused by other policies.
Objects in the monitoring queue live for the maximum allowed TTL.
The monitoring queue is required to handle cases when the TTL estimated much lower than actual TTL.

\section{RLCache: Multi-Agents Configuration}\label{sec:multi_agent}
Previous sections showed the design of independent agents that each specialises in a sub-task of the CM, the next natural step is to have all agents operate together in the same environment.
A naive implementation that uses all three agents is very expensive to operate, as each agent will need to learn all possible actions every other agent will make in addition to the task it is optimising.
This section describes our alternative approach that aims to reduce the number of training steps required.
The goal of this agent is to discover what possible bottleneck exists when using multiple agents concurrently in real systems and possibly open new ideas for future work.

\subsubsection{Agent Design}
Taking inspiration from game theory and recent work in cooperative games, Lowe et al. \cite{lowe2017multi} proposed a mechanism to coordinate cooperative agents who share the same goal by using a meta agent that orchestrates the action and assign credit to agents.
They used multiple actor-critic agents, where each agent had its own separate policy.
Their experiments showed promising results in small toy experiments. 
We took ideas from their work and adapted it into this one to apply it to a real-world application.
In their work, they stored the agents' learnt policies between episodes.
At the end of each episode, every agent sampled a policy and acted using it, by swapping the policy constantly, they managed to reduce the invariant between what each agent learn. 
Because the problem domain of the system cache is a continuous non-episodic problem, sampling per episode was not an option, and sampling at set interval is too disruptive to the real workload to be a viable option.
A simpler design that tolerates invariant between agents is better suited for system problems.

\begin{figure}[!htb]
	\centering
	\includegraphics[width=1.0\linewidth]{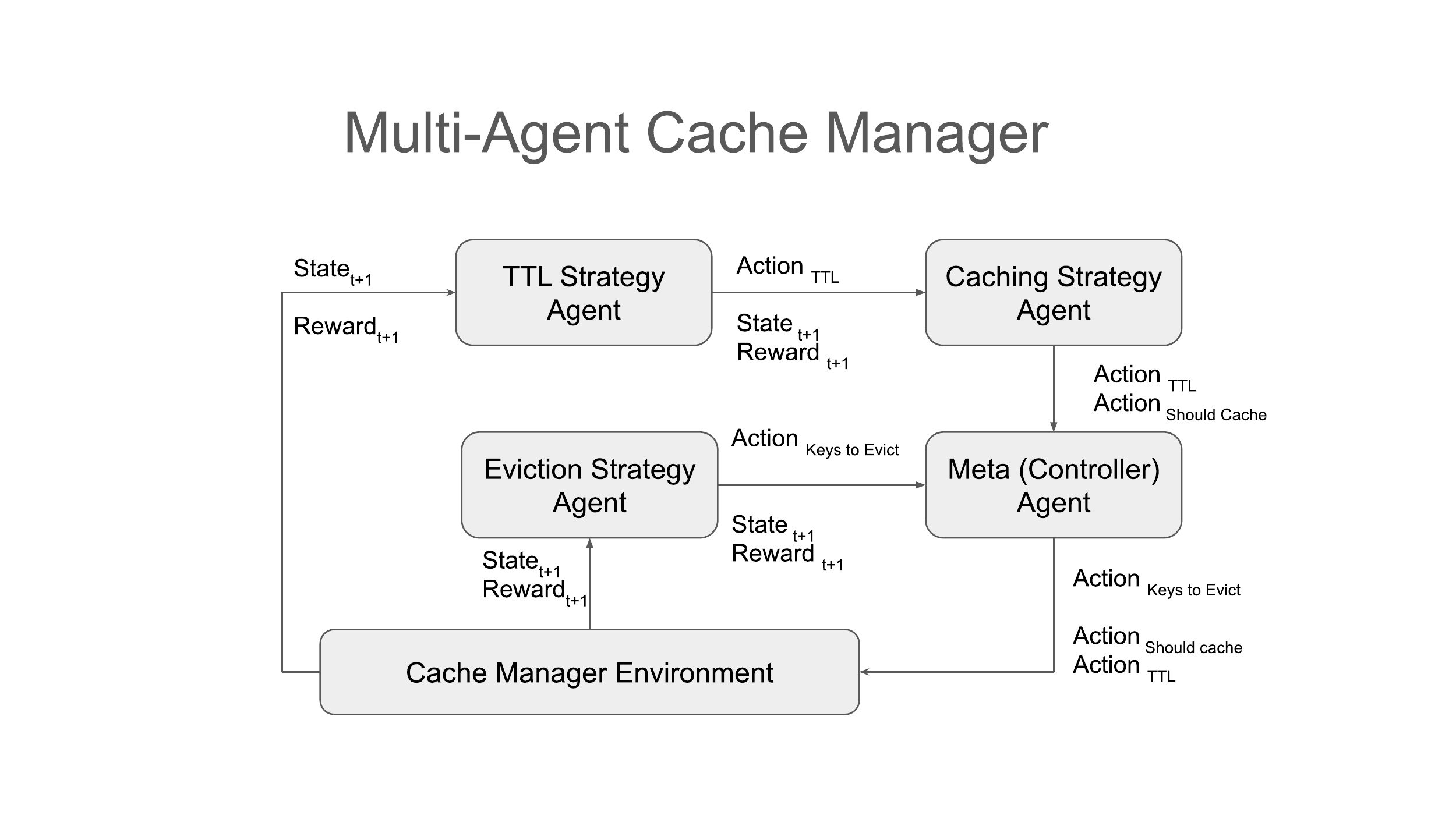}
	\caption{A simplified view of the cache manager's multi-agent setup.}
	\label{fig:multiagentfig}
\end{figure}

RLCache multi-agent architecture uses a separate agent for each task, then a meta (controller) agent to coordinate the output of the action from each agent with the goal that maximises the expected reward for all of the agent. \autoref{fig:multiagentfig} illustrates a simplified design of multi-agent architecture. 
Each agent states' and actions' mirrors the one described above in \autoref{section:rlcache_strategy}, \autoref{sec:ttl_agent}, and \autoref{section:rl_eviction}. 
The observer framework abstracts the multi-agent complexity by sending the state updates to all agents simultaneously. 
The reward function is also isolated, where each of the agent state, action, and reward is as described in their respective subsections above.
However, the meta agent has a separate state, action, and reward function.
The reward function for the meta agent takes into account the sum of all agents reward and aims to maximise that.
The meta-agent state encodes a vector of states of the agents, the action they took, and the reward they received.
A no-op is encoded as a zero in the vector to represent an agent that did not take action, such as the eviction agent only operating when the cache is full.
The meta agent action is a vector of the three actions that will be committed to the environment.

\section{RLCache: Single Agent Multi-Task Configuration}\label{sec:multi_task}
The agents described above had a comparable state representation and learnt common features (e.g. the hot-keys key value), but the previous designs did not allow to easily transfer that information.
They used separate networks to model data in a distribution plane despite using a similar reward signal based on the cache-hit rate.
However, it is not straightforward to merge the networks as they operate at different times and serve different purposes.
For example, the eviction strategy only called when the cache is full.
Therefore, combining all the networks into a single network can improve the learning significantly, also triples the number of actions (i.e. training steps) of a single network.
This approach is known as multi-task learning in the field of deep learning.
Multi-task learning \cite{caruana1997multitask, thrun2012learning} is an approach that builds one network that generalises to multiple tasks.

\subsubsection{Agent Design}
A single agent uses a shared network to decide on the three caching strategies: caching decision, TTL estimation, and key eviction.
Multi-task learning is difficult compared to a single task learning, as there needs to be a balance between the multiple tasks and their weights on the network and the reward the environment gives \cite{hessel2018multi}.
Since the three tasks have different action spaces (TTL continuous, caching and evicting are discrete), modelling these action spaces requires a mathematical justification. 

Tang et al. \cite{tang2019discretizing} identified that discrete action is contained within the continuous action-space.
Therefore, it is possible to use continuous action space to model all three task, and discretise the action for the two discrete tasks (eviction and caching).
Discretising continuous action space was looked at in \cite{van2009using, tang2019discretizing}.
They proposed several methods including approximating to nearest discrete value, approximating and removing some values, or discretising each dimension of the action space into K discrete actions and using the probability distributions of those dimensions to choose the nearest discrete value.
We experimented with the simple approach first of using a function approximator, which rounds a probability between $\pi_i = [0, 1]$ to the nearest integer.
The simple approach worked well due to the small (binary) discrete action space of both the eviction and caching.

We use SAC with continuous action space to allow the agent to learn a set of three actions using the same network. 
We further simplify the network by combining the caching decision and TTL estimation into the same task, if the agent estimates a TTL over a  configurable threshold, the object is cached.

\subsubsection{State}
The state encodes the union of all previously mentioned strategies' states, so the agent has enough information to handle any of its tasks.
Over-provisioning information is not an issue in deep learning, as the agent's network will adjust the weight assigned to the feature if it is not contributing to the overall benefit of the system.

\subsubsection{Action}
The agent outputs two values, one indicates if the object should be evicted, and the other a duration estimate of the TTL.
To simplify the agent decision, both of the actions are in the continuous action space (and indeed SAC only works in that space).
RLCache calls the Estimate TTL procedure when an object is not in the cache, based on the results of the estimation, RLCache decided whether to cache it or not and uses the estimated TTL as the TTL for the object. 
When the cache is full, the multi-agents scans the whole cache outputting a value between $[0, 1]$ of which we later rounded it to map it to the discrete set ${0, 1}$ that indicate an eviction or not, as described in the eviction strategy previously (\autoref{section:rl_eviction}).

\subsubsection{Reward}
Designing a complex reward function is beneficiary when all possible cases are captured, with a hand-crafted reward function the agent is guided towards convergence faster, such as the cases with the single agents described earlier.
However, in a multi-task agent where some of the tasks are contradictory (i.e. eviction and caching decisions), designing a correct reward function that captures all possible interaction is challenging and demolishes the flexible benefit of using RL.
Therefore, we opted for a simplified reward function that rewards each cache hit with $1$, reward a successful prediction of eviction or not caching with $10$, while each cache misses and invalidation with $-10$.
While it takes the agents longer to learn the pattern of the object in this way by doing so the agent also learns the optimal reward function without any hand engineering, and allows the multi-task agent to incorporate other  signals easier.
Algorithm \ref{algo:multi_reward} describes the reward function.
\begin{algorithm}[!htb]
	\caption{RLCache: Multi-Task Reward Function}\label{algo:multi_reward}
	\begin{algorithmic}
		\Procedure{reward}{$t_{inv}, t_{exp}, utility, hits$}
		\If{$observation == hit$}
			\State $r = 1$
		\ElsIf{$observation == invalidate $}
			\If{$action[ttl] < 10 or action[eviction] == 1$}
				\State $r = 10$
			\Else 
				\State $r = -10$
			\EndIf
		\ElsIf{$observation == miss$}
			\State $r = -10$
		\Else \Comment {Expires or evicted}
			\State $r = 0$ 
		\EndIf
		\State \Return $r * utility$
		\EndProcedure
	\end{algorithmic}
\end{algorithm}

\chapter{Evaluation}\label{chapter:evaluation}
\graphicspath{{04-evaluation/figs}}

\section{Evaluation Aims}
This chapter describes the experiments setup, the hyperparameters choice, and the evaluation of several agents.
The goals of the experiments are the following:
\begin{enumerate}
	\item Demonstrate the advantage of using RL in computer systems, namely CM, by optimising its SLO (cache-hit rate).
	\item Illustrate RLCache ability to dynamically adjust its caching strategies according to changes in the workload, by evaluating its SLO metrics:  hit-rate, the ratio of caching operations, the difference between estimated and optimal TTL, and the accuracy of objects evictions.
	\item Demonstrate the flexibility of the observer framework by using it as the backbone for all the methods and operations used in this chapter (including the baselines).
	\item Show that multi-task agent provides full automation to all components of a CM.
\end{enumerate}

CMs are logical components that group several methods together.
System developers choose a combination of methods that best suits their expected workload.
An open-source implementation of Redis CM \cite{redisspring} uses by default a write-through strategy for caching, Fixed TTL for object expiration, and LFU for the eviction strategy. 
While Caffeine \cite{caffeine} uses LFU for eviction, write-on-read for caching, and a fixed TTL.
To generalise and provide a fair comparison to standard systems, each component of RLCache evaluated in isolation against a rule-based method:
\begin{itemize}
	\item \textbf{Eviction.} LFU, LRU, and FIFO are widely used general purpose methods (\autoref{background:eviction}). 
	\item \textbf{Caching.} Write-through and Write-on-read (\autoref{sec:caching_decision}).
	\item \textbf{TTL.} Fixed TTL expiration (\autoref{background:ttl}).
\end{itemize}
We could not evaluate against a statistical augmented methods, due to lack of reproducible open-source implementations of the work we mentioned in the related work section.
Furthermore, this work, as far as we are aware, is the only one that evaluates and optimises for object based on their cache key, and as such we cannot draw a direct comparison to reported metrics from related work.
\section{Experiments Setup}\label{sec:simulator}
\subsection{Simulator}
RL algorithms require a large number of samples to learn meaningful actions, in production services, the agent learns from real traffic logs or metrics.
However, experimental environments lack real traffic so we used a simulator to generate synthetic traffic to compensate for that.
Yahoo Cloud Serving Benchmark (YCSB) \cite{cooper2010benchmarking} is a Java framework that provides a set of workloads for evaluating the performance of different key-value stores.
The framework comprised of two main components, a YCSB client, and a workload generator.
The client decides on which records to read or write by sampling from a random distribution,
YCSB has multiple predefined distributions, one of which is the Zipfian distribution that we used.
Zipfian distribution follows the popularity distribution where a small number of items are far more popular than the rest, and web caches do follow this distribution \cite{breslau1999web}.

We extended the YCSB framework and defined a set of workloads to test ranging from read-heavy, a mixture of reads and write, and write-heavy workloads, as shown in \autoref{table:workloads}.
While for simple web caches the workload might not change often, we argue that RLCache applicable to systems that are the building blocks of other systems.
For example, BigTable is the underlying storage for other systems such as databases like MegaStore (and its successors);  batch jobs such as MapReduce to store data between jobs.
Thus we argue that these systems exist and are already part of many large scale applications.
Furthermore, varying the workload demonstrates that RLCache does not require any manual configuring from the system developers.
Automating the CM reduces the barrier to developing systems and offloads the tedious task of profiling and experimenting with configurations.

\begin{figure}[!htb]
	\centering
	\includegraphics[width=1.0\linewidth]{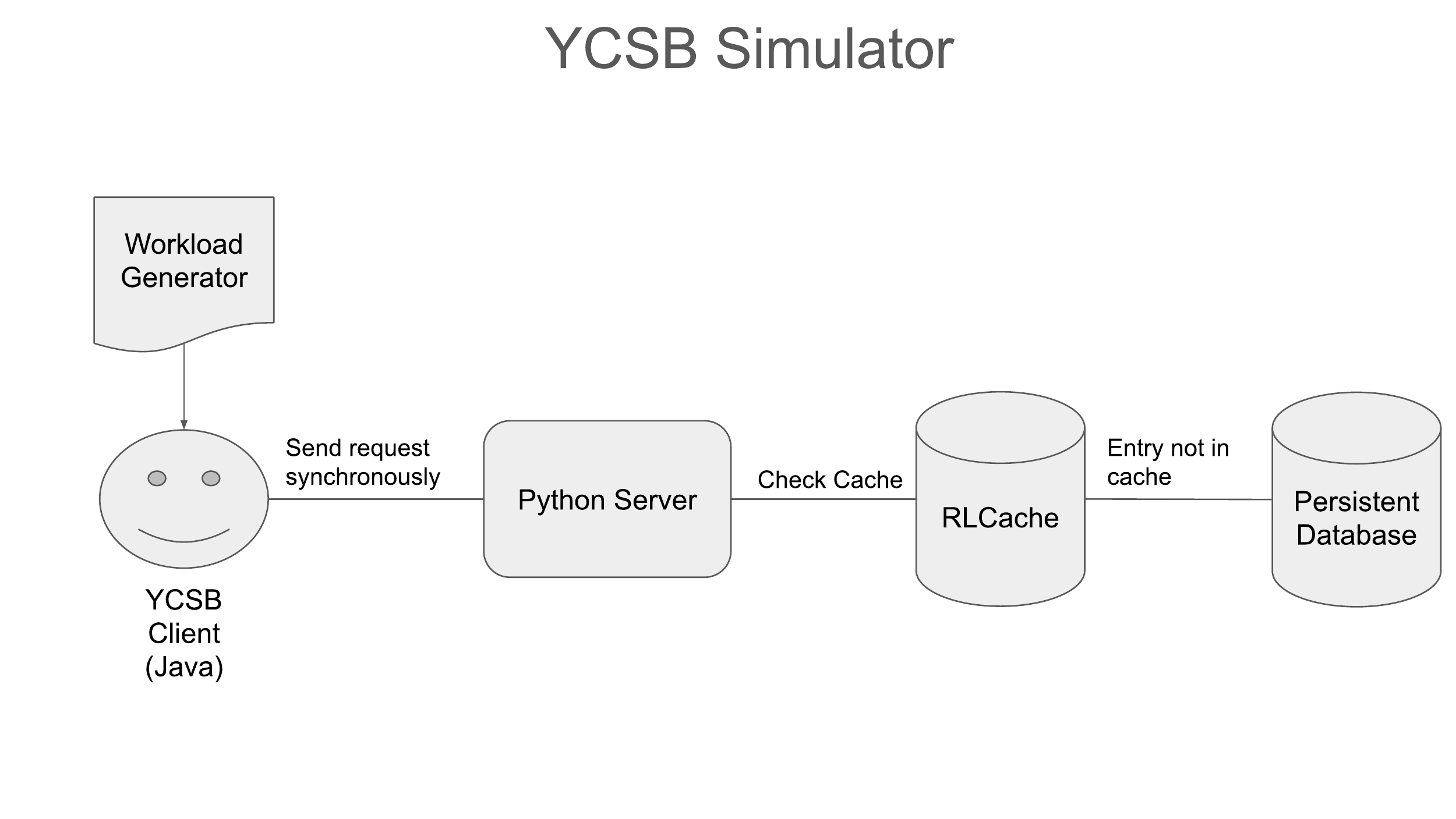}
	\caption[YCSB client to cache manager communication.]{The communication diagram between YCSB and the cache manager.}
	\label{fig:ycsbfig}
\end{figure}

Since the YCSB client needs to be in Java, we wrote a Java client that communicates with YCSB workload generators and then issues REST requests to the CM server.
The CM server is written in Python to be compatible with the established RL frameworks such as Tensorflow \cite{Abadi2016} and RLGraph \cite{Schaarschmidt2018rlgraph}.
\autoref{fig:ycsbfig} demonstrates the communication between the YCSB client and the CM.
The CM runs on the synchronous thread-per-request model to simplify the development process and avoid possible race conditions. 
We use in-memory storage for the cache in the simulation.
To build the RL agents, we used the Cambridge University's RLGraph \cite{Schaarschmidt2018rlgraph}, a modular RL framework with several high-performance agents implementations such as DQN and SAC. 

At the start of an experiment, the backend is cleared of any record, then YCSB client inserts $10,000$ records to prepare the simulator for upcoming queries. 
After the initial loading phase, the evaluation takes place.
Each workload generates $100,000$ queries, sampled from a the Zipfian distribution with hot spot fraction set to $20\%$ of the overall data. 
We run four workloads per experiments totalling $400,000$ queries, and each experiment ran three times, we reported the average of those runs.
We chose three runs to get reasonable confidence in the reported results without being it considerably expensive, as we have run a large number of experiments: five different agents, on four different capacity limits, each repeated three times.
Each experiment took $13$ hours to complete on an $i7-7700HQ@2.80GH$ with a $16GB$ memory. 

\begin{table}
\begin{tabular}{|c|c|c|c|}
	\hline 
	\textbf{Workload} & \textbf{Read \% }& \textbf{Write \% }& \textbf{Application Example} \\ 
	\hline 
	Read Only & $100$ & $0$ & User profile cache. \\ 
	\hline 
	Read Mostly & $95$ & $5$ & Photo tagging. Tag is heavily read  \cite{tang2015ripq,huang2013analysis}. \\ 
	\hline 
	Read Dominant & $75$ & $25$ & Database read and correct job.\cite{dellkrantz2012performance} \\ 
	\hline 
	Read/Write Mix & $50$ & $50$ & Session store of recent actions. \\ 
	\hline 
	Write Heavy & $0$ & $100$ & Batch write job.\cite{dellkrantz2012performance} \\ 
	\hline 
\end{tabular}
\caption{Request ratios in various workloads.}\label{table:workloads}
\end{table}

\subsection{Hyperparameters Selection}
RL algorithms have a large number of configurable parameters. 
This subsection covers the process we used to choose these parameters and their role in the experiments.

\textbf{Caching and eviction agents. }
We used the same parameters for both the caching and eviction agents, as they share the state and action representations despite fulfilling the opposite job.
For the agent's algorithm, we used DQN as described in \autoref{agent:dqn}, DQN is an off-policy discrete action algorithm that is proven to work well for general applications.
We configured the exploration constant $\epsilon = 1.0$ that decays to $0.2$ over $250,000$ steps, the large number of steps is based on the observation of the experiment that the agent's learning stabilises at that point..
The lower-bound of $0.2$ allows the agent to perform a random action $20\%$ of the time to ensure it is never stuck in local maxima when the workload changes.
As an example, when the agent learns best actions to perform when the workload is read-heavy, and then the workload changes to a write-heavy, with a $20\%$ probability the agent is taken random actions, with enough samples the agent will learn that the workload changed to write-heavy.
The impact of this constant for our experiment had the biggest impact, a lower value made the agent performs better for static workload, while a higher value ensured the agent is able to adapt to changes to the workload, choosing too high of a value and the agent will simply be a random agent and not learning, a value of $\epsilon \in [0.05, 0.25]$ is what we suggest to experiment with.

For the network structure, we used a two-layer network with $64$ hidden units.
We chose unit size $\le 128$ since the experiments ran on the CPU only. 
CPUs perform better tensor operations when the network size is smaller. 
In contrast, a GPU handles larger networks better as it can multi-thread and parallelise the operations efficiently. 
A larger network did not yield any better results as the state-value representations are relatively small (the encoded state vector is $8$ values and the action is a binary). 
Therefore, a larger network yields slower performance and result in unstable learning.

\textbf{Multi-task and TTL estimation agent. }
As both agents require an algorithm that supports continuous action-space and is an off-policy, we used SAC for the agent (\autoref{agent:sac}).
The network has three hidden-layers for both of SAC's value and policy networks, each with $128$ units.
While the reasoning behind the network size is similar to the one we had for caching and eviction, SAC, in general, requires a wider and deeper network by algorithm design \cite{haarnoja2018softApplications}.
SAC does not have an exploration constant and instead. 
It maximises the entropy that is used in choosing a random variable, making it robust to changes in the workload.

\textbf{Network optimiser and activation functions. }
For the neural networks activation function we used ReLu \cite{ramachandran2017searching}, and used Adam \cite{kingma2014adam} as the learning optimiser, with the learning rate set to $0.0003$. 
These parameters were researched and evaluated in \cite{henderson2018deep} and are considered what works best for most general cases.

\section{Experiments}\label{sec:evaluation}
\subsection{Caching Strategy}\label{sec:eval_caching}
\textbf{Intro.} The decision to cache an object directly impacts the cache hit rate as well as the slowdown caused by eviction policy once the cache is full.
A good caching strategy aims to maximise the cache hit rate and by caching only needed objects, thus reducing the number of commit calls to the cache.
For that, we measure two main performance metrics: the cache hit rate, and the caching rate.
Reducing the number of times an object is cached while achieving a similar hit rate results in reduced latency as fewer network hops are made, as well as, fewer objects are occupying the cache.

\begin{figure}[!htb]
	\centering
	\includegraphics[width=0.7\linewidth]{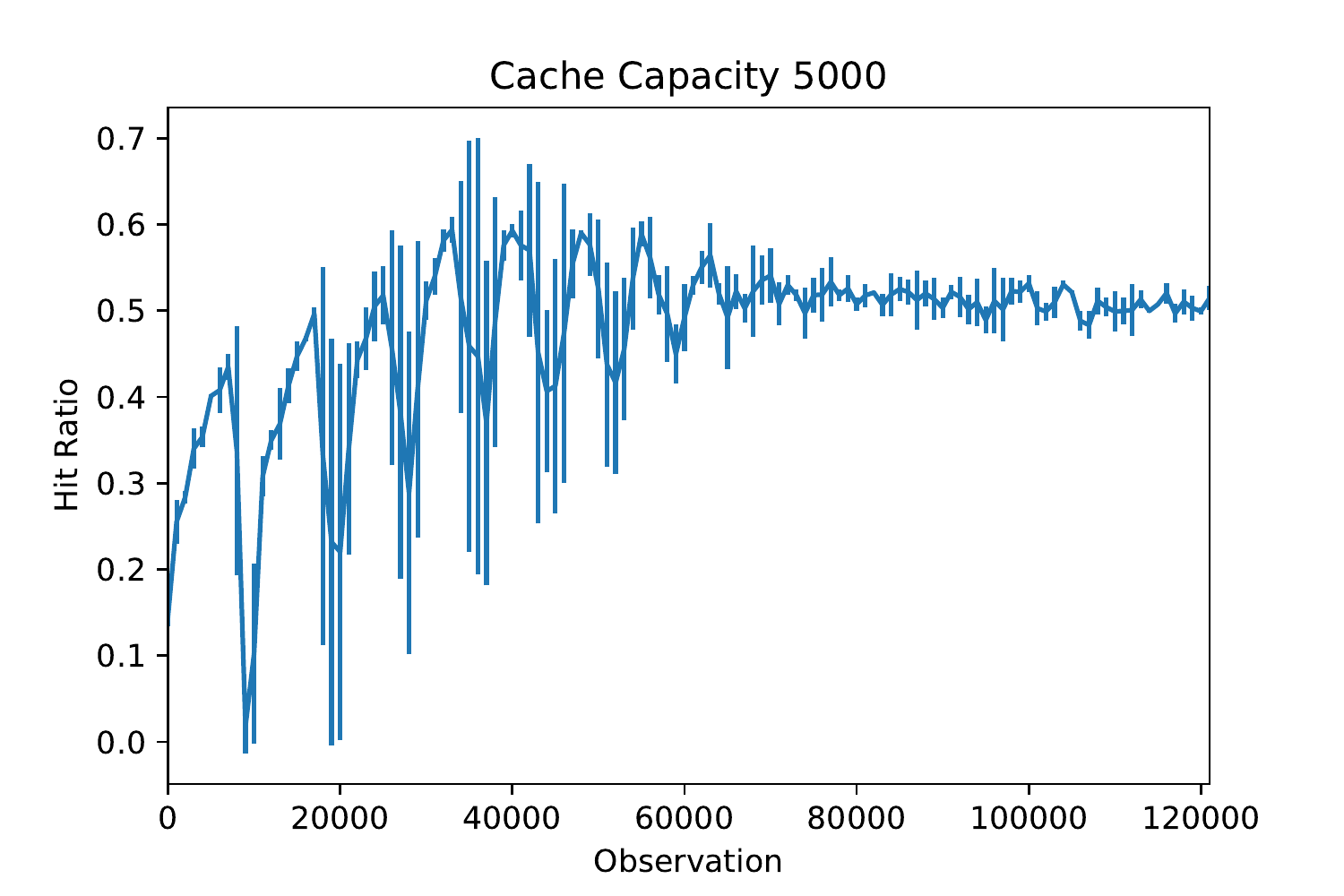}
	\caption[RLCache caching strategy learning overtime.]{RLCache agent learning a caching strategy overtime.}
	\label{fig:hitratezoomed5000}
\end{figure}

\textbf{Experiments.} First, we demonstrate the ability of RLCache's agent to learn caching strategy.
\autoref{fig:hitratezoomed5000} shows the cache hit rate performance for RLCache when the system is under $95\%$ read and $5\%$ write workload. 
The agent starts initially with no information about the environment as a result, the cache hit-rate starts as low as $20\%$. 
Overtime, as the agent is exploring and updating its neural network weights, it begins to learn optimal policy, and achieve $ \ge 50\%$ cache hit-rate.
We configured the agent to slowly decay its exploring parameter $\epsilon$ to 0.1 over a $50,000$ steps.
The agent chooses an action that maximises the Q value with the probability $1-\epsilon$, otherwise it samples uniformly from the available action space, hence the sine-wave shape of the cache performance early on. 
The reason the $\epsilon$ never decays to zero, is to allow the agent to adapt to when the workload changes, as can be seen in the next evaluation.

\begin{figure}[!htb]
	\centering
	\begin{subfigure}{.5\textwidth}
		\centering
		\includegraphics[width=1.0\textwidth]{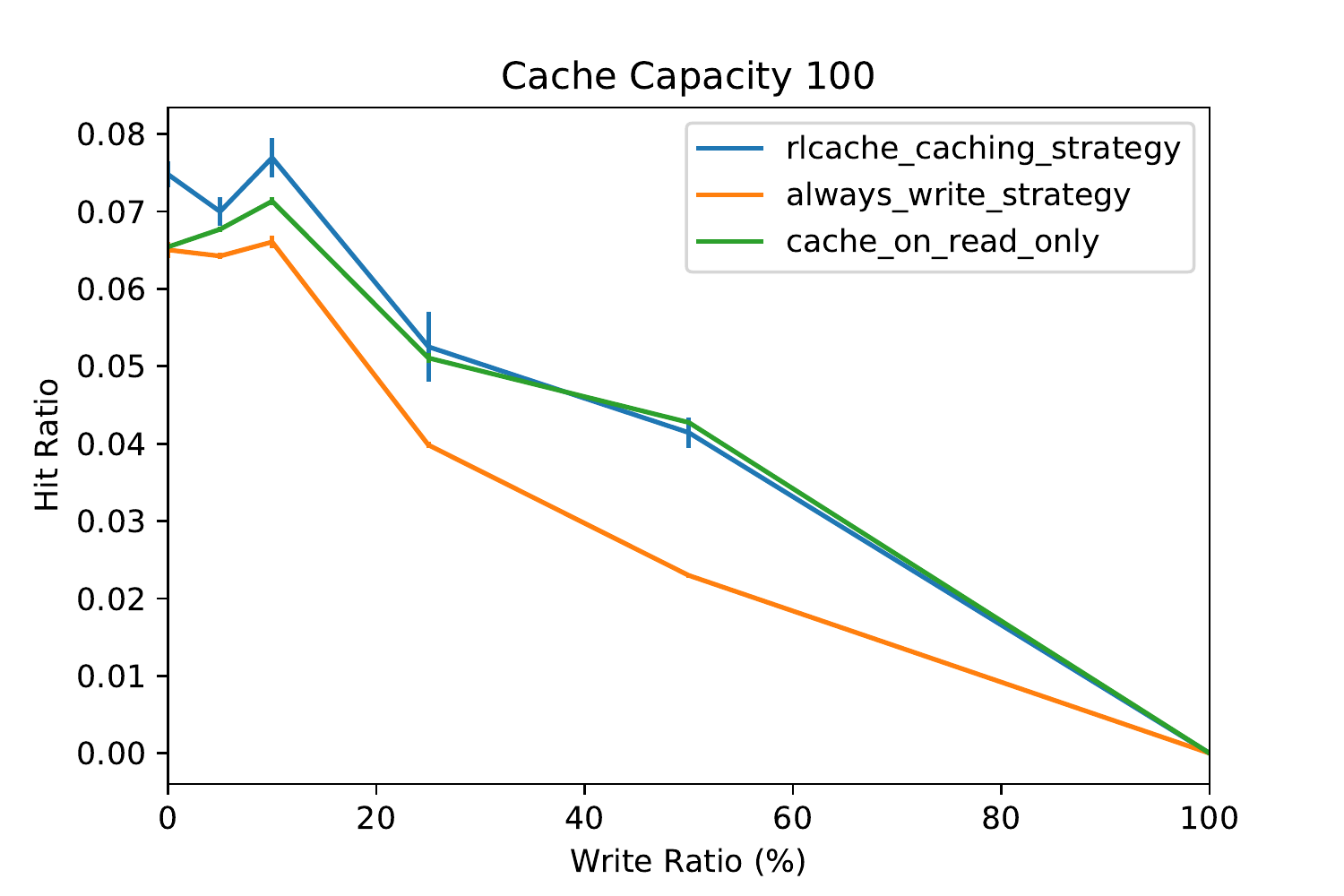}
	\end{subfigure}%
	\begin{subfigure}{.5\textwidth}
		\centering
		\includegraphics[width=1.0\textwidth]{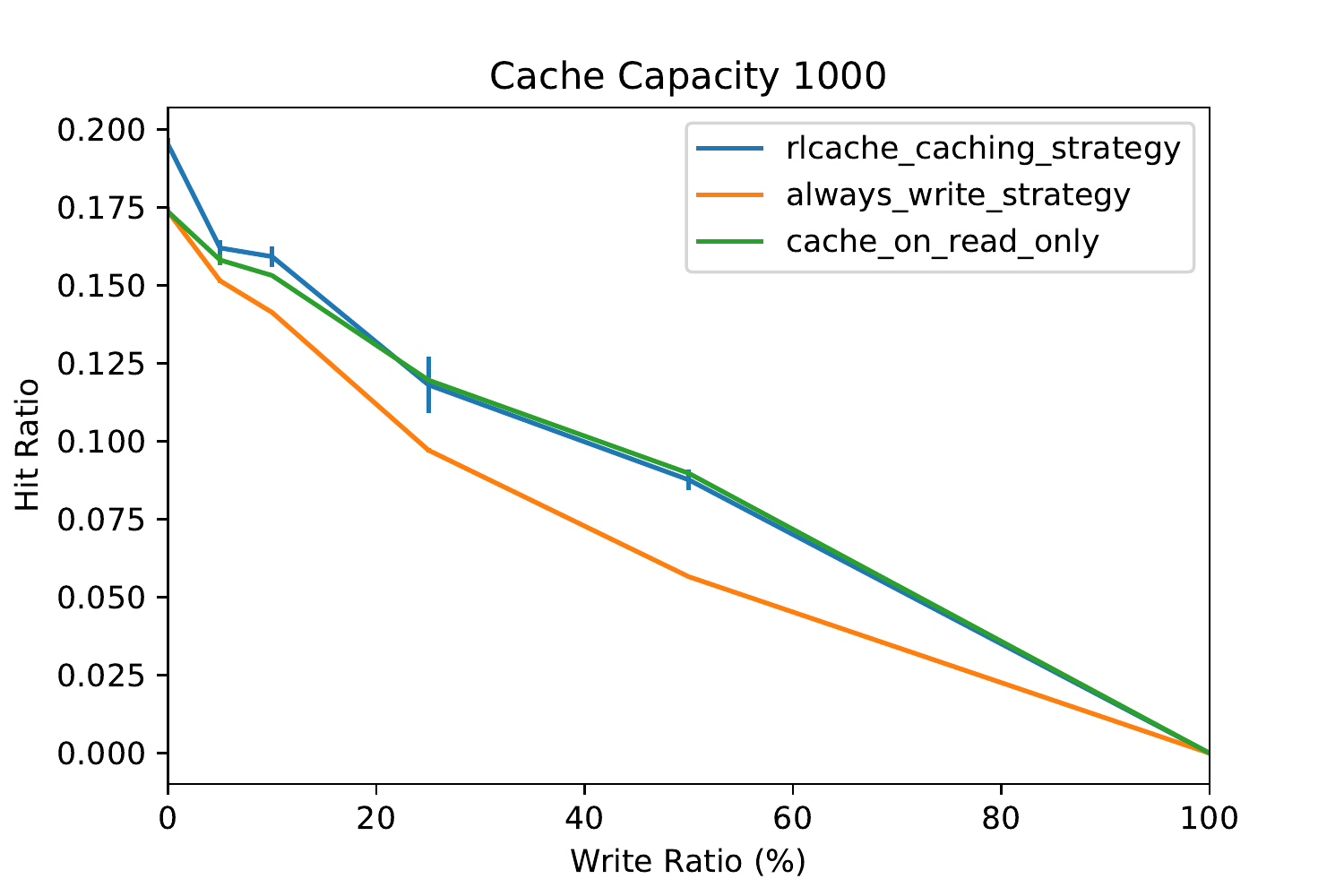}
	\end{subfigure}
	\begin{subfigure}{.5\textwidth}
		\centering
		\includegraphics[width=1.0\textwidth]{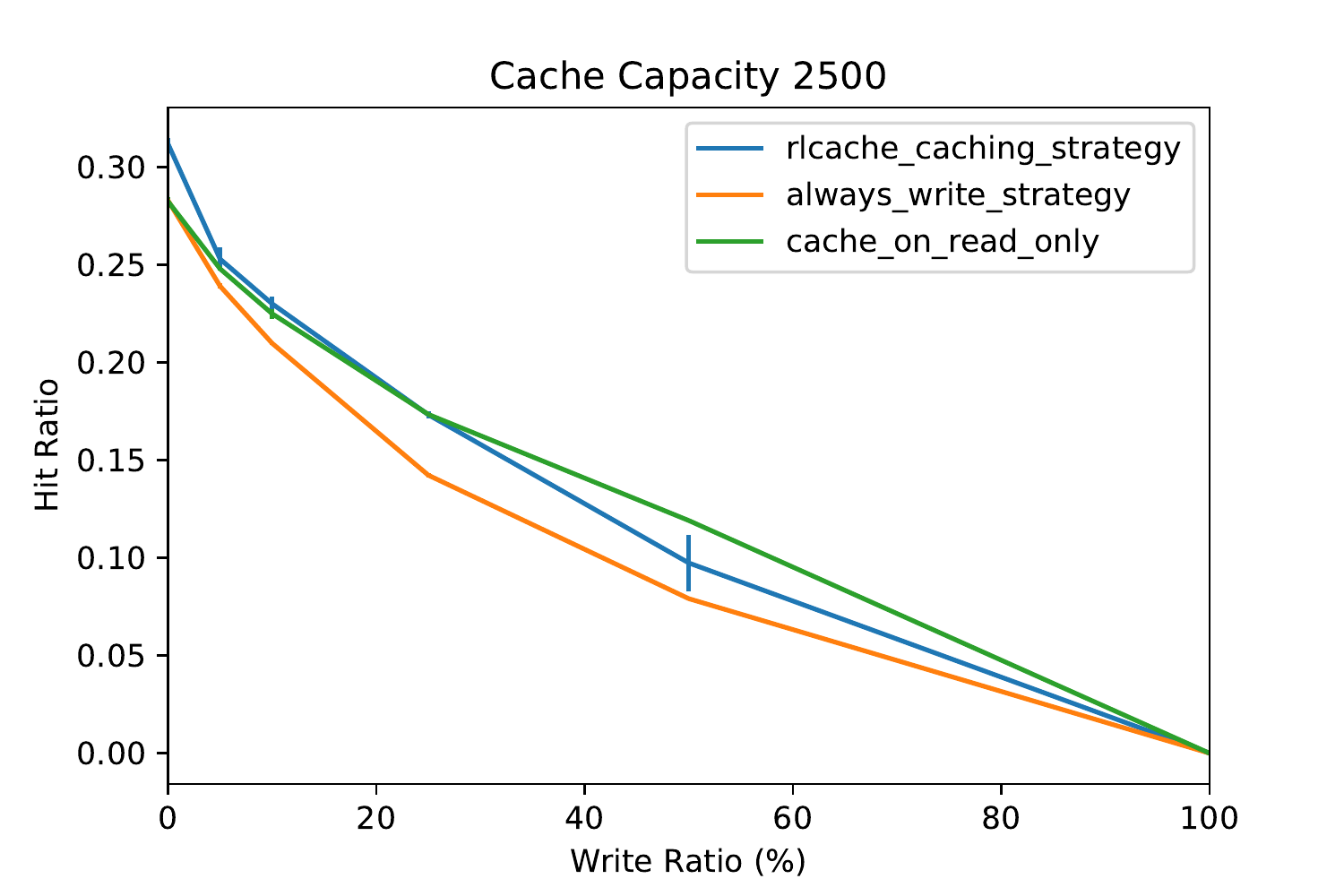}
	\end{subfigure}%
	\begin{subfigure}{.5\textwidth}
		\centering
		\includegraphics[width=1.0\textwidth]{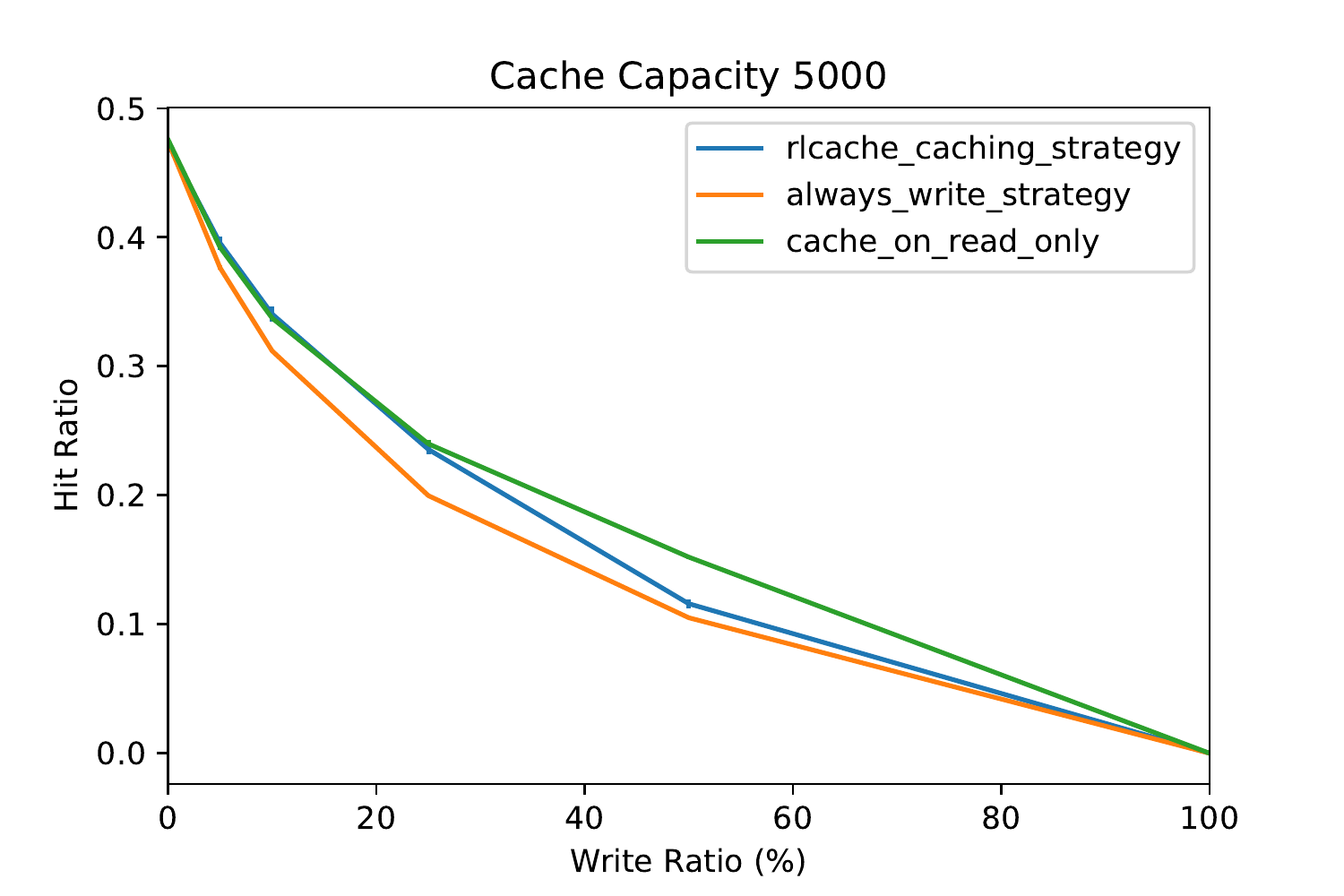}
	\end{subfigure}
	\caption[RLCache caching strategy hit-rate evaluation.]{Comparison between the hit rate of various cache capacities and RLCache caching strategy. Error bars with standard deviation less than 0.05 are omitted.}
	\label{fig:rlcache_caching_comparisons}
\end{figure}

\autoref{fig:rlcache_caching_comparisons} shows the percentage of the cache operations that results in cache hits.
It demonstrates the capability of RLCache while varying both the write ratio and the cache capacity.
RLCache is always able to perform better than the traditional methods because of its ability to favour objects with higher hits when caching.
The cache hit approaching zero with higher write ratio is expected, as the number of requests that attempt to retrieve cached objects also approaches zero. 
The simulation ran for only $100,000$ queries for each type of workload.
We anticipate, given more queries per evaluation, the agent will learn more distinct features thus resulting in higher cache hit rate, as the trend shows.

\begin{figure}[!htb]
	\centering
	\begin{subfigure}{.5\textwidth}
		\centering
		\includegraphics[width=1.0\textwidth]{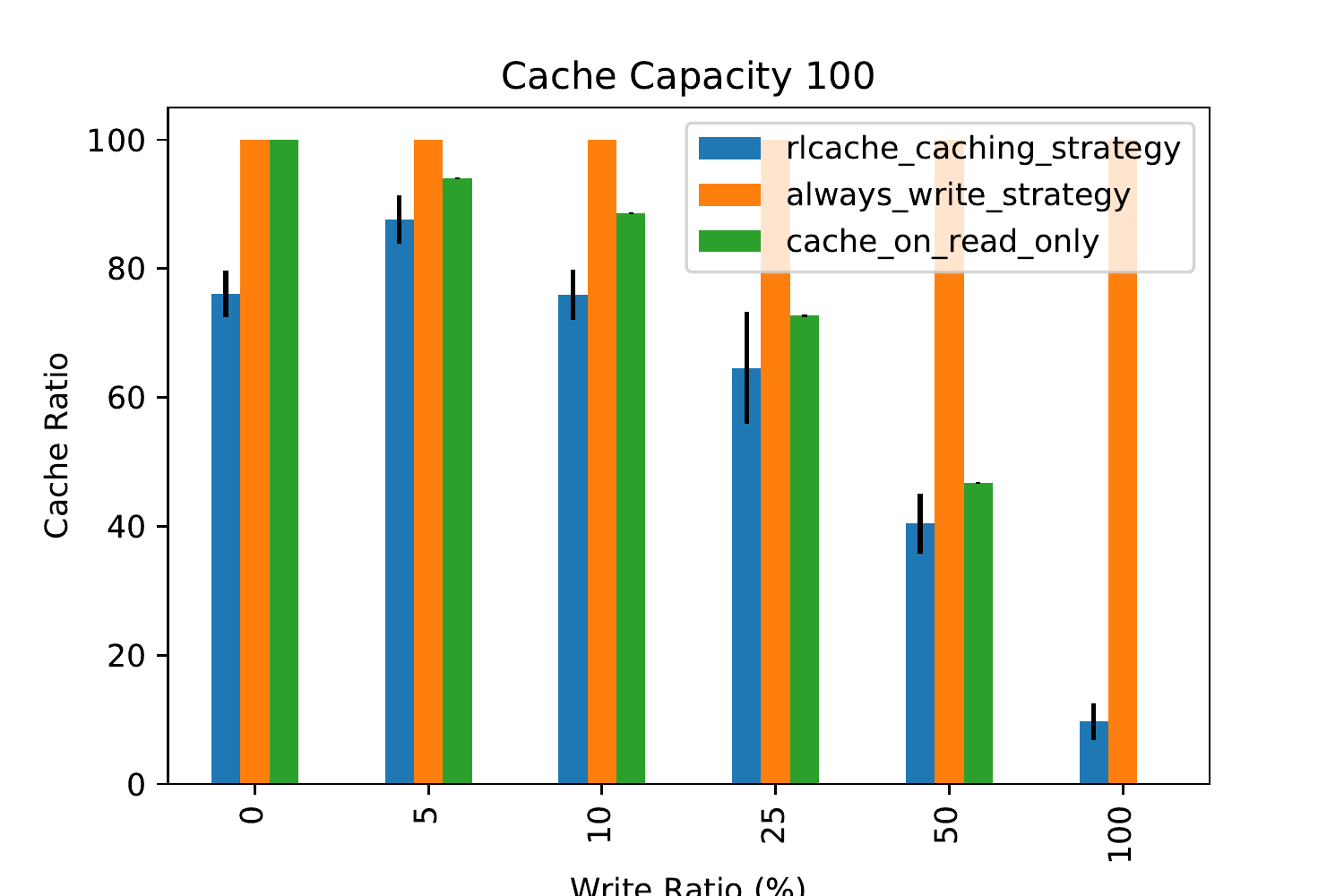}
	\end{subfigure}%
	\begin{subfigure}{.5\textwidth}
		\centering
		\includegraphics[width=1.0\textwidth]{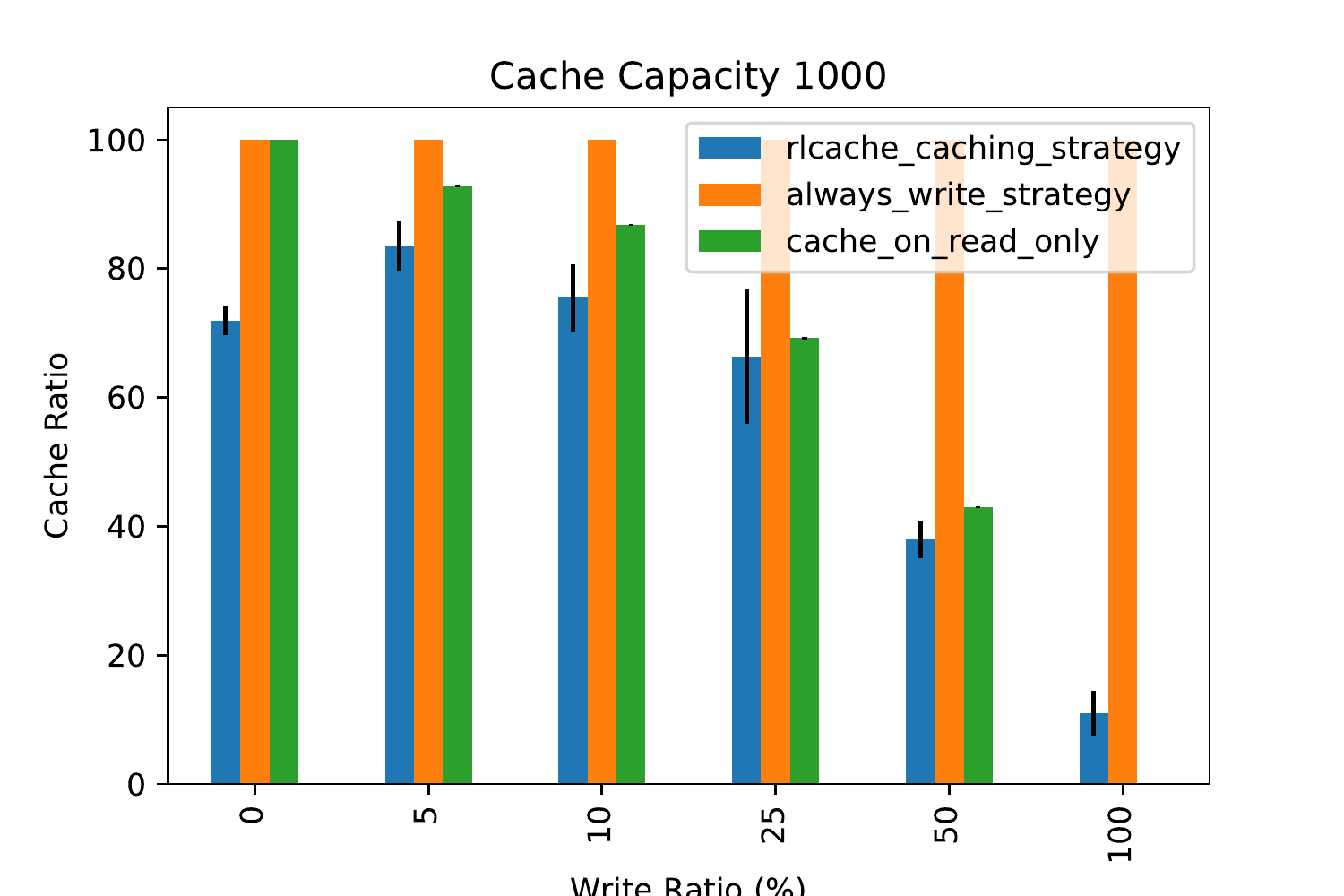}
	\end{subfigure}
	\begin{subfigure}{.5\textwidth}
		\centering
		\includegraphics[width=1.0\textwidth]{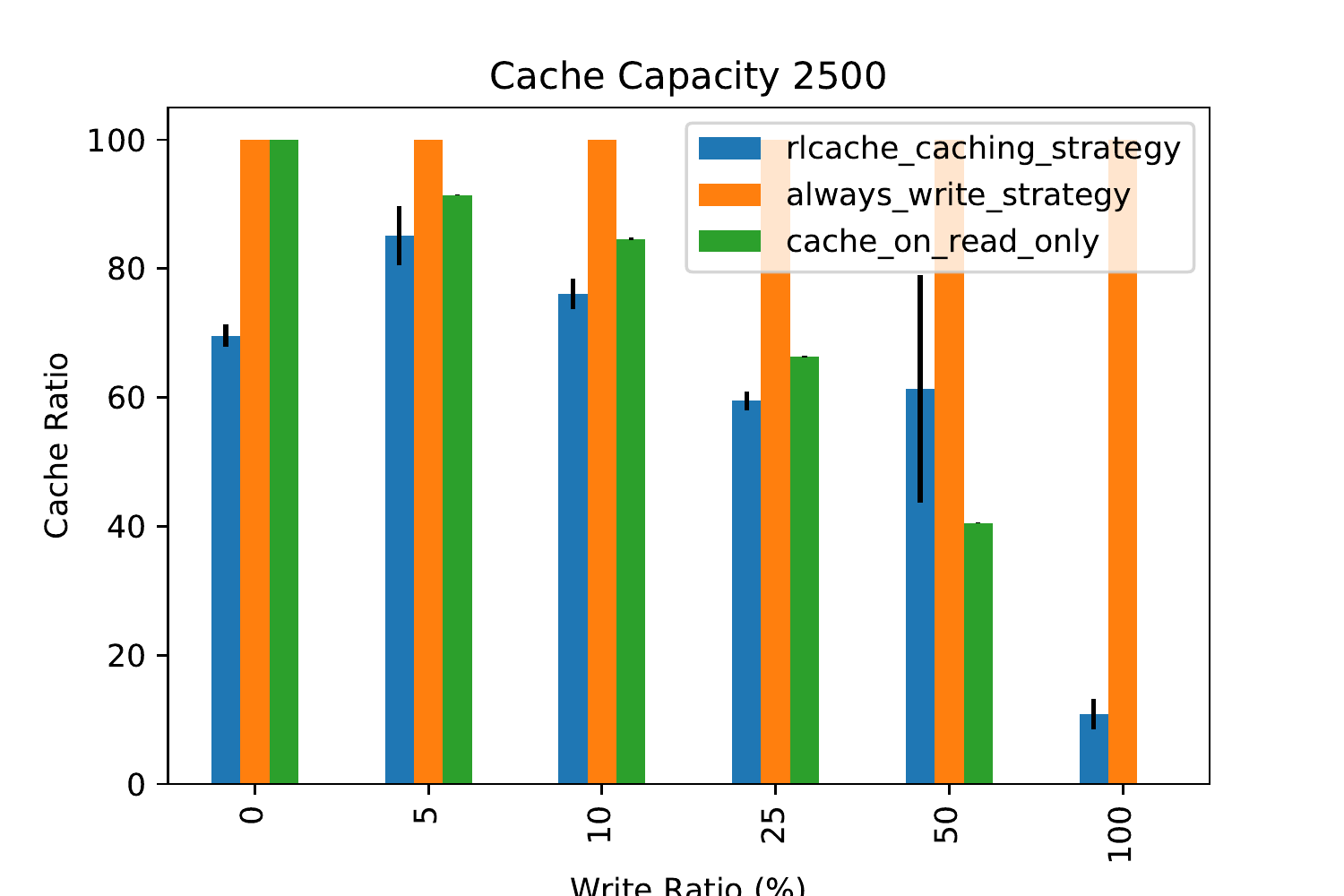}
	\end{subfigure}%
	\begin{subfigure}{.5\textwidth}
		\centering
		\includegraphics[width=1.0\textwidth]{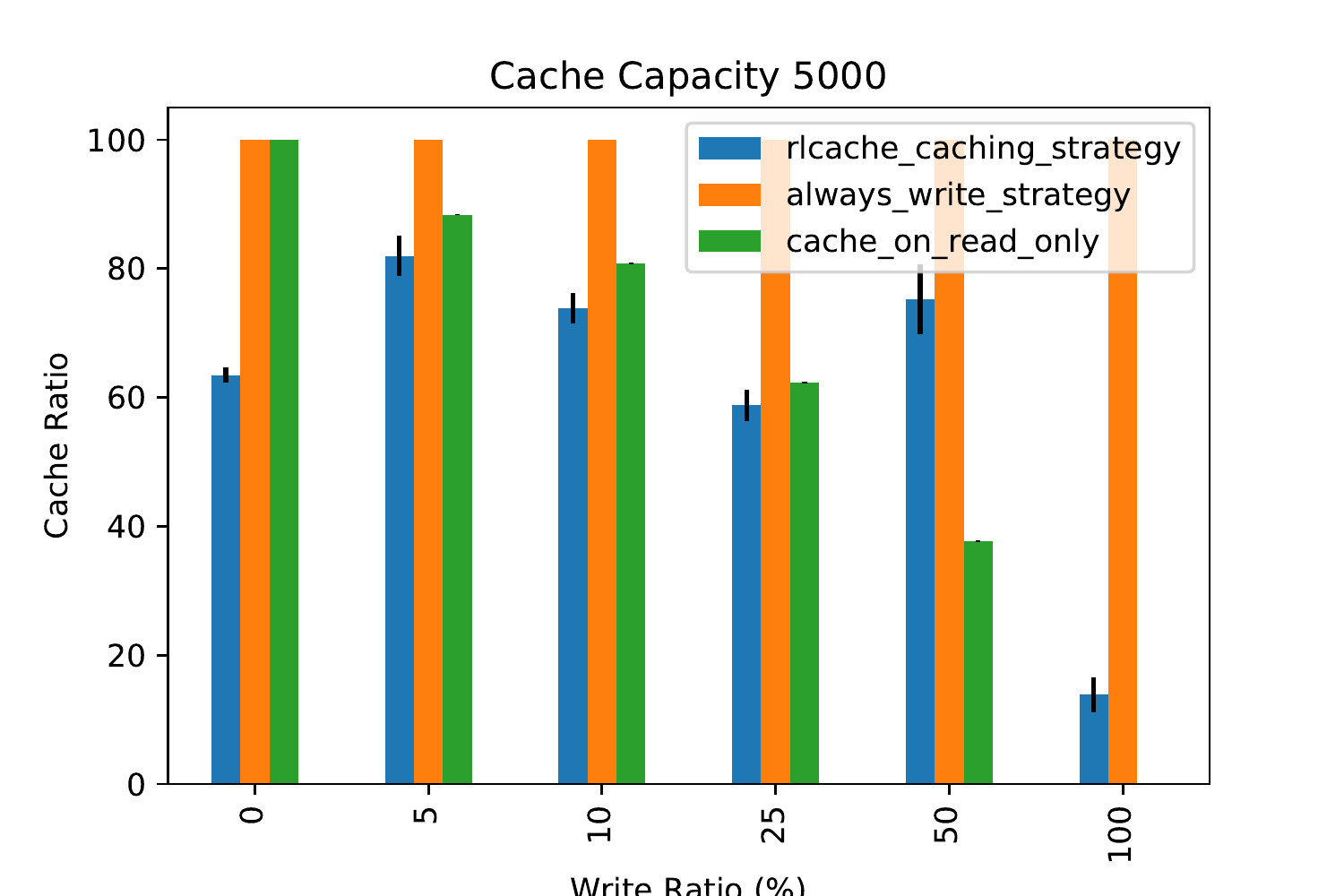}
	\end{subfigure}
	\caption[RLCache caching strategy and caching rate evaluation.]{Comparison between the caching rate of various cache capacities and RLCache caching strategy. Error bars with standard deviation less than 0.05 are omitted.}
	\label{fig:rlcache_caching_cache_ratio_comparisons}
\end{figure}

\autoref{fig:rlcache_caching_cache_ratio_comparisons} backs this claim, while RLCache achieves higher cache hit rate, it also does that with fewer calls to cache the objects, thus skipping objects that will not receive hits.
RLCache also opts to skip more objects when the write ratio increases, demonstrating its ability to change behaviour dynamically according to the workload.
This behaviour is strikingly obvious when the write rate is at 100\% but the caching rate is bellow 10\%, thus saving many trips to the cache when the backend is receiving a bulk update.

\textbf{Outro.} This comparison shows two things: the capability of the system in achieving a competitive cache hit rate, and its ability to adapt to a dynamic workload. 

\subsection{Eviction Strategy}\label{sec:eval_eviction}
\textbf{Intro.} The CM invokes the eviction strategy when the cache is full and needs to replace an old cached object with a newer one.
We configure the testing environment to track performance metrics by storing eviction decisions for the duration of the leftover TTL after evicting the object.
An optimal eviction strategy evicts the objects that are no longer useful and keeps the objects that will be used soon in the cache.
These properties can be measured using precision, recall, and f1 score.
The \textit{precision} is the number of correctly evicted or skipped objects to the total number of evictions, calculated according to \autoref{fig:precision}.
The \textit{recall} \autoref{fig:recall} is the ratio of correctly evicted objects to the total number of decisions, measured using \autoref{fig:precision}.
\textit{F1} \autoref{fig:f1} scores the balance between the precision and recall, per \autoref{fig:precision}.
The system collects the following metrics to calculate the above formulas:
\begin{itemize}
	\item \textbf{FalseEvict:} If in the period until the TTL is up the evicted object is requested again (thus causing a cache miss).
	\item \textbf{TrueEvict:} If the object is invalided within the TTL time or not requested.
	\item \textbf{TrueMiss:} If the object is requested after the eviction policy did not evict it.
	\item \textbf{FalseMiss:} If the object is invalided or not requested after the eviction policy did not evict it.
\end{itemize}
\begin{equation}\label{fig:precision}
precision = \frac{TrueEvict + TrueMiss}{TrueEvict + TrueMiss + FalseEvict}
\end{equation}
\begin{equation}\label{fig:recall}
recall = \frac{TrueEvict + TrueMiss}{TrueEvict + TrueMiss + FalseMiss}
\end{equation}
\begin{equation}\label{fig:f1}
f1 = 2 * \frac{precision * recall}{precision + recall}
\end{equation}

\textbf{Experiments.} We have modified an open source implementation \cite{cacheout} of LRU and FIFO to capture the above performance metrics.
To simplify comparison between runs, we set the TTL to be 60 seconds for all entries in the cache, and attempt to cache every entry that is read or written.
We varied the caching capacity to show the flexibility of RLCache in constrained space. 

\begin{figure}[!htb]
	\centering
	\begin{subfigure}{.5\textwidth}
		\centering
		\includegraphics[width=1.0\textwidth]{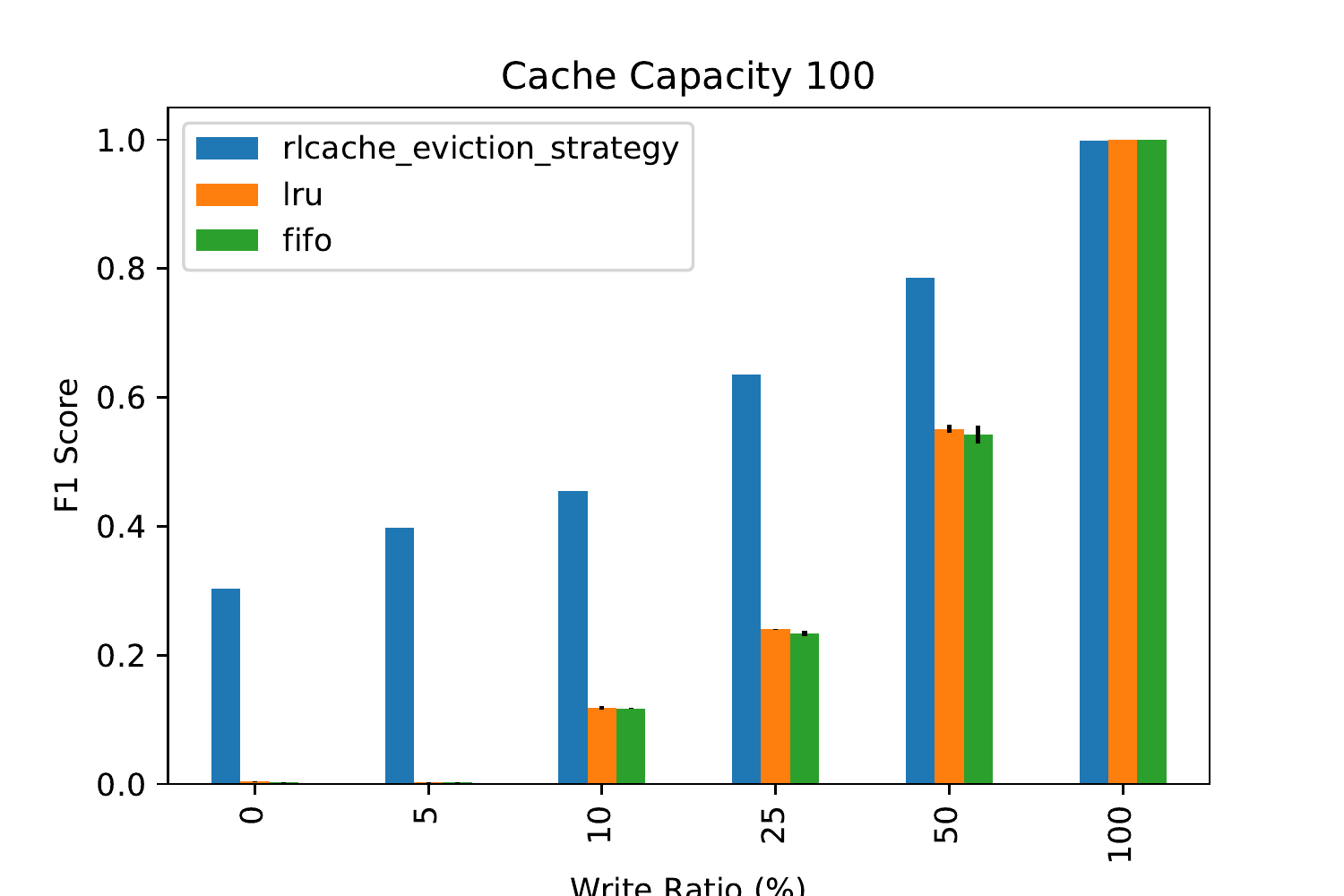}
	\end{subfigure}%
	\begin{subfigure}{.5\textwidth}
		\centering
		\includegraphics[width=1.0\textwidth]{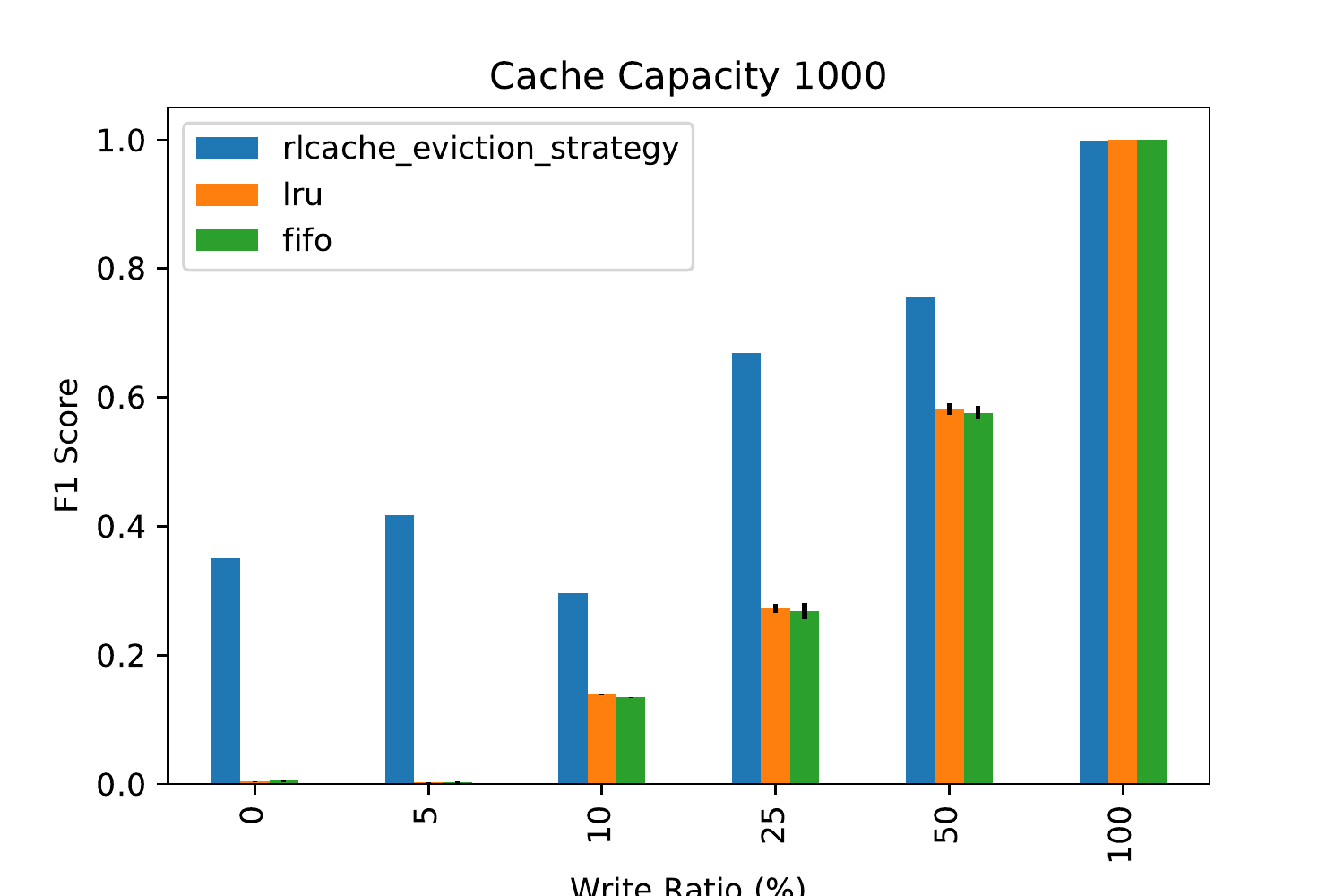}
	\end{subfigure}
	\begin{subfigure}{.5\textwidth}
		\centering
		\includegraphics[width=1.0\textwidth]{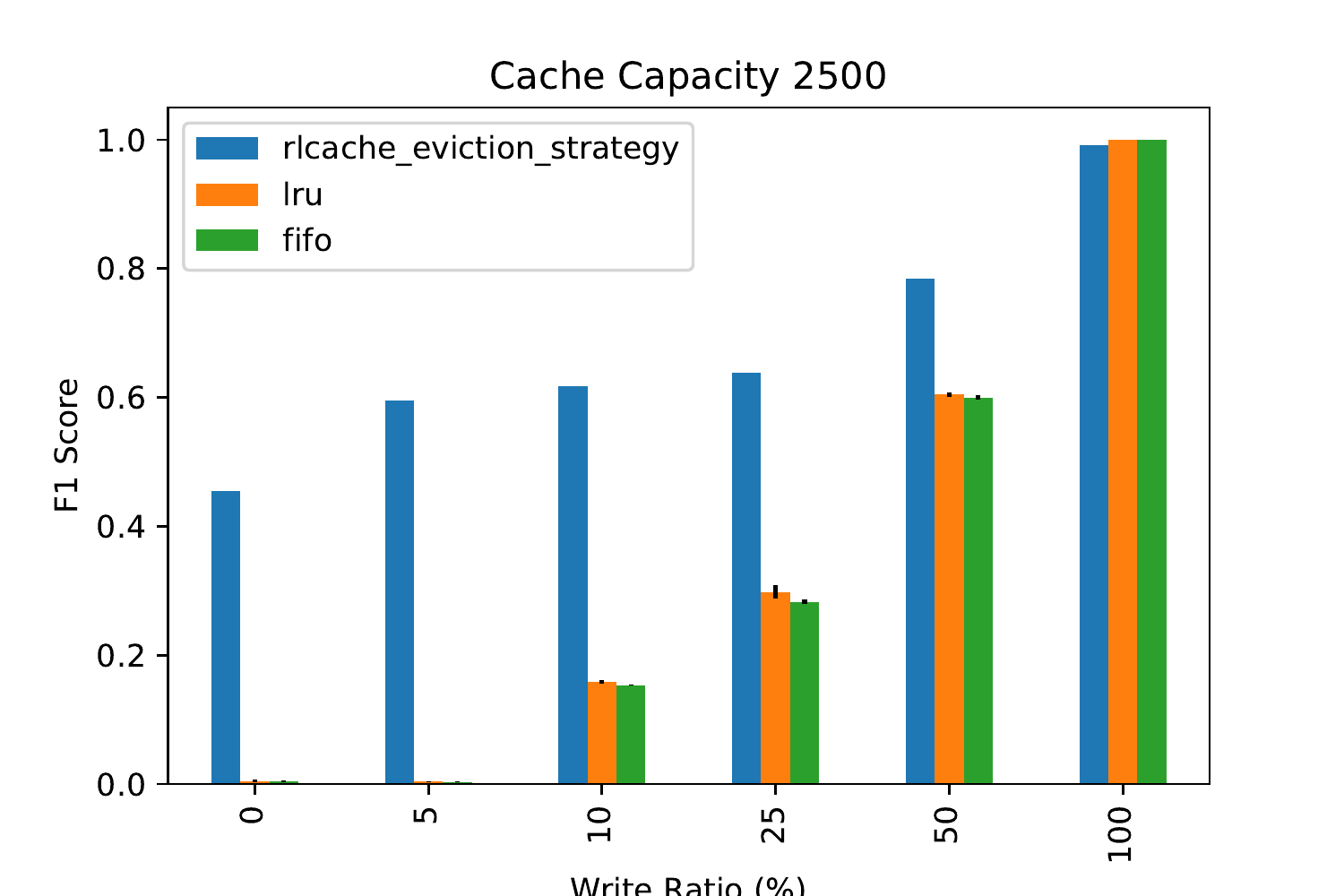}
	\end{subfigure}%
	\begin{subfigure}{.5\textwidth}
		\centering
		\includegraphics[width=1.0\textwidth]{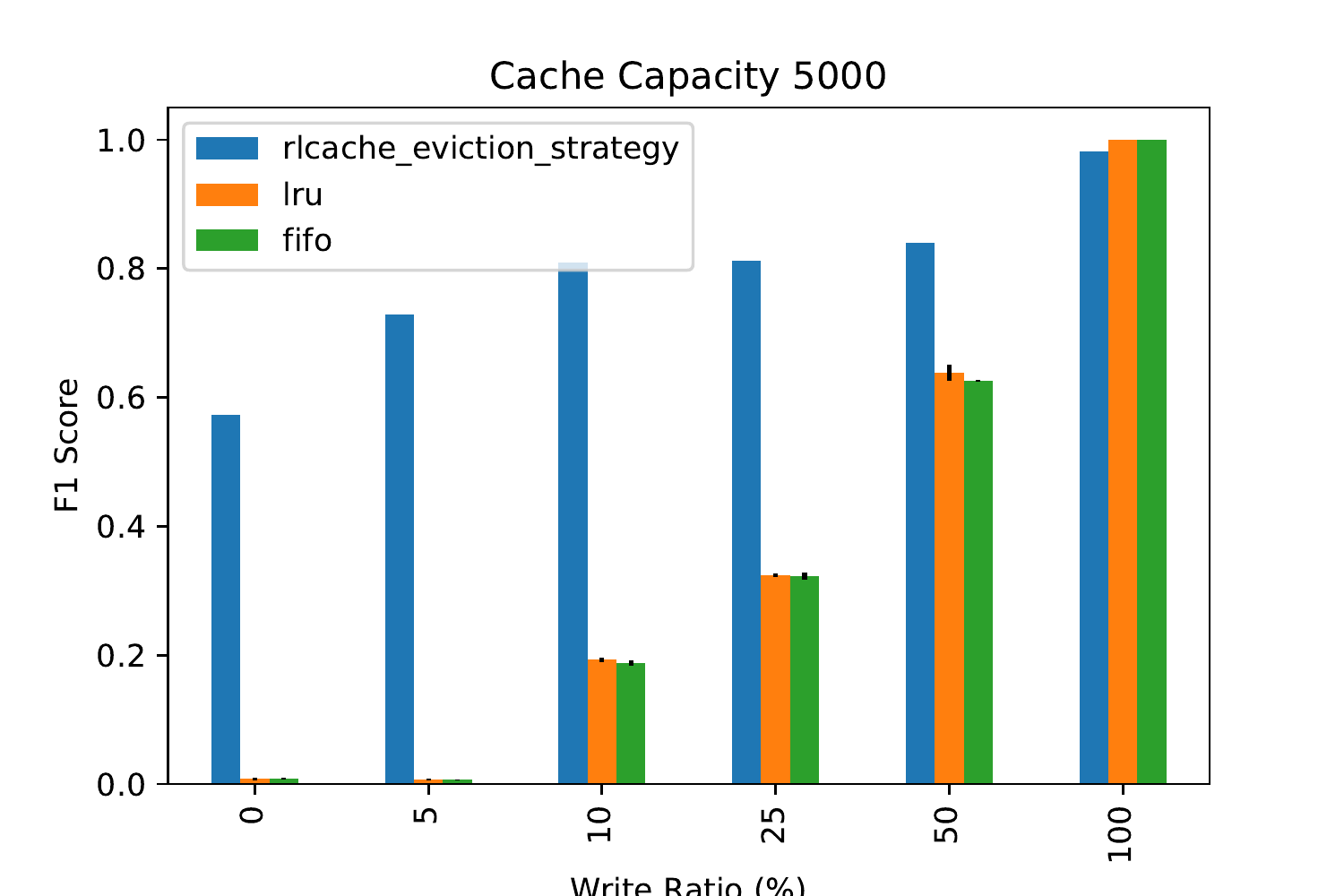}
	\end{subfigure}
	\caption[RLCache eviction strategy F1 evaluation.]{Comparison between the F1 score of various cache capacities and RLCache eviction strategy. Error bars with standard deviation less than 0.05 are omitted.}
	\label{fig:rlcache_f1_score_comparisons}
\end{figure}

\begin{figure}[!htb]
	\centering
	\begin{subfigure}{.5\textwidth}
		\centering
		\includegraphics[width=1.0\textwidth]{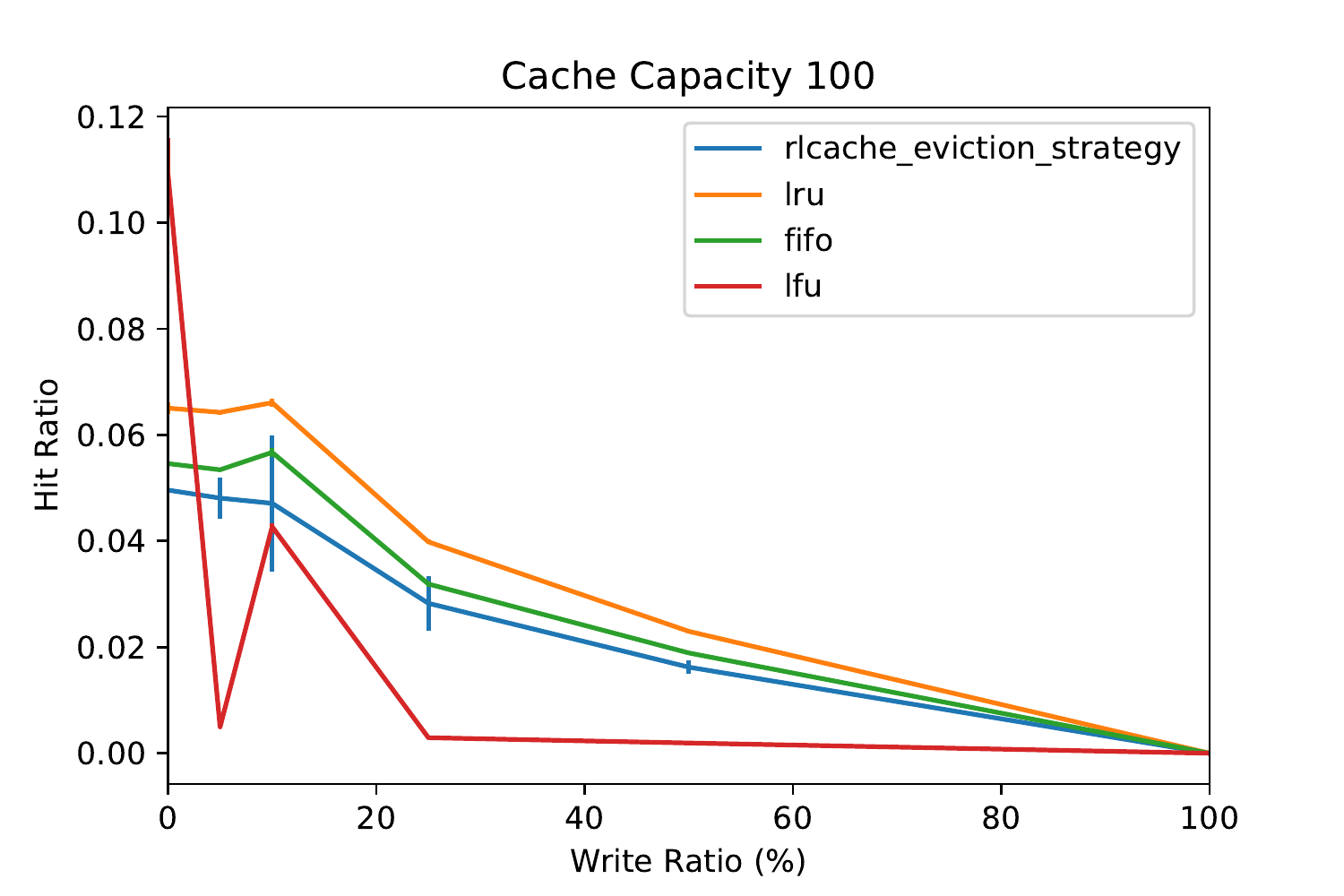}
	\end{subfigure}%
	\begin{subfigure}{.5\textwidth}
		\centering
		\includegraphics[width=1.0\textwidth]{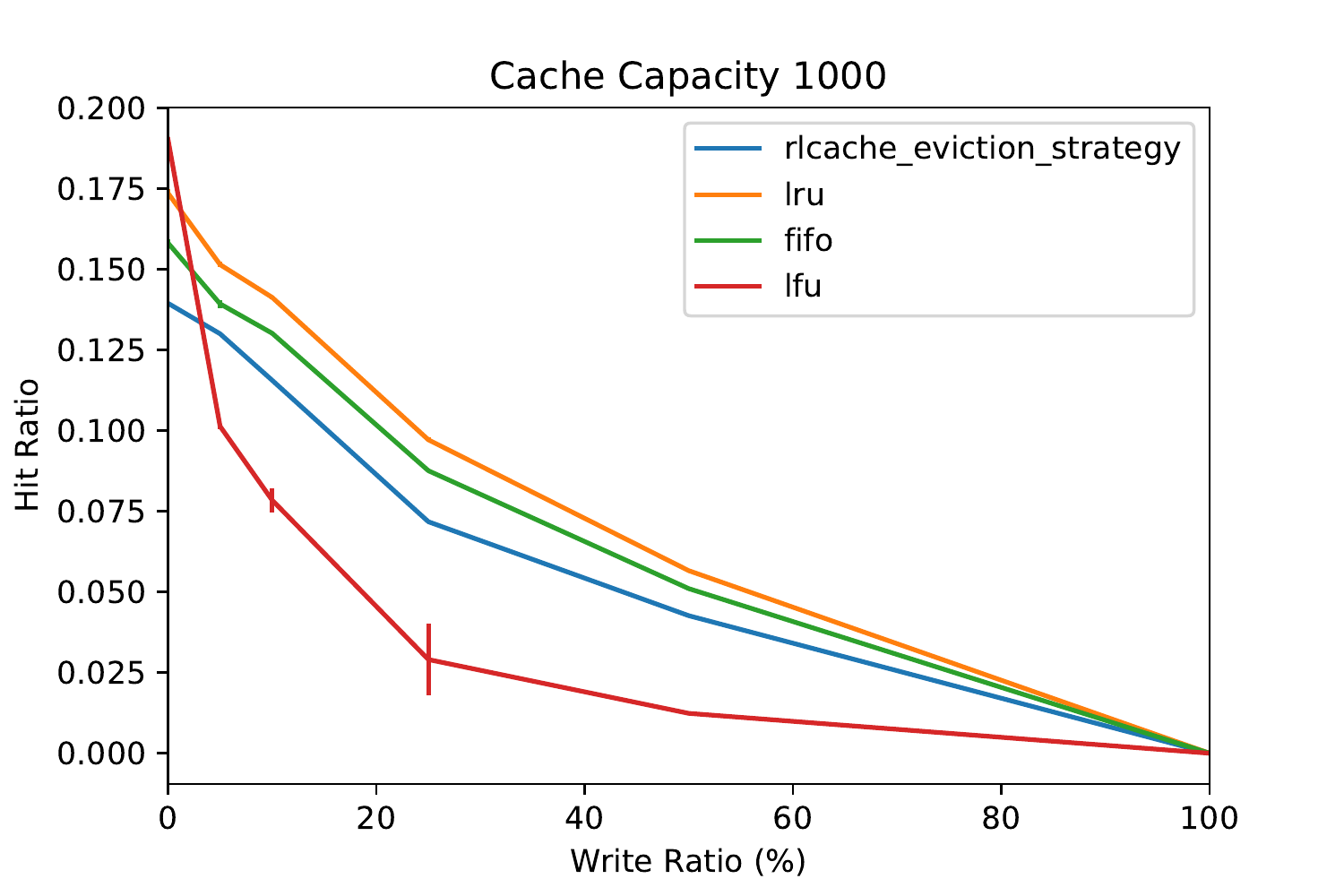}
	\end{subfigure}
	\begin{subfigure}{.5\textwidth}
		\centering
		\includegraphics[width=1.0\textwidth]{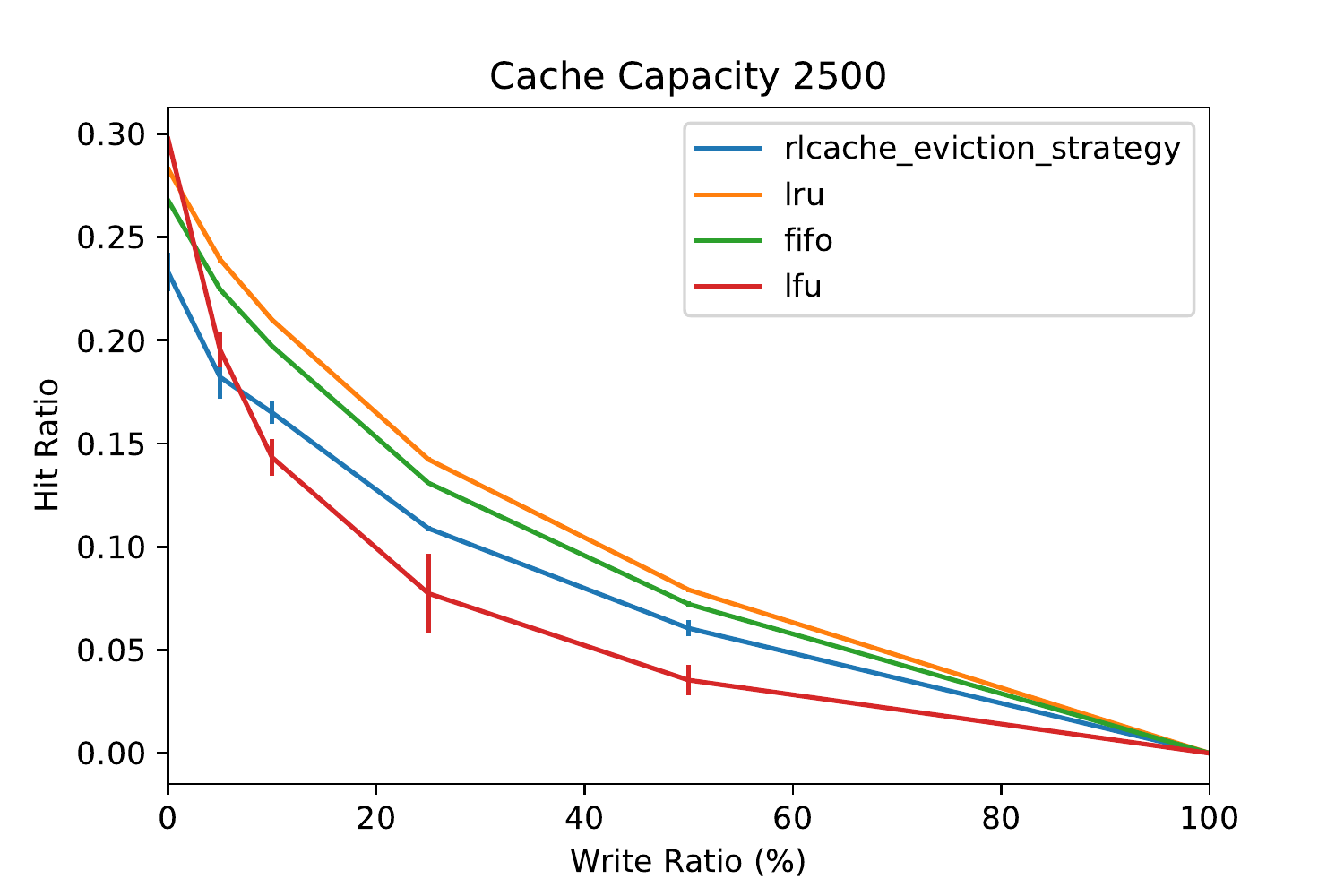}
	\end{subfigure}%
	\begin{subfigure}{.5\textwidth}
		\centering
		\includegraphics[width=1.0\textwidth]{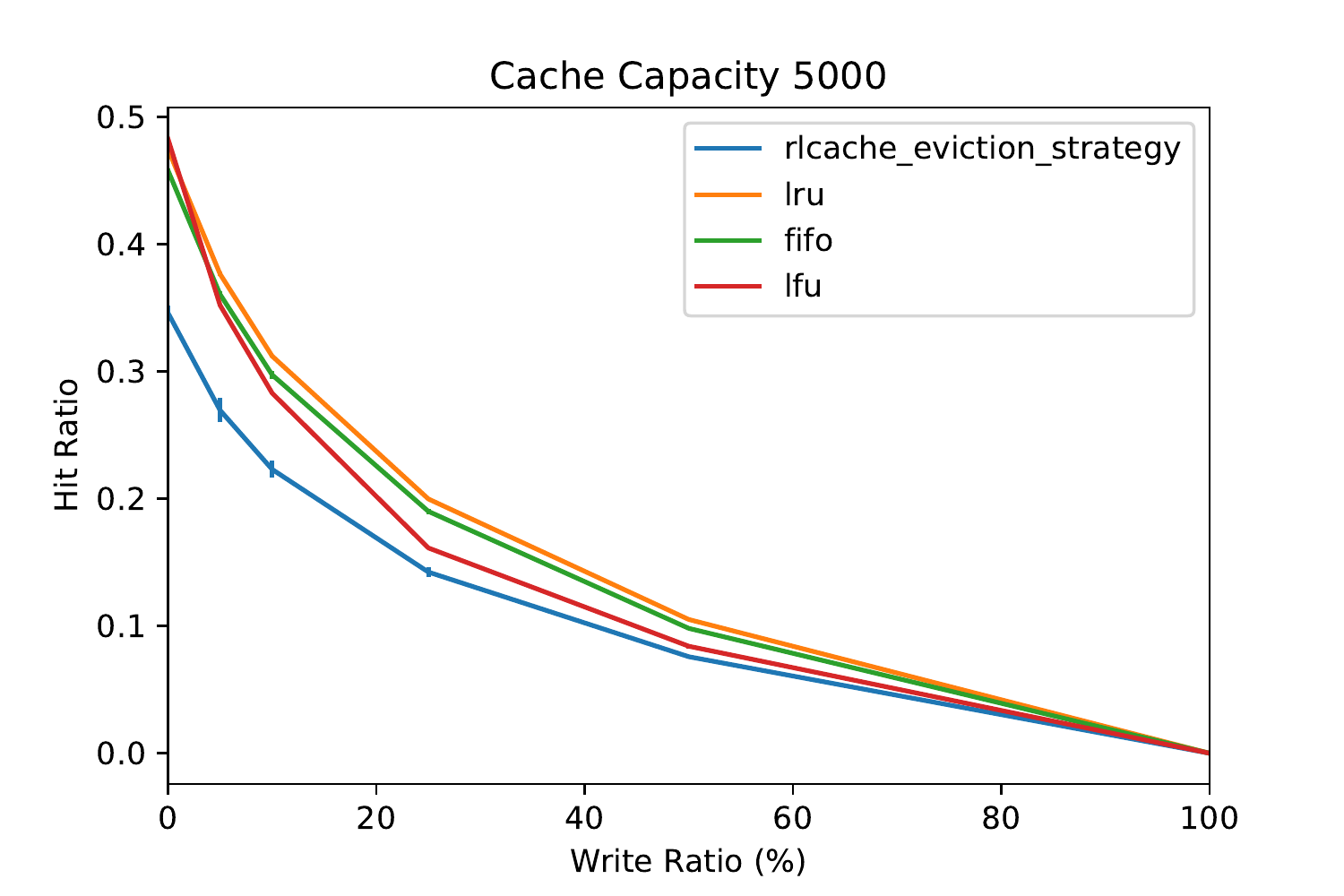}
	\end{subfigure}
	\caption[RLCache eviction strategy hit-rate evaluation.]{Comparison between the hit rate of various cache capacities and RLCache eviction strategy. Error bars with standard deviation less than 0.05 are omitted.}
	\label{fig:rlcache_eviction_hitrate_comparisons}
\end{figure}

RLCache's F1 score as shown in \autoref{fig:rlcache_f1_score_comparisons} far outperform the other baseline, meaning that RLCache eviction policy results in the eviction of undesirable objects more often than the baselines, and that is especially true for when the workload has some reads mixed in.
It can be noted that when the write ratio approaches $100\%$, the F1 also approaches $1.0$. 
Because in a write-heavy workload followup reads are rare, therefore any evict is a good evict.
This is a great property to have in an eviction policy, predicting when an object is no longer needed and prematurely evicting it provides the host cache with more space to store other newer objects.
However, \autoref{fig:rlcache_eviction_hitrate_comparisons} shows this did not result in a higher cache hit rate as we thought it would.
We believe this is due to several factors:
\begin{itemize}
	\item RLCache is able to correctly predict when an item will expire soon and evicts it before it naturally expires, therefore it accumulate much higher F1 score.
	\item RLCache outputs multiple evictions decisions at a time, while it has high precision (60\% at write ratio of 0\% and cache capacity of $5000$), because it outputs order of magnitudes more evictions it results in lower hit rate overall.
\end{itemize}

\textbf{Outro. }The eviction strategy is a promising direction for when evicting the 'wrong' item is costly in comparison to a cache miss. 
While the strategy itself does not improve the cache-hit rate, it has high accuracy in predicting which objects are likely to be invalidated soon.
Example use case of this is a on device cache in a smartphone, where the cached entry can be a video that will take several minutes to download, as opposite of a cache miss to a website logo.

\subsection{TTL Estimation}\label{sec:eval_ttl}
\textbf{Intro. }Estimating the TTL has two main objectives. 
The first is to give objects that often are requested a long time to stay in the cache.
The second is to reduce the time in the cache for objects that rarely receive reads or often gets invalidated.
From our experiments, we found maximising for these objectives (and as such the agent's reward function reflect that) yields the highest overall cache hit rate.
Earlier work such as the one in \cite{schaarschmidt2016learning} aimed to learn the time an object receives an invalidation request, and assign that estimated time to the object.
However, objects that rarely receive updates and do not receive any reads either will live in the cache longer occupying space without contributing to the performance of the cache.

\begin{figure}[!htb]
	\centering
	\begin{subfigure}{.5\textwidth}
		\centering
		\includegraphics[width=1.0\textwidth]{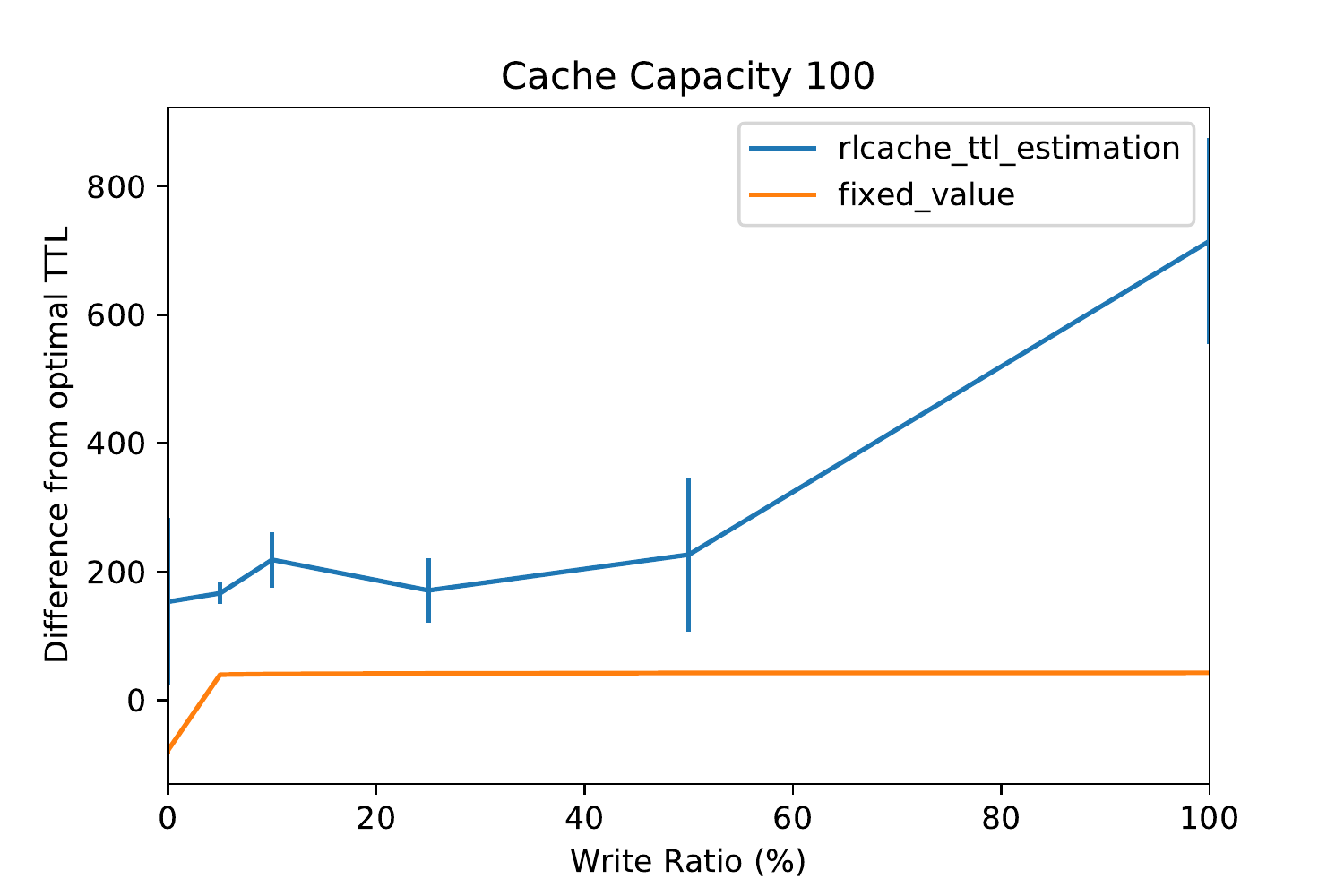}
	\end{subfigure}%
	\begin{subfigure}{.5\textwidth}
		\centering
		\includegraphics[width=1.0\textwidth]{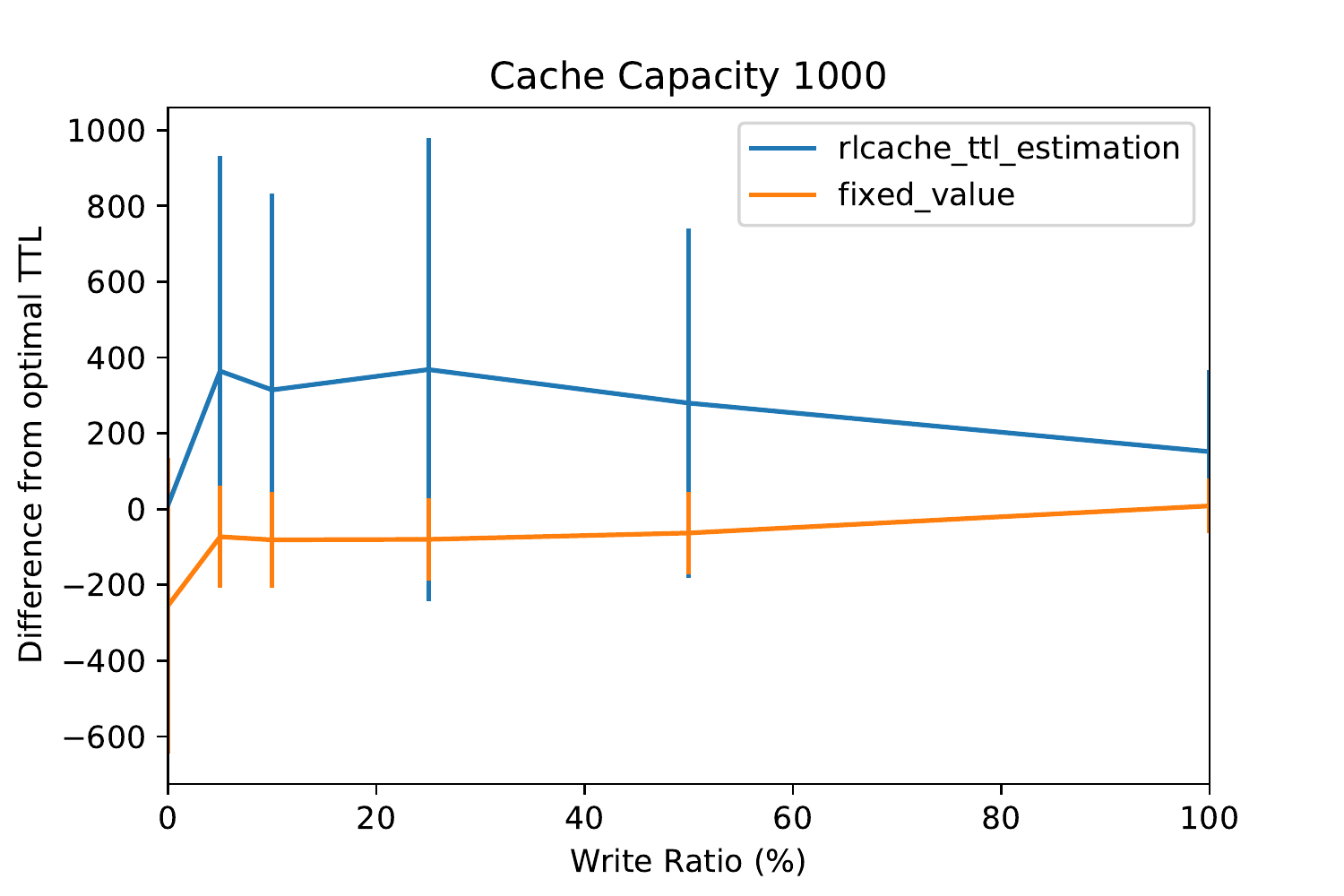}
	\end{subfigure}
	\begin{subfigure}{.5\textwidth}
		\centering
		\includegraphics[width=1.0\textwidth]{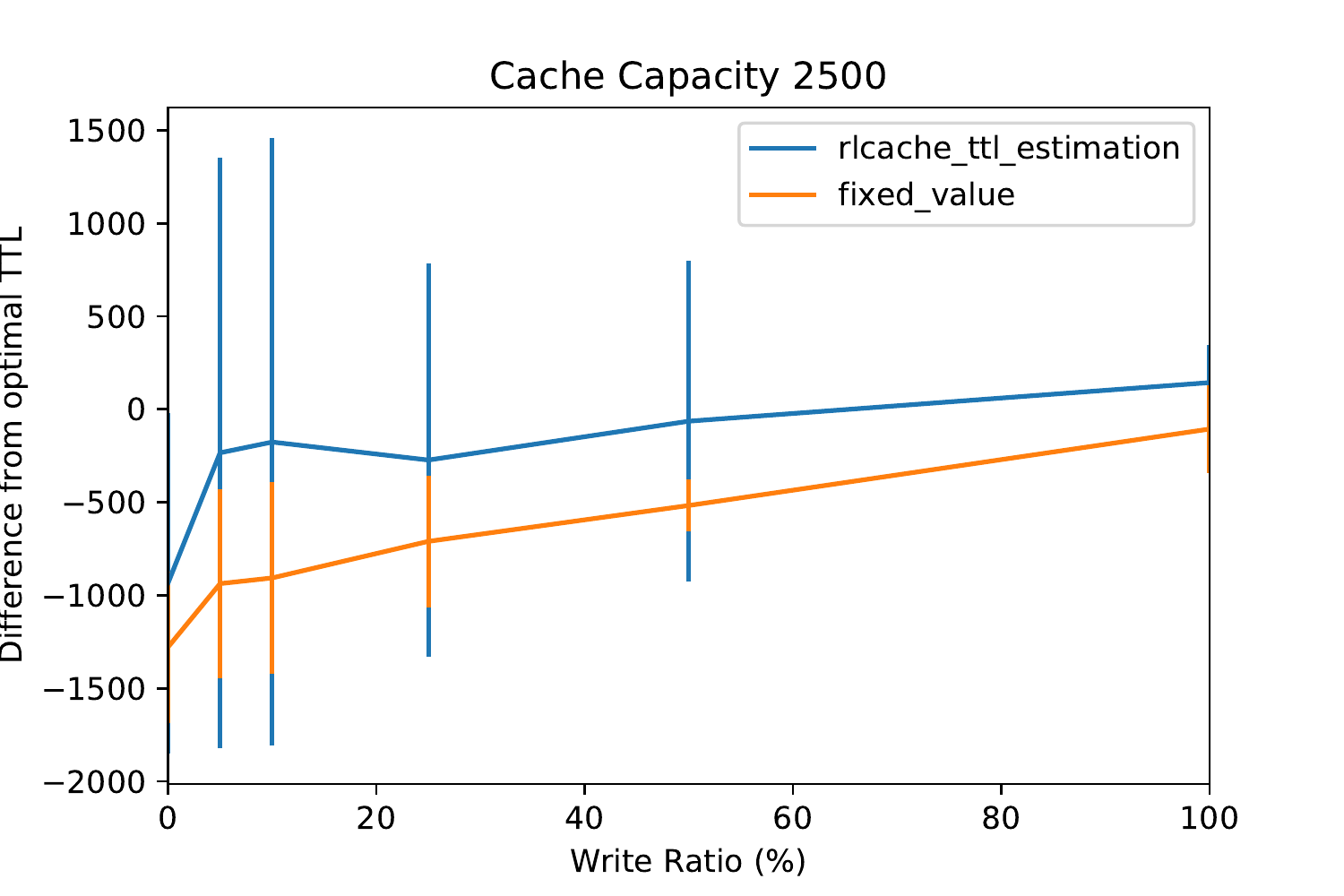}
	\end{subfigure}%
	\begin{subfigure}{.5\textwidth}
		\centering
		\includegraphics[width=1.0\linewidth]{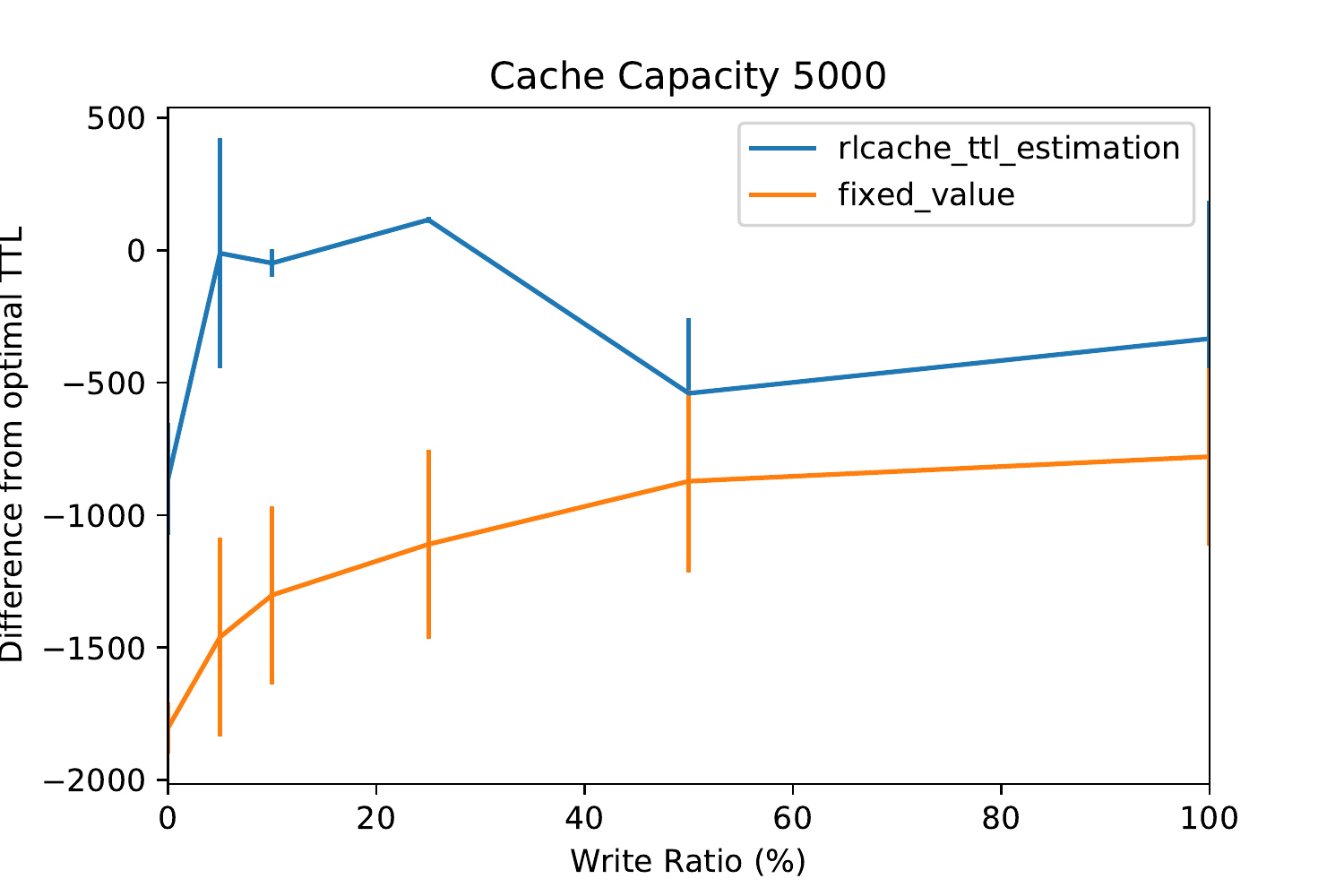}
	\end{subfigure}
	\caption{TTL difference between optimal value and TTL estimation strategy.}
	\label{fig:ttl_diff}
\end{figure}

\begin{figure}[!htb]
	\centering
	\begin{subfigure}{.5\textwidth}
		\centering
		\includegraphics[width=1.0\textwidth]{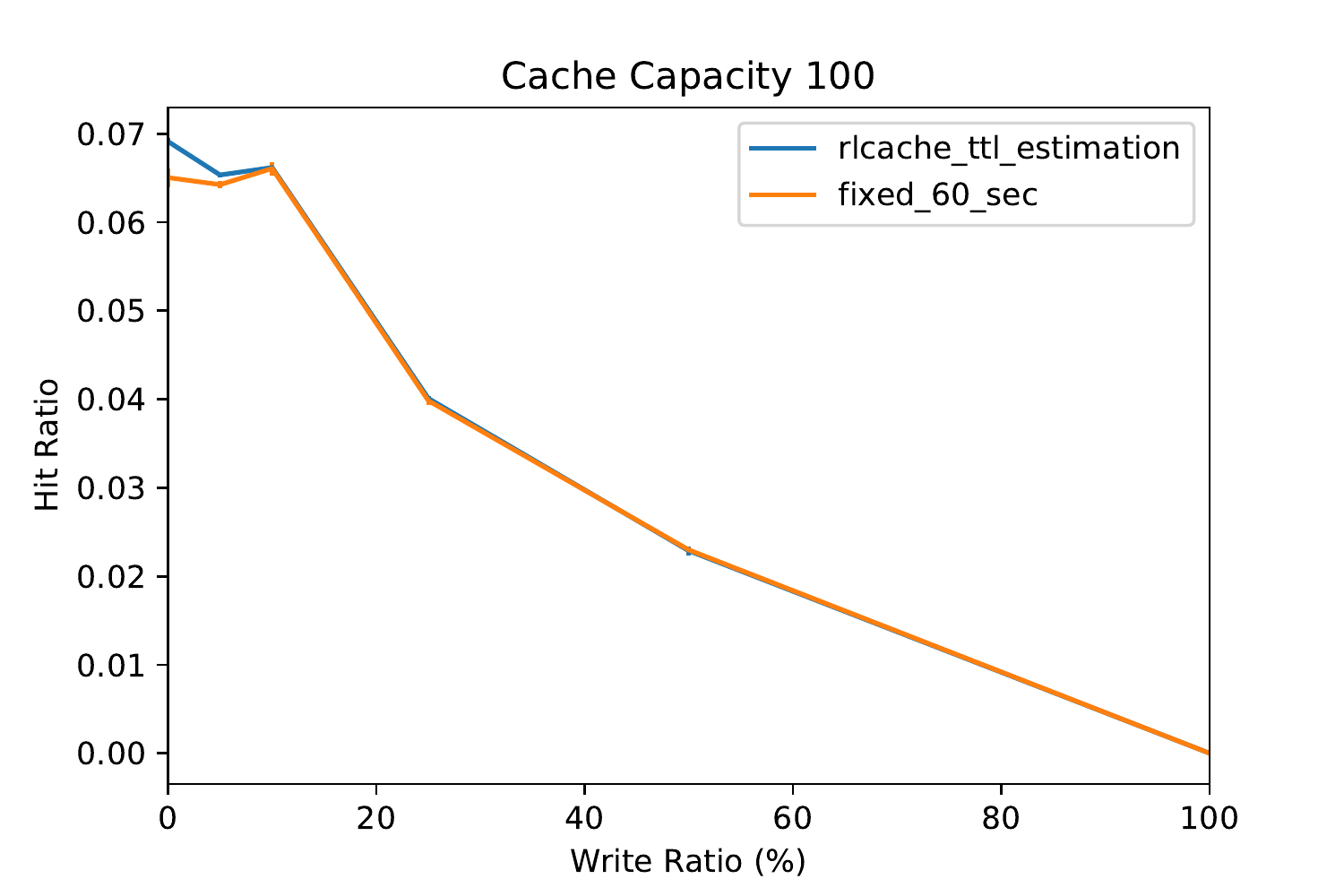}
	\end{subfigure}%
	\begin{subfigure}{.5\textwidth}
		\centering
		\includegraphics[width=1.0\textwidth]{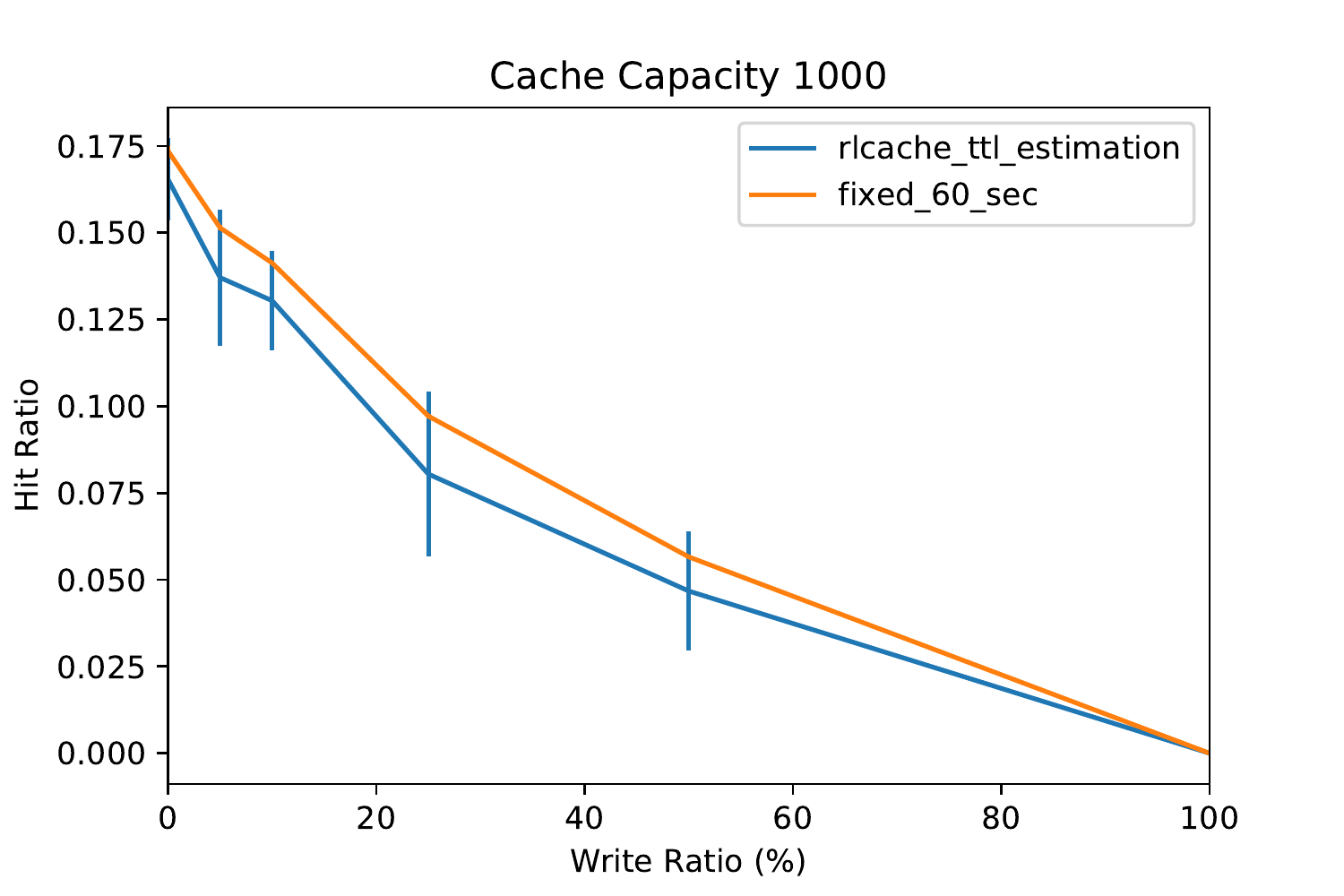}
	\end{subfigure}
	\begin{subfigure}{.5\textwidth}
		\centering
		\includegraphics[width=1.0\textwidth]{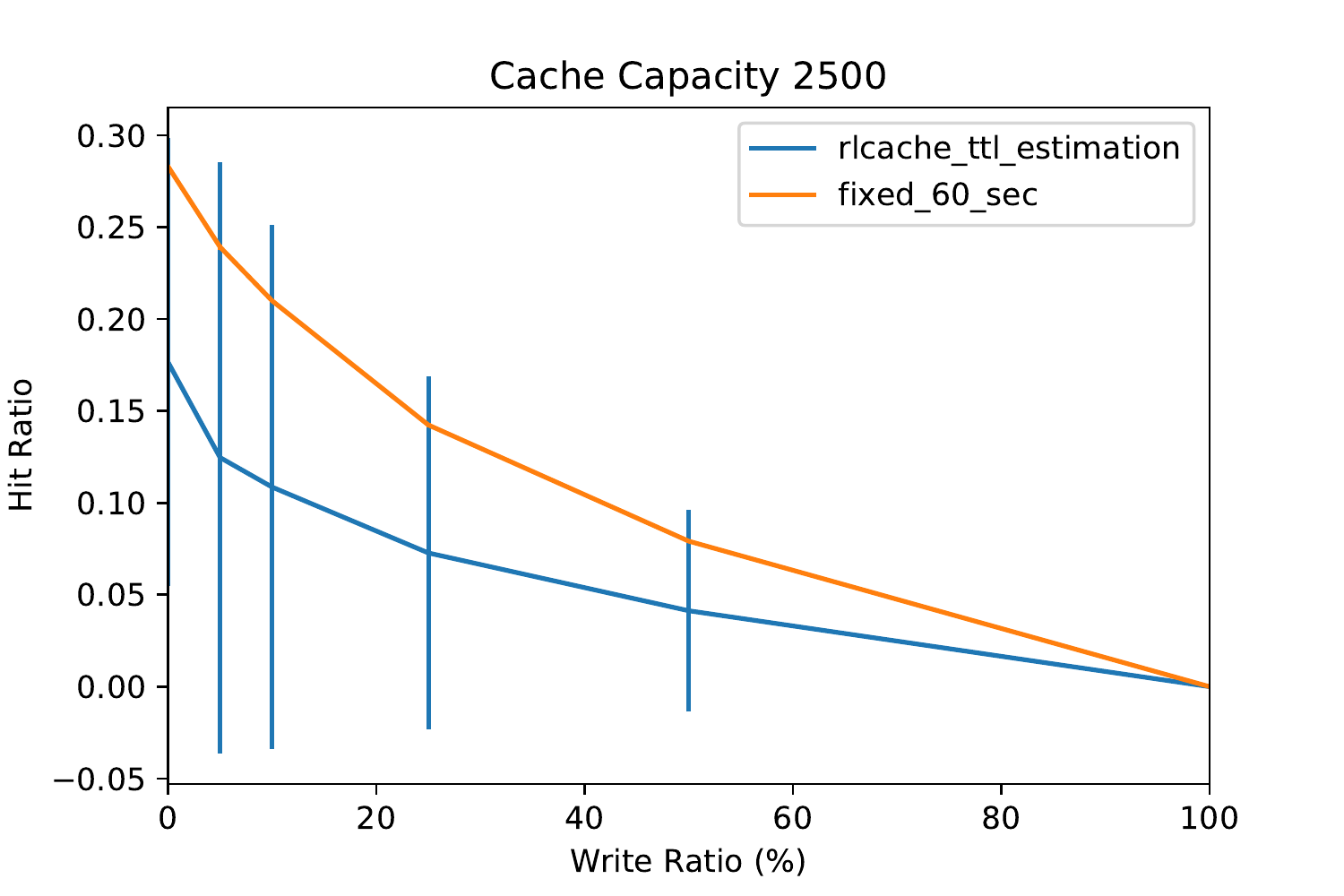}
	\end{subfigure}%
	\begin{subfigure}{.5\textwidth}
		\centering
		\includegraphics[width=0.8\linewidth]{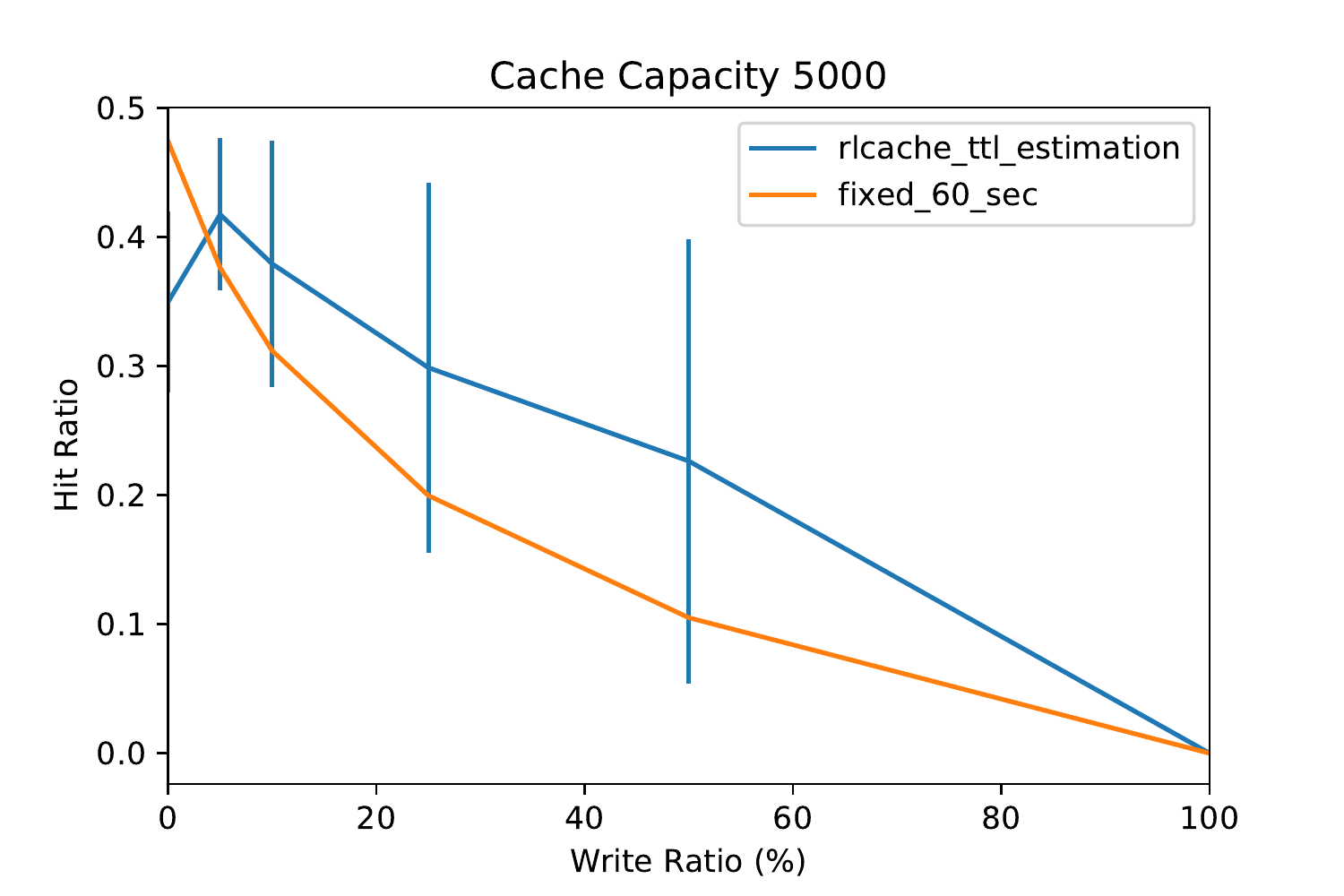}
	\end{subfigure}
	\caption{Cache hit-rate comparison when using a dynamically adjusted TTL strategy.}
	\label{fig:ttl_hit}
\end{figure}

\textbf{Experiments.} We evaluated the performance of this agent by how far off from an optimal TTL it is.
The optimal TTL is the duration between multiple hits and invalidation if the object does not receive any hits in $30$ minutes, the optimal TTL is then set to $0$ to discourage caching objects for longer than they are needed.
\autoref{fig:ttl_diff} demonstrates the learning of the RLCache overtime, and it is approaching an optimal TTL despite changes in the workload.
In the plot, the value that is closer to zero yields better results, as y-axis is set to the difference between the optimal TTL and target algorithm.
Initially, the agent explores by overestimating the TTL value and explores further by underestimating the value, before it settles on overestimating on average as that yield better cache hit rate.
Despite the agent's ability to learn over time, achieving $100\%$ estimation of optimal TTL is very difficult due to several factors, including the cache is limited in size and some values will be forced to leave the cache despite what the TTL agent predicted.
Furthermore, optimal TTL is a continuous value and the simulators samples from a distribution.
Therefore, the exact timing is not guaranteed to be a constant value that can be predicted. 
Dynamically adjusted TTL struggles to learn correct values when the cache capacity is limited (1\% of the backend size) due to eviction policy interfering with the TTL policy and evicting entries despite the estimated TTL, causing the agent to learn undesirable values, it is evident when the write ratio increases and therefore the evictions also increases

Figure \ref{fig:ttl_hit} shows that cache hit-rate with a dynamically adjusted TTL. 
The benefit of using dynamically adjusted TTL is visible in larger cache sizes.
RLCache learns near optimal TTL estimation in a larger cache. 
When the TTL estimation is near optimal the objects are less likely to stay in the cache longer than needed and the cache will not be over-populated, and as a result, the objects will not be evicted prematurely and they will live to near maximum useful lifespan, which increases to the cache hit-rate.

\textbf{Outro.} RLCache dynamically adjusts its TTL estimation and predicts near-optimal TTL despite changes in the workload pattern. 
It achieves higher cache-hit rate as a result and reduces the time objects needs to occupy the cache for (or increase it if they are useful objects).
Accurately predict a TTL is a significant property in push-based caches such as CDN.
In CDN sending an invalidation request to objects is very expensive and instead, system developers relay on setting a TTL through trial-and-error. 

\subsection{Multi-Task and Multi-Agent}
\textbf{Intro. }
The previous evaluations looked at caching, eviction, and TTL estimation decisions, each of these decisions had a direct impact on the cache system and sub-sequentially the cache hit-rate.
By improving the caching decision, fewer objects will need to be evicted as fewer objects will be put in the cache. 
Similarly, by improving TTL estimation, fewer objects have to be manually evicted.
When the eviction decision is smarter, fewer objects will cause cache miss resulting in fewer objects that need to be evaluated by the TTL estimator and cache decision maker.
The complex interaction between these agents motivates this experiment to combine all actions into one. 

This evaluation is mostly experimental, as far as we are aware, the literature lacks applications of multi actions systems outside toy problems. 
The goal of this experiment was to investigate the possibility of combining all agents to get the best of all in one setup.
There are two architectures that we looked at that facilitate the multi-action paradigm.
\begin{itemize}
	\item \textbf{Multi-agent learning} This architecture uses a separate agent for each task, then a meta (controller) agent to coordinate the output of the action from each agent with the goal that maximises the expected reward for all of the agent. Section \ref{sec:multi_agent} goes over the details of how this agent works.
	\item \textbf{Multi-task learning} In this architecture, a single agent learns how to perform multiple separate actions using a single neural network. 
	This architecture discussed in depth in \autoref{sec:multi_task}.
\end{itemize}

\subsubsection{Multi-agent}
\begin{figure}[!htb]
	\centering
	\includegraphics[width=0.7\linewidth]{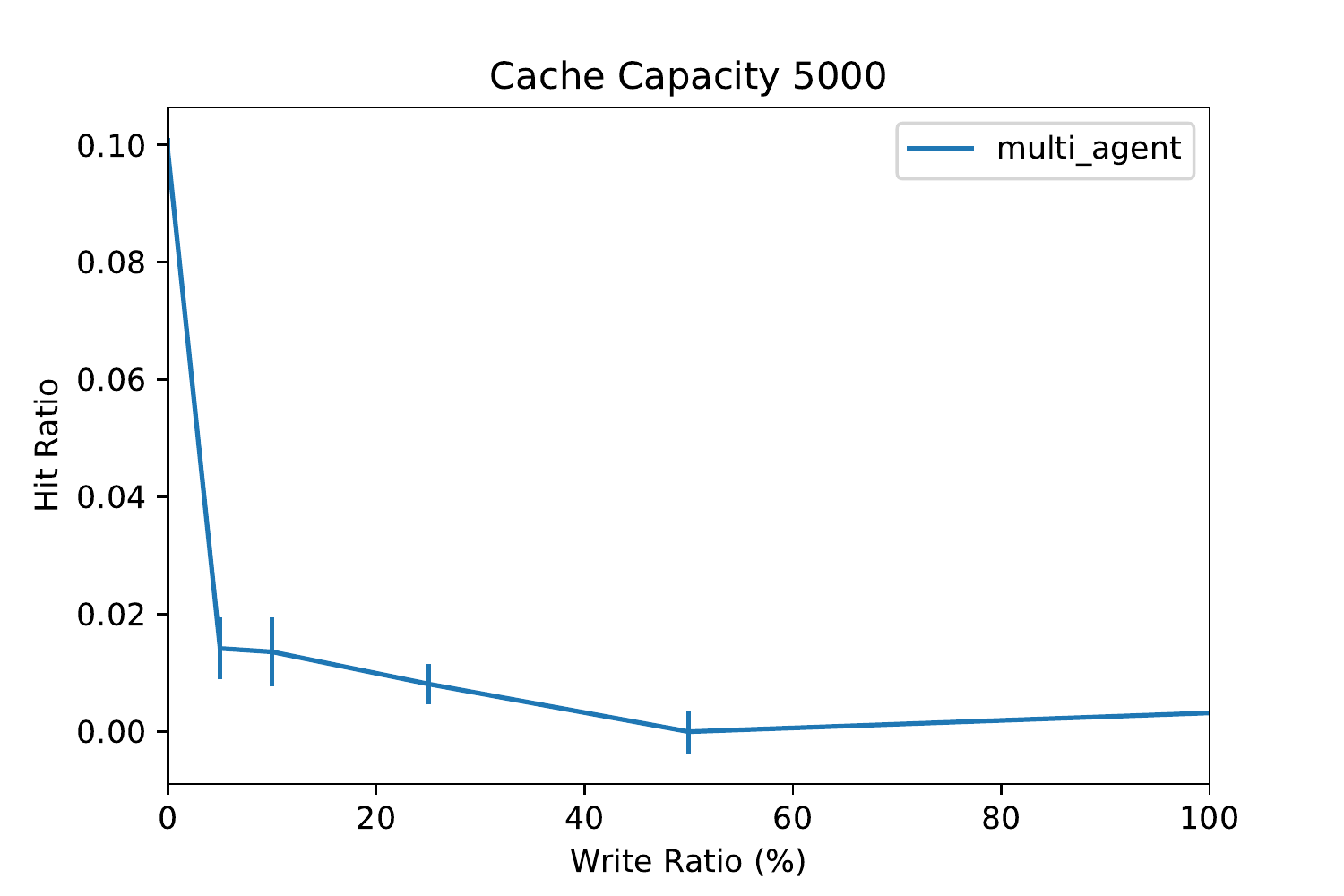}
	\caption{Cache hit-rate for multi-agent while varying workload.}
	\label{fig:multiagent_hitrate}
\end{figure}

\textbf{Experiments. }\autoref{fig:multiagent_hitrate} shows the result of evaluating the multi-agent by itself.
As can be seen, the results are underwhelming, while the agents share the same reward for the goal of maximising the cache hit rate and reducing space, they do so by invoking contradictory actions.
For example, a TTL agent estimates that an object should have a long lifespan, but then the eviction agent evicts, leaving no space for either agents to learn from their actions.
There are two possible mitigation to this problem.
The first, when inspecting the plot at closer scale, it becomes apparent that none of the agents converged, this is caused by low number of training samples, as not only each of the four agents requires a large number of samples to learn its task, but also need to learn the other agents actions and predict them before making decisions.
This can be mitigated by running an expensive experiment over several weeks that produces enough training sample for all agents to converge. 
However, due to time and resources constraints, we decided against doing so.
The alternative approach is using a model-based RL to run significantly faster simulation inside each of the agents and cause them to converge in fewer training samples.
We discuss this approach at length in the future work section \autoref{sec:future_model}.

\subsubsection{Multi-task}
\textbf{Intro. }Multi-task agent combines the neural network between all CM tasks and aims to learn multiple tasks at the same time.
This approach works well when the tasks have similarities between them.
In RLCache the tasks are set to maximise the same objective function and learn the same key features in each unique cache entry, e.g. learning the hotkeys of the cache.
To demonstrate the practicality of multi-task agent, we look at four key metrics: 
\begin{itemize}
	\item \textbf{Cache hit-rate}: The percentage of the requests that are in the cache when clients query it, it is the main objective function.
	\item \textbf{Caching rate}: Indicates how many objects were cached to achieve the cache hit-rate.
	A higher value is better when the workload is read heavy, and lower is better when the workload is write-heavy.
	\item \textbf{F1 score}: Visualises the accuracy of an eviction decision (higher is better).
	\item \textbf{Optimal TTL}: The difference between the TTL's optimal value and the estimated value, the value that is closer to zero is better.
\end{itemize} 

\begin{figure}[!htb]
	\centering
	\includegraphics[width=0.7\linewidth]{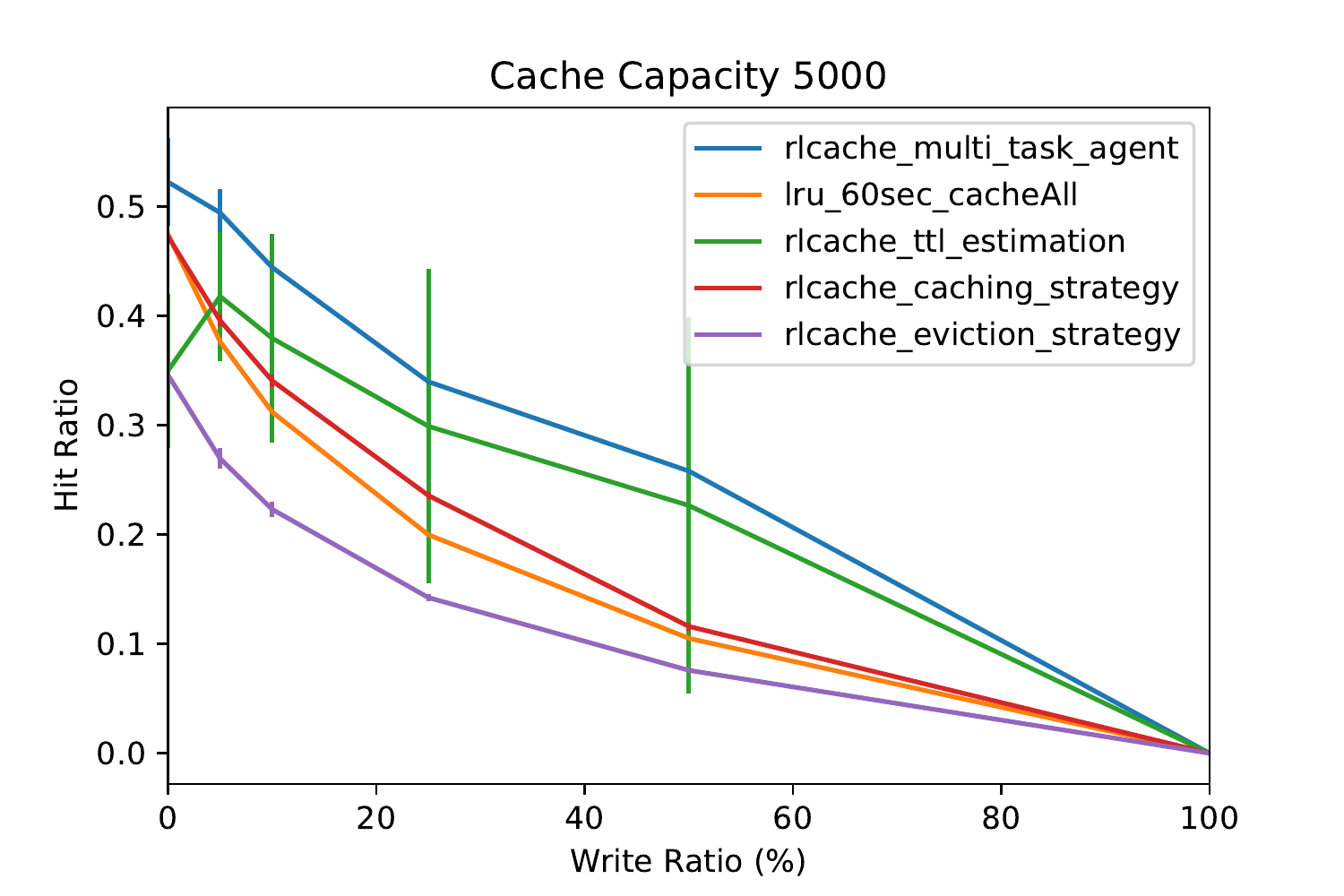}
	\caption{Cache hit-rate for multi-task agent on various workloads.}
	\label{fig:multi_task_hit_rate}
\end{figure}
\textbf{Experiments.} First, figure \ref{fig:multi_task_hit_rate} visualises the cache hit-rate and compares it to all previously mentioned methods.
The multi-task agent outperforms all previously mentioned methods even when varying the write ratio,  showing that the combination of all tasks, when optimised and learnt together, achieve a higher cache hit rate, and this is the main hypothesis this work was studying.
To ensure that is indeed the case, we look at the other metrics.

\begin{figure}[!htb]
	\centering
	\includegraphics[width=0.7\linewidth]{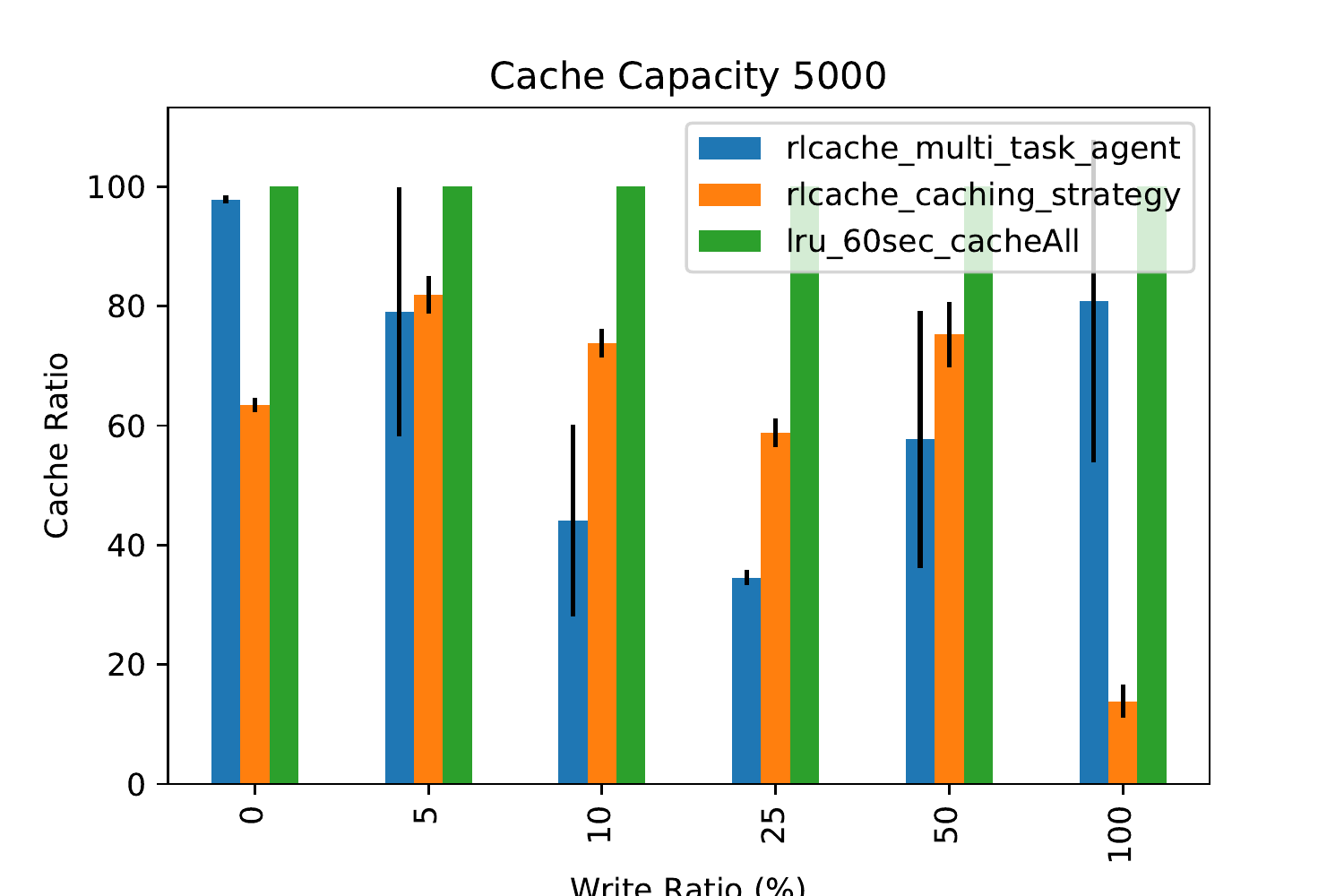}
	\caption{Caching rate for multi-task agent on various workloads.}
	\label{fig:multi_task_cache_rate}
\end{figure}
\autoref{fig:multi_task_cache_rate} shows the caching rate of objects. 
While the multi-task agent did not outperform the caching decision specialist agent, it still achieved significant improvements over the baseline.
In our experiment, the exploration parameter starts at 100\% of all decisions are sampled uniformly, but decays over time until it reaches $10\%$, and the agent reaches that value around the time the $25\%$ workload is executed.
The agent at that time learns the weights that work well for when there is a significant percentage of the requests being a read request, as a result when the workload switches to a heavy-write workload, the agent did not have the chance to update its policy.
Therefore, if the CM expects a sudden shift in the workload all the time, a higher exploration lower bound (i.e. $25\%$) allows the agent to react faster to those shifts.
If the CM has a large number of observation samples or the changes in the workload are not as fast, a lower exploration bound (like the one we used) provide better consistent performance for when the agent converges.

\begin{figure}[!htb]
	\centering
	\includegraphics[width=0.7\linewidth]{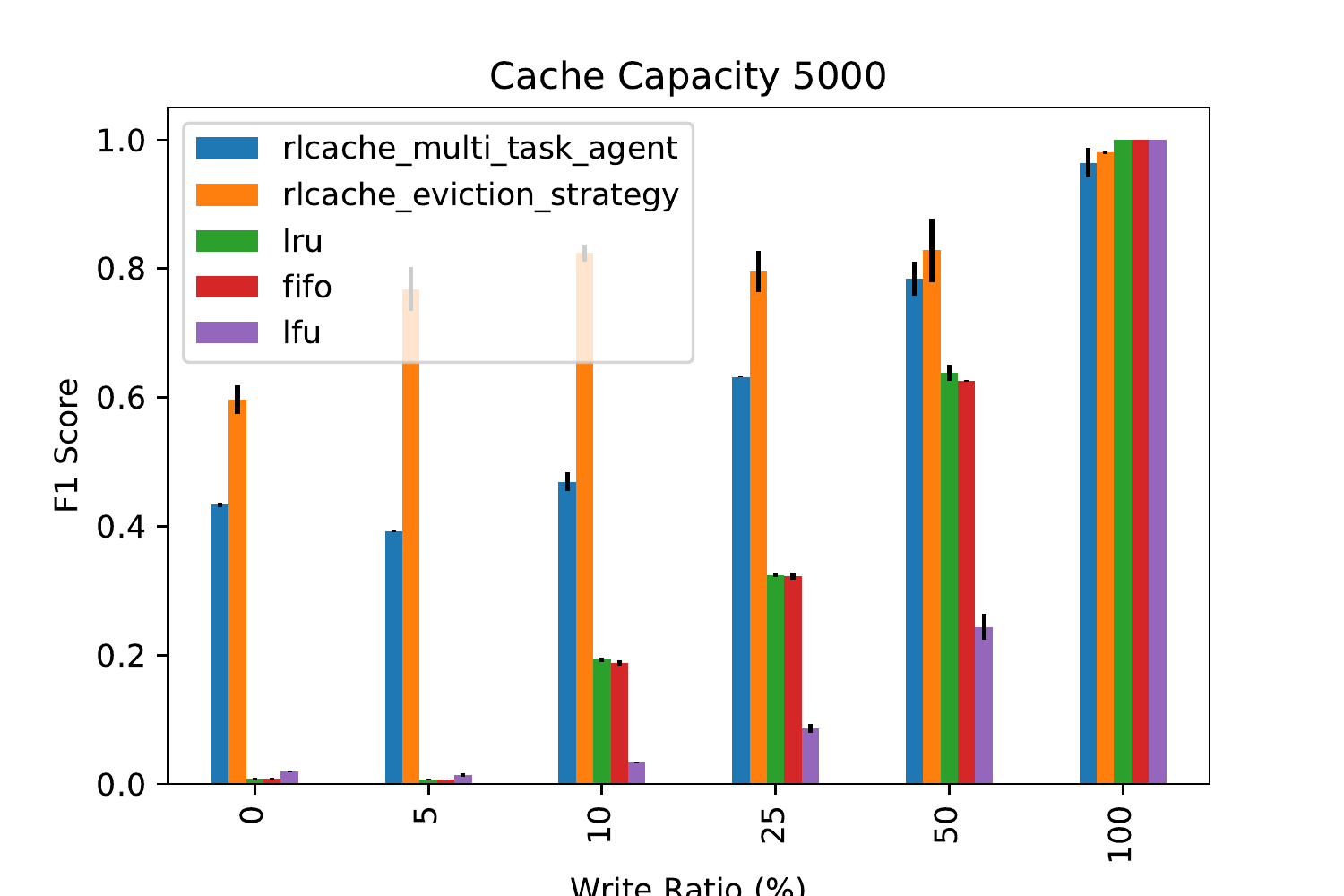}
	\caption{F1 score of multi-task agent on various workloads.}
	\label{fig:multi_task_f1}
\end{figure}
Next, we look at the eviction task that figure \ref{fig:multi_task_f1} shows.
The F1 score shows the trade-off between precision and recall for evicted objects. 
We compared RLCache multi-task agent against RLCache single-task agent and the baselines.
The figure shows that the single-task agent's F1 score outperforms the multi-task agent's score.
That is expected since the single-agent needs to learn a subset of what the multi-tasks needs to learn. Additionally, the single agent reward function is hand-crafted that captures all edge cases and guides the agent, while in the case of multi-task it is challenging to design similar reward function due to challenges we discussed earlier in \autoref{sec:multi_task}.
However, the multi-task agent was significantly better than the baselines, and when the workload shifted its traffic patterns, the multi-agent was able to adapt just as well as the single task agent.

\begin{figure}[!htb]
	\centering
	\includegraphics[width=0.7\linewidth]{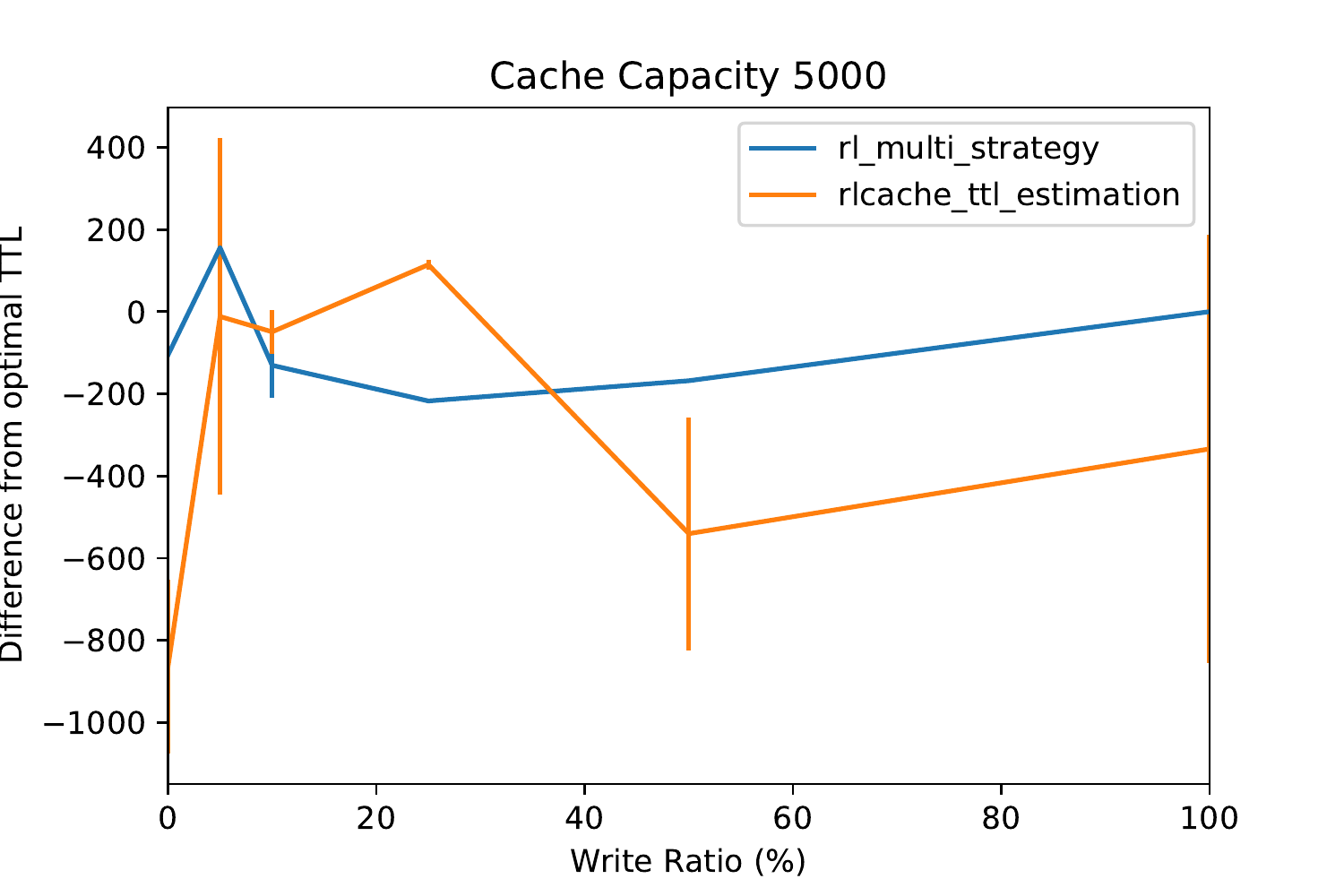}
	\caption{TTL difference between optimal value and RL multi-task agent.}
	\label{fig:multi_task_ttl_diff}
\end{figure}
Finally, we look at the TTL estimation task of this agent.
\autoref{fig:multi_task_ttl_diff} compares the accuracy of the TTL estimation of both the RLCache single-task agent and the multi-task agent.
This task heavily depends on the other two tasks because an accurate TTL estimation needs information about what objects are likely to be evicted or cached.
As a result, the multi-task, in fact, outperforms the single-task (unlike the other two tasks) and estimates almost near optimal TTL for all the generated workload.

\textbf{Outro.} The multi-task agent showed that a fully automated cache management system outperforms all other methods significantly, due to the knowledge sharing between tasks and ability to predict the actions of the other tasks at any point in time.
While the multi-task agent is not able to outperform the single-agent in their own sub-tasks, we hypothesise, given enough training samples the multi-task agent will perform similar to the single task agent in their respective tasks, and as a result, produces even better overall cache-hit rate, however, such an experiment is expensive to conduct and confirm.

\subsection{RL Hyperparameter Initialisation}
\textbf{Intro.} The goal of this experiment to investigate how susceptible RLCache is to the random initialisation of the RL hyperparameters.
Henderson et al. \cite{henderson2018deep} investigations show that RL is well known to be sensitive to the random initialisation, as a stochastic approximator are used at various levels of any RL algorithm. 

\textbf{Experiments.}
We repeated the caching strategy evaluation $13$ times and visualised it, the interesting metric to look at is the error bar as that shows the deviations between runs and as such the impact of the random seed.
\begin{figure}[!htb]
	\centering
	\begin{subfigure}{.5\textwidth}
		\centering
		\includegraphics[width=1.0\textwidth]{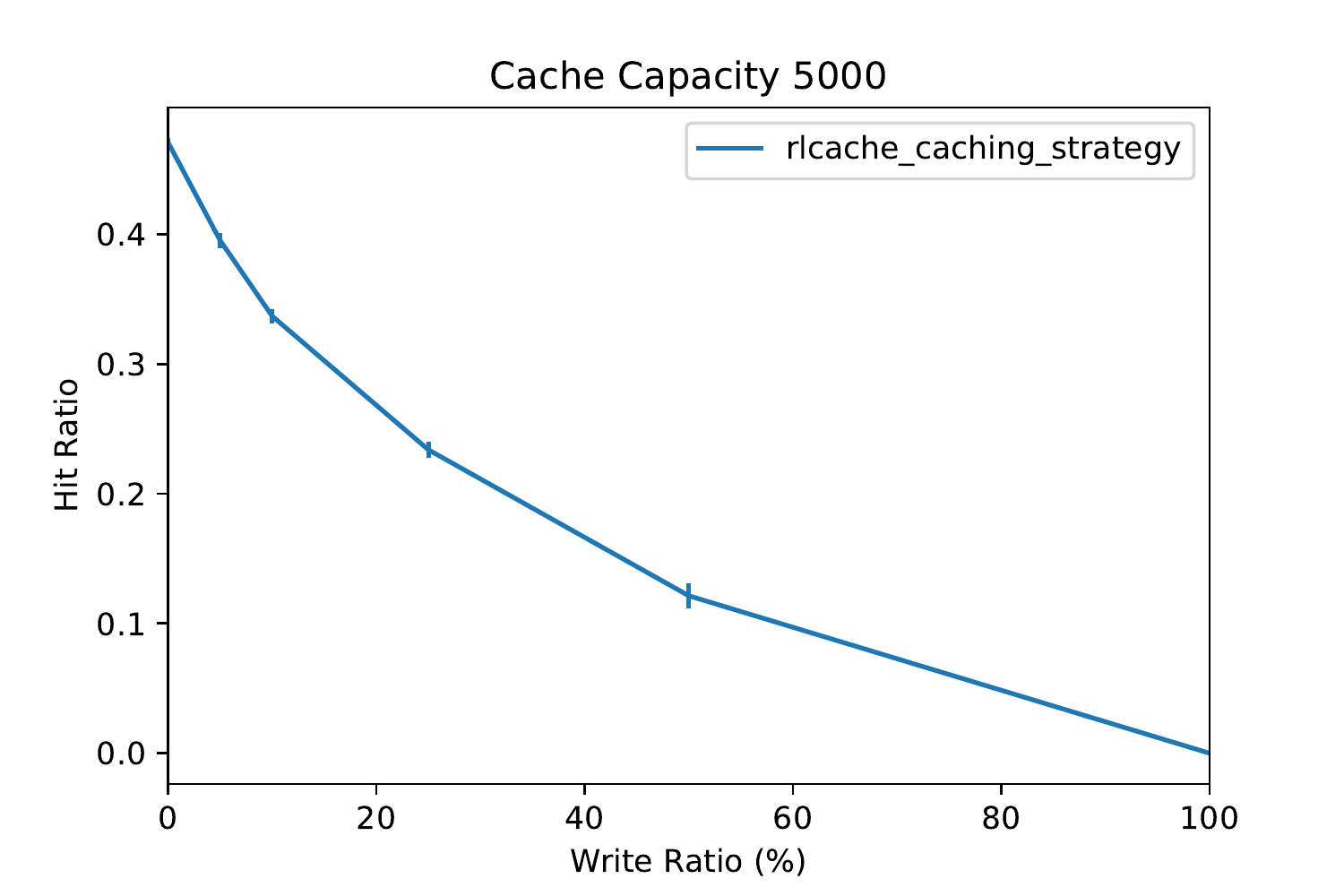}
	\end{subfigure}%
	\begin{subfigure}{.5\textwidth}
		\centering
		\includegraphics[width=1.0\textwidth]{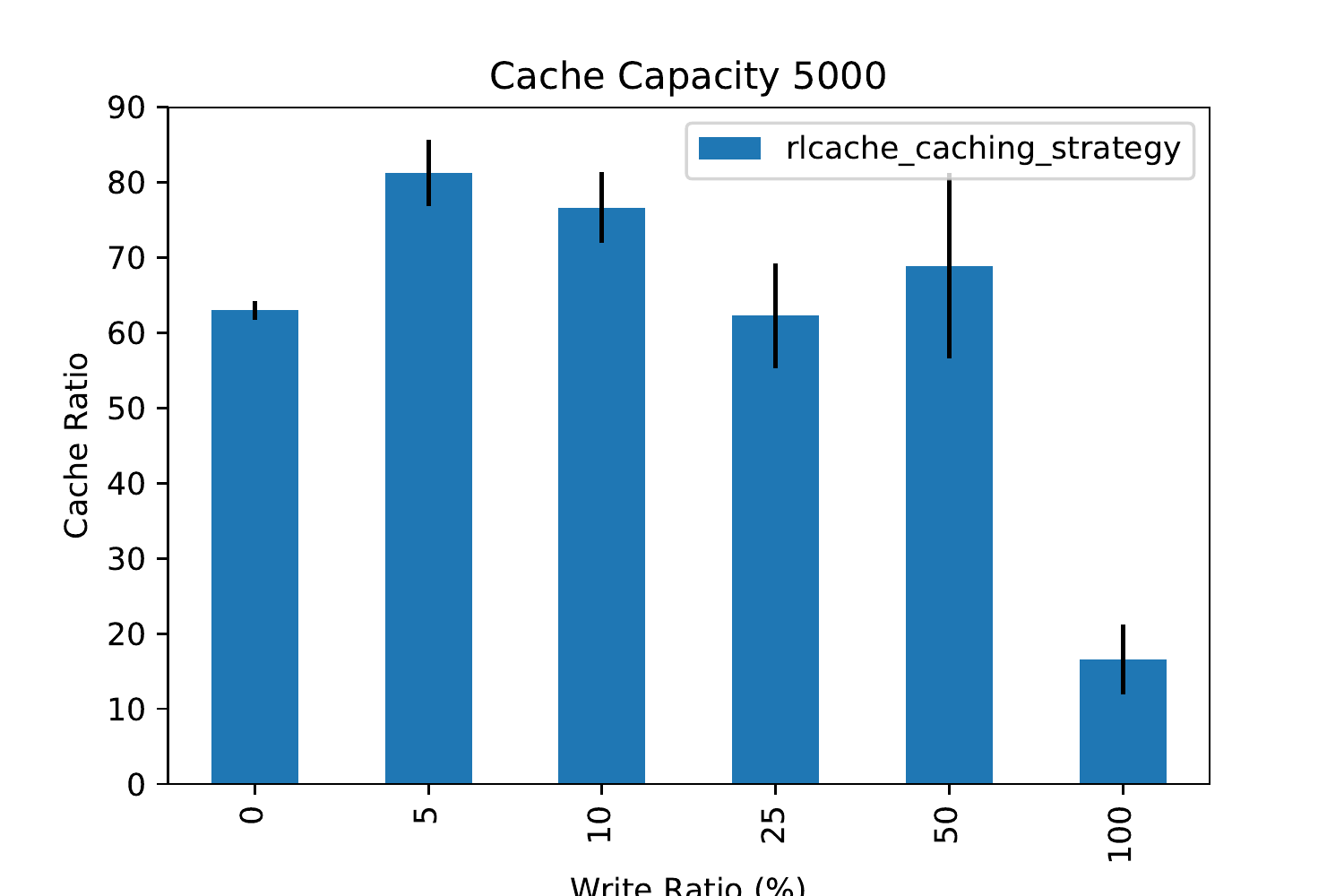}
	\end{subfigure}
	\caption[Hyperparameter initialisation impact on hitrate]{The hyperparameter initialisation impact on cache hitrate.}
	\label{fig:repeatedcacherate5000}
\end{figure}

\begin{figure}[!htb]
	\centering
	\includegraphics[width=0.7\linewidth]{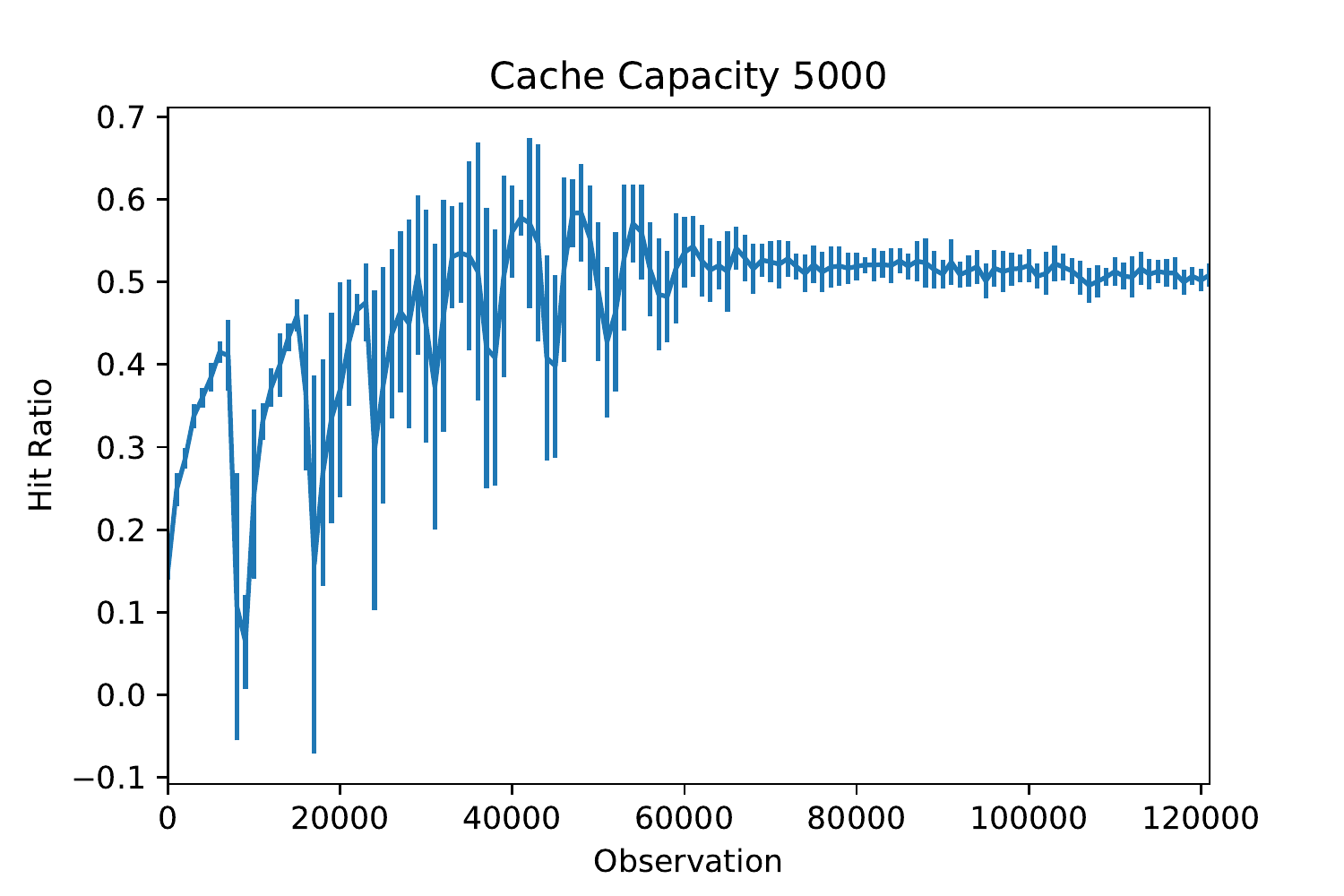}
	\caption[Hyperparameter initialisation impact on hitrate per step]{The hyperparameter initialisation impact on hitrate per 1000 observation.}
	\label{fig:repeatedhitratezoomed5000}
\end{figure}

The first figure \ref{fig:repeatedcacherate5000} shows the impact against the objective functions the cache-hit rate and cache rate, the largest deviation is observed when the read-write ratio is at $50\%$ the hypothesis, as at that workload depending on the seed the agent can fall in either of the local maxima that favours the reads or the one that favours the write.
Otherwise, the agent is robust and has a low standard deviation.

\autoref{fig:repeatedhitratezoomed5000} shows the impact when looked at as per observation, in this figure it is obvious that the agent first $60,000$ observations are the most susceptible to random seed and the deviations can vary massively between steps, however the agent soon coverages and stabilises after, showing its ability to recover from a bad seed.

\textbf{Outro. } 
The experiment showed that while random initialisation has a big impact on the initial decisions. 
Over time the agent converges towards the optimal policy and recovers from a bad seed. 

\subsection{Outlook}
This chapter evaluated this thesis against claims made in the introduction, and summarised in the following points:
\begin{itemize}
	\item RLCache is an online control that adapts its caching strategies to changes in the workload and with information about the system such as capacity.
	\item RL-driven decisions outperform heuristic-driven one:
	\begin{itemize}
		\item \textbf{Caching decisions: }Achieved similar cache hit-rate as the baseline doing so while using significantly less caching operations. 
		\item \textbf{TTL estimation: } Estimated a near-optimal TTL while achieving a higher hit-rate than comparable baseline.
		\item \textbf{Cache eviction: } Accurately identifying objects that will not be used again, a property that is useful when evicting the wrong object is a costly operation.
	\end{itemize}
	\item Demonstrated that multi-task learning, a single RL agent learnt all the actions needed to manage the cache effectively, achieving better results in all benchmarks.
	\item While not directly visible in the evaluation, but these agents were only possible because of the RLObserver ability to link delayed experiences and distribute them to interested agents.
	\item Showed that the stochastic initialisation of RL algorithm only has impact at very first few observations, after which the system able to adapt to the changes in the workload and recover from a bad initialisation seed.
\end{itemize}

The key takeaway is that RL is useful for computer system problems when applied to the right problem. 
It enables a new class of computer systems that adapts to the changes in the workload and learn complex interactions between various components of the system.
\chapter{Summary and Conclusions}\label{chapter:summary} 
\graphicspath{{05-summary/figs/}}

\section{Conclusion}
This work showed the result of applying reinforcement learning to a general purpose cache manager.
RLCache estimates the time-to-live for cached objects, predict the advantage of caching an object, and which object to evict if the cache is full.
RLCache captures the objective of maximising the cache hit-rate while minimising populating the cache with useless objects.
We developed three independent RL agents that each is responsible for one of the three tasks and a multi-task agent that learns all three tasks at the same time.
Furthermore,  we designed a framework to tackle the problem of delayed experience in computer systems.
The framework allows the system to subscribe as many agents to the delayed observation, and abstract the complication associated with the delay of state transition in computer systems.

We evaluated our results using a database workload generator, YCSB,  that provides an established benchmark on various workloads databases.
RLCache adapts to changes in workloads pattern and outperforms the hit-rate of traditional methods while minimises the number of objects in the cache. 
We identified three potential future works for RLCache, applying it in a distributed setting, improving the training time by building a RL-model of the environment, or incorporating it with an enterprise storage system.

\section{Future work}
\subsection{Distributed Cache Manager}\label{section:distributed_cache_manager}
This work introduced a single instance manager that has a full view of all cache transactions, that single instance has the responsibility to learn from the observations and inference to do actions.
Scaling-out is a concern when an application grows such that there is a large number of keys that do not fit in a single instance of a CM, or there are availability requirements.

Decoupling the RL learning and inference processes is required to allow the CM to scale out.
The observer architecture discussed in \autoref{sec:observer} simplifies the process of decomposing the system into multiple independent logical chunks.
RLObserver framework commits transactions and observation to logs, that later synchronised across all distributed listeners, the synchronisation delay of the logs is mitigated by using off-policy agents that learn from the data without interacting with the environment.

The workflow of RLCache in a distributed setting is the following:
A worker is assigned the responsibility of conducting inference operations only.
Every new observation and system metrics are pushed to all the workers' queue through the fan-out architecture described in \autoref{sec:observer}.
The distributed workers then learn by consuming the observations from their independent queues.
At set intervals, all workers to sync their weights with the main CM.
This architecture is typical in distributed systems and allows the system to scale effortlessly and reliably.

The future work for this is to investigate distributing the cache, study the latency impact of using RLCache, and bundle the CM into an established cache system such as Redis \cite{carlson2013redis} or Memcached \cite{jose2011memcached}.

We expect the RL induced latency to remain a bottleneck even in a distributed setting.
For example, Kraska et al. \cite{kraska2018case} proposed an approach to learn an indexing data structure, while their proposal resulted in promising results, they identified a critical bottleneck in the inference time it takes for an indexing decision to be made.
RL is still an evolving field and its toolchains are written in Python, as such they are not easily parallelisable and are ill-suited for mission-critical fast response tasks.
Rewriting the core engine of the CM and the inference engine in a high-performance language (e.g. C++) will further improve the performance of the cache.
Once the CM is both distributed and parallelised, it can operate in mission-critical systems.
 
 
\subsection{Model-based RL}\label{sec:future_model}
The work here uses a model-free RL approach to learn caching related decisions. 
While it is a flexible and powerful approach, it requires a large number of observations before the agent learn meaningful actions.
Lack of observations is a big issue for actions that are infrequent, for example, evicting cached object.

In a model-based RL, the agent interacts with the environment and build a model that generates more training samples that the agent uses to learn from actions without having to experience them in the environment, which results in a faster converging.
There are two main approaches to model-based RL: Given a model of the environment, the agent learns through self-play to collect training samples as done in AlphaZero \cite{silver2017mastering}. 
Alternatively, if the model does not exist, World Models \cite{ha2018world} learns features of the environment in an unsupervised manner and creates a compressed version of the environment, it then trains the agent inside the compressed representation before outputting actions in the real environment.

The difficulty with this approach is constructing an accurate model as an inaccurate model will cause the agent to learn sub-optimal actions.

\appendix
\singlespacing

\bibliographystyle{abbrv} 
\bibliography{references} 

\end{document}